\pdfoutput=1

\documentclass[11pt]{article}

\usepackage[]{EMNLP2023}

\usepackage{times}
\usepackage{latexsym}

\usepackage[T1]{fontenc}

\usepackage[utf8]{inputenc}
\usepackage{enumitem}

\usepackage{microtype}

\usepackage{inconsolata}

\usepackage{multirow}
\usepackage{array}
\usepackage{graphicx}
\usepackage{longtable}
\usepackage{booktabs}
\usepackage[most]{tcolorbox}

\title{Pragyaan: Designing and Curating High-Quality Cultural Post-Training Datasets for Indian Languages}

\author{ Neel Prabhanjan Rachamalla, Aravind Konakalla, Gautam Rajeev, \\  \textbf{Ashish Kulkarni, Chandra Khatri, and Shubham Agarwal } 
\\ \\
Krutrim AI, Bangalore, India\\ \\
\textsuperscript{Contact: \{neel.rachamalla1, ashish.kulkarni, shubham.agarwal1\}@olakrutrim.com} 
}

\begin{document}
\maketitle
\begin{abstract}

The effectiveness of Large Language Models (LLMs) depends heavily on the availability of high-quality post-training data, particularly instruction-tuning and preference-based examples. Existing open-source datasets, however, often lack multilingual coverage, cultural grounding, and suffer from task diversity gaps that are especially pronounced for Indian languages. We introduce a human-in-the-loop pipeline that combines translations with synthetic expansion to produce reliable and diverse Indic post-training data. Using this pipeline, we curate two datasets: \textit{Pragyaan-IT} (22.5K) and \textit{Pragyaan-Align} (100K) across 10 Indian languages covering 13 broad and 56 sub-categories,  leveraging 57 diverse
datasets. Our dataset protocol incorporates several often-overlooked dimensions and emphasize task diversity, multi-turn dialogue, instruction fidelity, safety alignment, and preservation of cultural nuance, providing a foundation for more inclusive and effective multilingual LLMs.


\end{abstract}


\section{Introduction}

Recent developments around Large Language Models (LLMs) \citep{touvron2023llama,grattafiori2024llama,abdin2025phi, guo2025deepseek} have demonstrated that post-training data, comprising both instruction-tuning and preference data, plays a critical role in enhancing model alignment, task generalization, and usability \citep{ouyang2022training, bai2022constitutional, chung2022scaling}. 
Particularly in a multilingual and multicultural landscape, like India, 
the availability of high-quality, culturally grounded post-training data is crucial to address performance gaps in low-resource languages that often arise from the scarcity of relevant and representative training data \citep{joshi2020state}.


While several open-source datasets for post-training \citep{longpre2023flan, selfinstruct, bercovich2025llamanemotronefficientreasoningmodels} exist, they are predominantly English-centric and often suffer from limitations such as inconsistent quality, restricted coverage,  insufficient task complexity, and limited multilingual coverage. These challenges extend to Indic post-training data as well, 
focus of our work. 


Direct translations of existing English post-training datasets are prone to translation biases, errors \cite{hartung2023measuring,savoldi2021gender,muennighoff2022crosslingual} and loss of cultural grounding~\citep{wang2022measuring,pudjiati2022post}. For instance, a prompt like \textit{``Tell me about a small herb to plant in backyard"} might yield Western herbs such as \textit{thyme or rosemary}, whereas Indian users would expect culturally familiar options like \textit{tulsi (holy basil), pudina (mint), or curry leaves}. Similarly, when asked \textit{``What is a good comfort meal for a rainy day?"}, English-centric answers such as \textit{tomato soup or grilled cheese} overlook Indian preferences like \textit{masala chai with pakoras or khichdi with ghee}. Even in wellness contexts, \textit{``Recommend a workout routine for beginners"} may default to \textit{squats and push-ups}, neglecting practices like \textit{Surya Namaskar or yoga exercises}. Such mismatches highlight the need for post-training data that reflects not just language, but also local traditions and cultural context.




The recent popularity of LLM-based synthetic data generation \citep{wang2022self} for creating post-training datasets, while promising, still suffers in quality due to 
linguistic inaccuracies, grammatical inconsistencies, and reduced fluency, especially in multilingual settings, that could degrade the performance of models trained on them. Moreover, the lack of fine-grained control over output complexity and the potential for hallucination can lead to the generation of low-quality, unreliable data. 





With the aim of addressing these gaps, we present an approach to curate high-quality post-training datasets, especially in multilingual settings. Our curation approach combines the above techniques with post-hoc manual editing, leading to a scalable human-in-the-loop pipeline, with specific focus on several aspects of quality, like task coverage, multilingual representation, task complexity, culture, multi-turns, reasoning, and others. We leverage our approach to curate high-quality Indic post-training datasets: \textit{Pragyaan-IT} comprising $22.5K$ instruction tuning examples and \textit{Pragyaan-Align}, a dataset of $100K$ preference examples in $10$ Indian languages covering $56$ task categories. 
Our contributions could thus be summarized as follows:

\begin{figure*}[htbp]
    \centering
\includegraphics[width=1\linewidth]{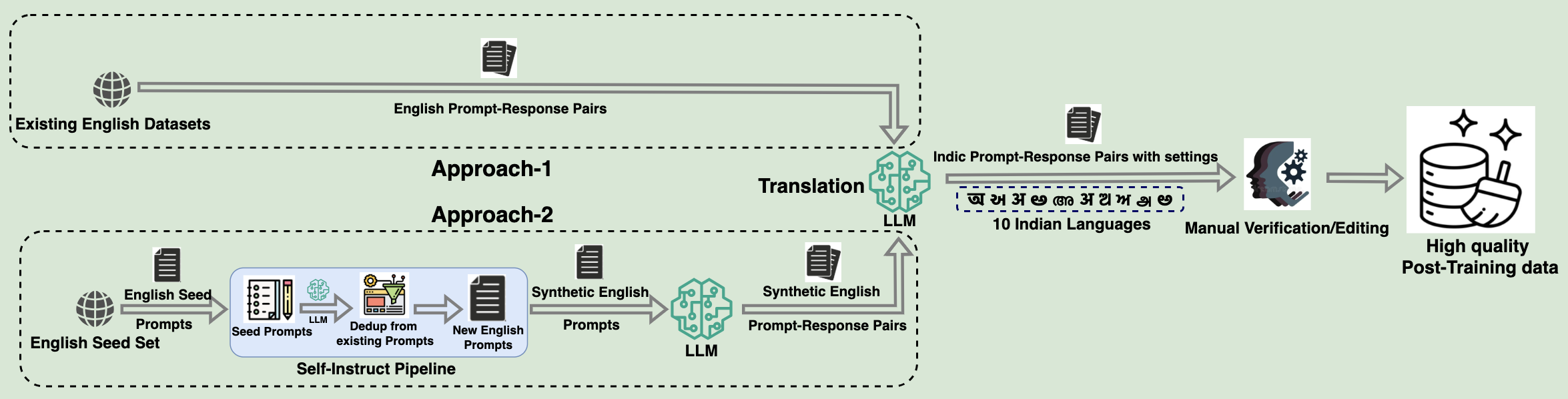} \caption{Workflow for building Indian language post-training data: English prompts are either translated or expanded via modified self-instruct pipeline to generate synthetic prompts. In both cases, responses are then produced with an LLM, translated into one of the 10 Indian languages, and manually refined (Section \ref{sec:method}).}
    \label{fig:approaches_pipeline}
\end{figure*}


%
\begin{itemize}[noitemsep]
    \item We present a scalable pipeline for curating high-quality post-training data. Our approach emphasizes a human-in-the-loop (HITL) that is more efficient,  reliable and ensures higher quality than direct synthetic generation or translation of existing English datasets.
    \item We introduce high-quality, manually curated, and culturally-inclusive post-training \textit{Pragyaan} dataset series, consisting of 1) $22.5K$ \textit{Pragyaan-IT} and 2) $100K$ \textit{Pragyaan-Align}, 
    designed for aligning LLMs to the diverse Indian cultural context.
    \item Our dataset includes a broad spectrum of instruction-following tasks with varying levels of complexity, ensuring the resulting models can handle a wide range of real-world scenarios. We provide a detailed analysis of the dataset's characteristics, including its language distribution and domain representation, also showcasing its suitability for robust instruction-following capabilities through a small-scale pilot experiment.
\end{itemize}



\section{Related Work}

Post-training is a key step in aligning large language models (LLMs) with human intent, commonly achieved through instruction tuning \citep{wei2022finetuned} and preference tuning \citep{bai2022training} datasets, constructed in several ways. Task template based resources such as Flan 2021 \citep{wei2022finetuned}, Flan 2022 \citep{longpre2023flan}, and P3 \citep{sanh2022multitask} adapt NLP datasets into instruction–response format. Human-authored datasets like Open Assistant \citep{kopf2023openassistant}, Dolly \citep{DatabricksBlog2023DollyV2}, and LIMA \citep{zhou2023lima} demonstrate the value of curated instructions but face scalability challenges. To overcome this, synthetic generation approaches leverage LLMs to expand from small human-annotated \textit{seeds}, with efforts such as Self-Instruct \citep{wang2022self}, Alpaca \citep{alpaca}, and Guanaco \citep{joseph_cheung_2023}, often distilling knowledge \citep{hinton2015distilling} from stronger teacher LLM models. Advanced pipelines like Evol-Instruct \citep{xu2023wizardlm} iteratively increase instruction complexity, while later works extend these methods to reasoning and code generation \citep{luo2023wizardcoder,gunasekar2023textbooks}. Complementing these, user-contributed datasets such as InstructionWild \citep{instructionwild} and ShareGPT\footnote{\url{https://sharegpt.com/}} provide naturally occurring conversational data, and Unnatural Instructions \citep{honovich2022unnatural} show how seed tasks can be scaled into diverse synthetic corpora.
Subsequent work expanded into specialized domains, including dialogue systems \citep{kopf2023openassistant}, structured knowledge grounding \citep{xie2022unifiedskg}, and chain-of-thought reasoning \citep{wei2022chain,kim2023cot}. 
More recently, the Magpie dataset \citep{argilla2024magpie} introduced a fine-grained taxonomy spanning creative writing, math, role-playing, planning, and data analysis, emphasizing the importance of broad coverage in post-training resources. Preference datasets such as UltraFeedback \citep{cui2023ultrafeedback} and Tulu3 \citep{lambert2024tulu3} comprises human and synthetic preference pairs for LLM alignment. Building on these advances, we construct our approach and dataset tailored to Indian languages and cultural contexts leveraging manual annotations in complement with synthetic generation.

While large-scale post-training datasets have become increasingly available, they remain predominantly English-centric, with limited coverage for other languages. A few exceptions incorporate some proportions of multilingual data \citep{kopf2023openassistant,  longpre2023flan, muennighoff2023octopack, zhuo2024astraios, oig2023}, but they remain limited in cultural and linguistic diversity compared to English resources. Prior efforts to extend post-training resources beyond English have typically followed three strategies: (1) translating English datasets into additional languages \citep{li2023bactrian, khan2024indicllmsuite}, (2) generating template-based datasets \citep{yu2023large, gupta2023targen}, and (3) manually curating instruction datasets in non-English languages \citep{li2023m3it, wang2022supernaturalinstructions}. Amongst these, template-based efforts such as xP3 \citep{muennighoff2022crosslingual} extend the P3 taxonomy with 28 multilingual datasets. However, xP3 relies on uniform templates across languages, leading to limited task diversity and frequent repetition.
Translation-based approaches face similar limitations, such as Bactrian \citep{li2023bactrian}, which translated Alpaca \citep{alpaca} and Dolly \citep{DatabricksBlog2023DollyV2} into 52 languages. In contrast, our work introduces human-edited datasets across 10 Indian languages, addressing issues of redundancy and cultural grounding while providing a more diverse and representative resource for multilingual alignment.

\section{Methodology}
\label{sec:method}
We present our multi-stage post-training dataset creation process that encompasses a variety of task categories at both broad and fine-grained levels, critical settings (complexity, interaction depth, constraints, safety, indian context, thinking trails) and leverages human annotators alongside curated data sources to ensure quality and coverage. While our approach itself is generic, we discuss how we used it to create high-quality Indic post-training datasets that we collectively refer to as \textit{Pragyaan-IT} and \textit{Pragyaan-Align}.



\subsection{Data Construction Approaches}

We employ two complementary approaches (Figure \ref{fig:approaches_pipeline}) that both combine translation and synthetic generation with post-hoc manual editing to ensure linguistic accuracy, fluency, and cultural appropriateness in our datasets.

\subsubsection{Approach 1: Translation with Human Refinement}
Here, we directly source English prompt-response pairs from existing English datasets (more details later in Section \ref{subsec:dataset}).

\textbf{Prompts:} 
We begin with English prompts which
are first translated into Indic languages using an LLM, then refined by human annotators. During verification, annotators correct linguistic errors, improve readability, and adapt expressions where needed to reflect Indian cultural norms. This results in two categories, i.e.\ \textit{1) Indic Generic Prompts:} direct translations of the English originals.
\textit{2) Indic Context Prompts:} culturally adapted and edited versions incorporating Indian references and contexts by human annotators.

\textbf{Responses:}
Corresponding English responses undergo a similar pipeline independently, with LLM-based translation into Indic languages, followed by human editing for grammar, relevance, length, and cultural appropriateness. Thus, we have
\textit{1) Indic Generic Responses:} literal translations of the English outputs.
\textit{2) Indic Context Responses:} refined versions adapted to Indian discourse norms.

\subsubsection{Approach 2: Synthetic Expansion with Human Refinement}

This approach introduces an additional intermediate stage of synthetic prompt expansion in English. 

\textbf{Prompts:}
Starting with a seed set of English prompts (sourced or created), we use the Self-Instruct pipeline \citep{selfinstruct} to iteratively expand this set into a larger synthetic pool.
While in the original pipeline they generate new prompts for classification and non-classification types via different strategies, in our adaptation, we use the same prompt template for both the cases. 
The resulting synthetic English prompts are then translated into Indic languages with LLMs and refined by human annotators to ensure correctness, clarity, and cultural grounding. This yields \textit{1) Synthetic Indic Generic Prompts:} literal translations of synthetic English prompts. \textit{2) Synthetic Indic Context Prompts:}  culturally enriched translations.

\textbf{Responses:}
For each synthetic English prompt, we generate English responses using an LLM independently. These are then translated into Indic languages and refined through human editing. Annotators correct factual or linguistic errors, polish style, and, when appropriate, enrich with cultural nuances. This process produces \textit{1) Synthetic Indic Generic Responses:} faithful translations of the English responses. \textit{2) Synthetic Indic Context Responses:} culturally adapted versions aligned with the Indian local context. 

While our first approach ensures fidelity through the translation of existing English datasets, it remains constrained in both scope and diversity. The second approach complements this by introducing synthetic expansion prior to translation, enabling broader task coverage, richer cultural representation, and greater scalability with reduced dependency on large English resources. 
Together, the two approaches strike a balance between reliability and diversity, yielding multilingual datasets that are both high-quality and contextually rich. 

\subsection{Task Categories}
\label{sub:categories}

Building on the design principles of several existing datasets, we curate a broad set of task categories that 
combine core language tasks such as reasoning (limited to CoT and self-thinking), inference, natural language understanding (NLU) and generation, question answering (QA), dialogue and interaction, information extraction, mathematics, coding, function calling, and instruction following, while also extending into culturally grounded domains like Indian states, religions, geo-political questions, etc. 
To promote robustness and responsible deployment, we additionally include safety and non-compliance, Indian contentious content, and self-identity tasks. Collectively, these categories establish a structured yet comprehensive framework that spans diverse sub-categories, enabling richer and more inclusive post-training curation. A detailed breakdown of sub-categories is provided in Table~\ref{tab:task_categories}.

\begin{table*}[htbp]
\tiny
\centering
\begin{tabular}{p{3cm}p{3cm}p{3.7cm}p{4.7cm}}
\hline
\textbf{Broad Category} & \textbf{Sub Categories} & \textbf{Broad Category} & \textbf{Sub Categories} \\ \hline

Reasoning \& Inference & \begin{tabular}[c]{@{}l@{}} Non-Math Reasoning \\ Math Reasoning \\ Code Reasoning \\ Indian Relationships \\ Inference \\ Data Analysis \end{tabular} 
& Natural Language Understanding \& Generation & \begin{tabular}[c]{@{}l@{}} Named Entity Recognition \\ Text Classification \\ Grammar Correction \\ Translation \\ Creative Writing \\ Paraphrase Identification \\ Paraphrase Generation \\ Text Summarization \\ Headline Generation \\ Question Generation \\ Sentiment Analysis \end{tabular} \\ \hline

Question Answering & \begin{tabular}[c]{@{}l@{}} General Question Answering \\ Fact Check \end{tabular}
& Interaction \& Dialogue & \begin{tabular}[c]{@{}l@{}} Multi Turn Conversation \\ Role Playing \\ Advice Seeking \\ Planning \\ Brainstorming \end{tabular} \\ \hline

Information Extraction & \begin{tabular}[c]{@{}l@{}} Information Seeking \\ Indian Cultural Context \\ Comprehension \end{tabular}
& Sanskrit Cultural \& Creative Usage & \begin{tabular}[c]{@{}l@{}} Sanskrit Festival Greetings \\ Sanskrit Auspicious Day / Other Occasions \\ Sanskrit Subhashitas (Quotes) \\ Sanskrit Captions and Mottos \\ Sanskrit Person's Name \\ Sanskrit Building / Institution / Company Name \\ Sanskrit Product Name \end{tabular} \\ \hline

Mathematics & \begin{tabular}[c]{@{}l@{}} Math QA \\ Math Instruction Tuning \\ Math Proofs \end{tabular}
& Coding & \begin{tabular}[c]{@{}l@{}} Code Generation \\ Code Debugging \\ Code Editing \\ Code Explanation \\ Code Translation \\ Unit Test Generation \\ Code Theory \\ Code Review \\ Repository level Code Generation \end{tabular} \\ \hline

Function Calling & Function Calling
& Instruction Following & Instruction Following \\ \hline

Safety \& Non-Compliance & Safety \& Non-Compliance
& Indian Contentious Questions & \begin{tabular}[c]{@{}l@{}} Indian Geo Political \\ Indian Politicians \\ Indian States \\ Indian Languages \\ Indian Religions \end{tabular} \\ \hline

Self-Identity & \begin{tabular}[c]{@{}l@{}} Model-name \\ Person-based \end{tabular}
& & \\ \hline

\end{tabular}
\caption{Taxonomy of NLP task categories and sub-categories for creating post-training datasets in Section \ref{sub:categories}.}
\label{tab:task_categories}
\end{table*}


\subsection{Task Settings}
\label{subsec:settings}
For each task category, we additionally define 
several task settings that encourage diversity of prompt complexity, interaction depth, instruction-following, safety considerations, cultural grounding, and explicit reasoning trails. We provide systematic descriptions of these settings next. 

\subsubsection{Complexity}

We categorize tasks by complexity to ensure models are trained for both simple and challenging scenarios, with two primary levels: \textit{1) Easy} prompts are direct and clearly defined, usually requiring minimal reasoning (e.g.\, a single factual or descriptive query). \textit{2) Hard} prompts feature greater structural complexity, often embedding multiple sub-questions within a single query, requiring nuanced reasoning and fine-grained understanding. Importantly, complexity is defined within each task category, enabling fair assessment across heterogeneous task types (see Figures \ref{fig:complexities_1}–\ref{fig:complexities_6} in Appendix).




\subsubsection{Multi-Turn Interactions}

Multi-turn settings capture tasks where contextual continuity is critical, such as dialogue, planning, or role-play. These scenarios require models to maintain memory of prior turns while generating coherent and adaptive responses. We consider three levels of interaction depths: \textit{1) Single-turn (1 turn):} A response to an isolated prompt; \textit{2) Short multi-turn (3 turns): }Three back-and-forth exchanges, ensuring local continuity; \textit{3) Extended multi-turn (5 turns):} Five exchanges, for long-range memory and coherence in extended conversations (e.g.\, planning a festival with evolving constraints).





\subsubsection{Instruction Following}
We categorize instruction-following into three levels, defined by the number and type of constraints imposed on the response such as \textit{``answer in 100 words''}, \textit{``respond in json format''}, etc. (Figure~\ref{fig:constraints} in Appendix). This ensures coverage of tasks that range from loosely guided prompts to highly structured outputs. Different combinations of these constraints are applied depending on the nature of the prompt: \textit{1) Simple instruction following}: prompts include minimal or no explicit constraints on the format or content of the response; \textit{2) Medium instruction following}: prompts introduce two to three explicit constraints, requiring the model to accommodate multiple conditions at once; \textit{3) Complex instruction following}: prompts impose several simultaneous constraints, demanding precise control and structured outputs.





\subsubsection{Safety}

We define safety settings to ensure that models behave responsibly when faced with sensitive, controversial, or potentially harmful content in a real-world setting. This dimension helps guide appropriate responses while maintaining ethical standards. We include \textit{1) Safe}: prompts are neutral and non-controversial, allowing the model to provide direct answers without ethical or policy concerns; \textit{2) Non-safe}: prompts involve sensitive or harmful material, where the model is expected to either refuse politely (e.g., \textit{``Sorry, I cannot assist with that ...''}) or generate a safe response.




\subsubsection{Thinking Trails}
We define `thinking trail' settings to capture the role of explicit reasoning in model responses, ensuring that outputs range from direct answers to more reflective reasoning styles.
\textit{1) Normal}: direct response generation without intermediate reasoning traces; \textit{2) Chain-of-Thought (CoT)}: step-by-step reasoning \cite{wei2022chain} articulated explicitly before the final answer; \textit{3) Self-Thinking}: inspired by recent ``deep thinking'' paradigms \cite{guo2025deepseek,bercovich2025llama, abdin2025phi}, where models produce more elaborate, self-reflective reasoning trails prior to the final response.




\subsubsection{Indian Cultural Context}
Given the centrality of Indic languages and cultural alignment in our framework, we explicitly model contextual grounding through three progressively richer levels. \textit{1) IC-1} represents generic prompts leading to generic responses, with no explicit India related anchoring (e.g., Prompt: \textit{``Suggest some breakfast items.''}, Response: \textit{``Pancakes, cereal, toast, scrambled eggs.''}). Such responses are accurate but remain culturally neutral, with no particular alignment to cultural settings. \textit{2) IC-2} represents generic prompts that nonetheless yield Indic-grounded responses. For instance, the same prompt above, in this setting, would elicit responses such as \textit{``Idli, dosa, paratha, poha''}, which are the most popular breakfast items in India. 
\textit{3) IC-3} involves prompts that are themselves explicitly Indic, thereby eliciting fully Indic-based responses. For example, the prompt itself mentions ``Suggest some Indian breakfast items'' with a similar response as in the IC-2 setting. This setting encourages grounding and diversity of responses with respect to Indian cultural context that is required for training Indic focused LLMs. 

\begin{figure*}[htbp]
    \centering
    \begin{minipage}[t]{0.472\textwidth}
        \centering
        \includegraphics[width=\linewidth]{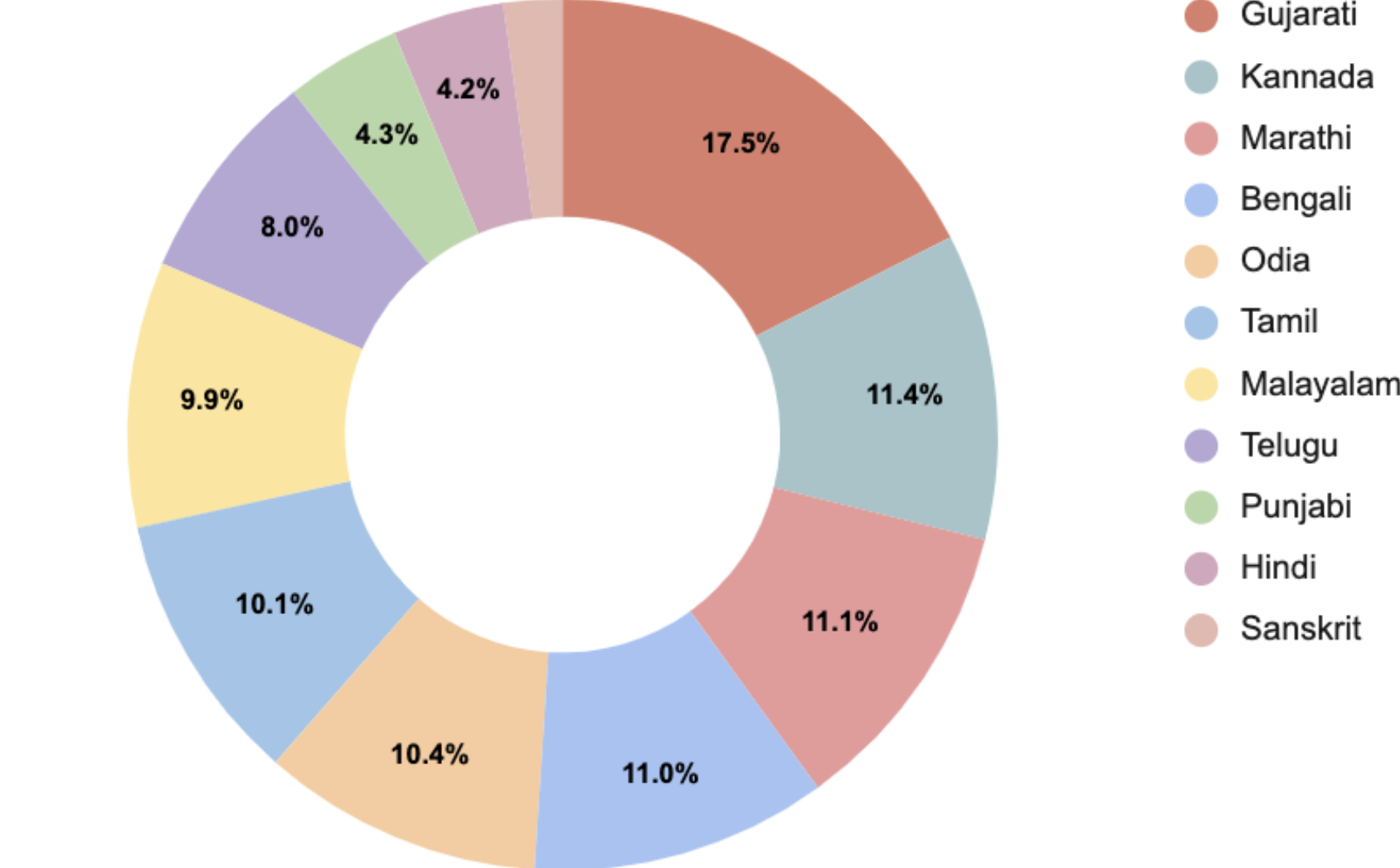}
    \end{minipage}%
    \hfill
    \begin{minipage}[t]{0.528\textwidth}
        \centering
        \includegraphics[width=.9\linewidth]{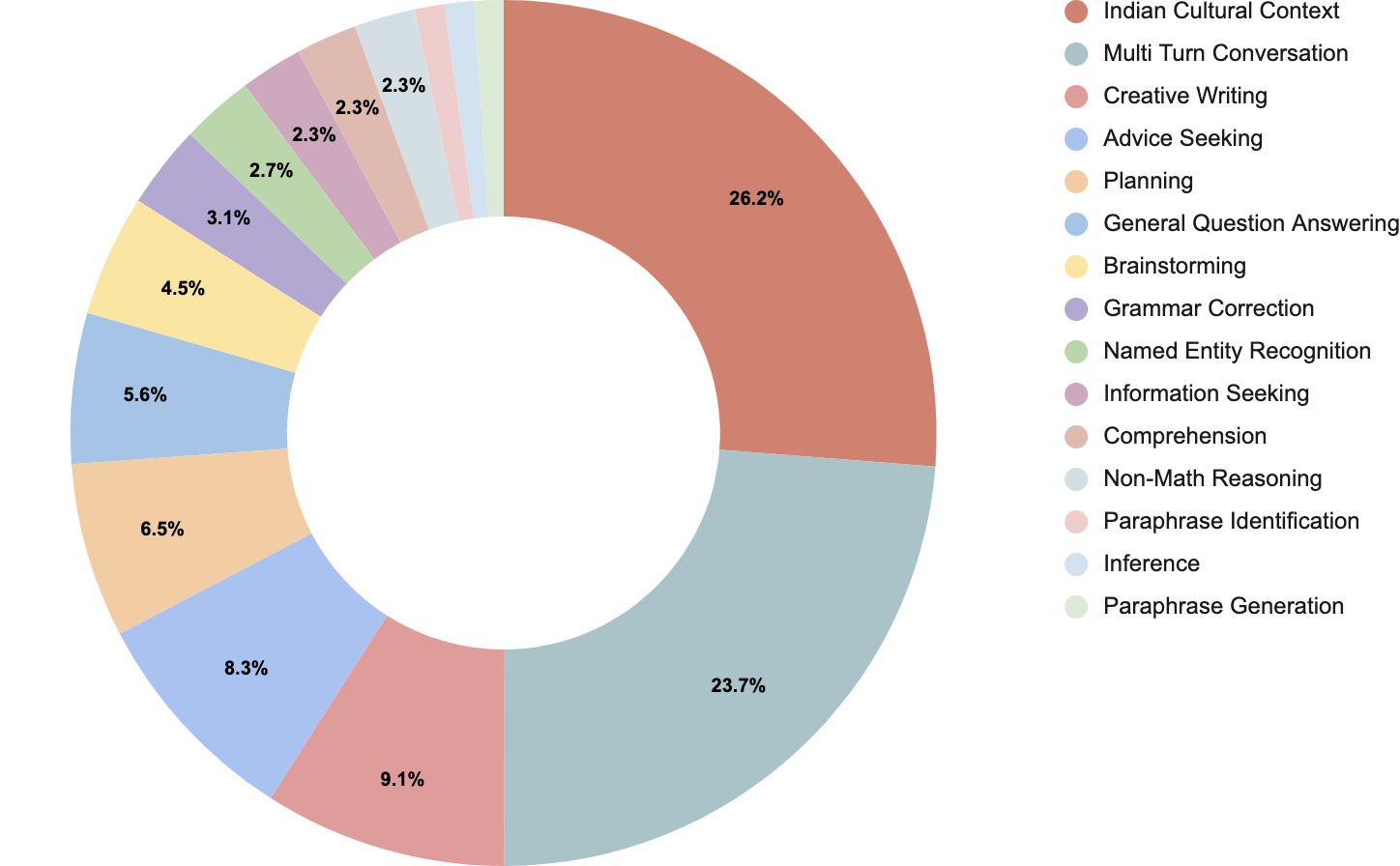}
    \end{minipage}
    \caption{Distribution of \textit{Pragyaan-IT} (Instruction-Tuning) data across languages (left) and categories (right).}
    \label{fig:sft-lang-cat-dist}
\end{figure*}


\subsection{Human-In-The-Loop (HITL) Refinement}

While synthetic generation and automated translation provides the backbone of our dataset creation pipeline, human annotators play an equally central role in shaping its final form. Each prompt–response pair, once generated and assigned a configuration of settings (e.g., \textit{easy, 1-Turn, Simple-IF, Safe, IC-3, Normal (No Thinking Trails)}), enters a stage of manual intervention where annotators act not merely as reviewers, but as curators of the data.
If a pair does not fully align with its designated configuration, annotators may either (i) adapt the configuration to better reflect the pair, or (ii) create a new prompt–response pair that correctly conforms to the specified configuration. This decision balances efficiency with fidelity to the framework.
In cases where manually generating a new response is especially time-intensive, the annotators flag the prompt for regeneration and they undergo another iteration through the pipeline.

Crucially, manual intervention goes beyond mechanical verification. Annotators conduct linguistic quality checks ensuring fluency, grammatical accuracy, syntactic correctness, and appropriate response length, but are also encouraged to exercise creative judgment. This includes refining awkward phrasings, restructuring unclear outputs, or enriching responses with culturally relevant details. Even when a pair formally satisfies all defined constraints, annotators may modify it to adjust tone, improve readability, contextual appropriateness or pedagogical value as well as elevate the overall communicative value of the response. These interventions ensure that the dataset is not only consistent with the defined settings, but also meaningful and robust for deployment in real-world post-training scenarios.


\subsection{Dataset Curation}
\label{subsec:dataset}
Our curation process draws from a broad pool of existing resources while systematically adapting them for Indic languages and tasks. In total, we considered $57$ diverse language
datasets, together with their relevant splits, spanning different categories and task families. These resources serve either as direct candidates for translation into Indic languages or as seed data for synthetic expansion through a modified Self-Instruct pipeline \citep{selfinstruct} described earlier. By combining translation and generation in a complementary manner, we ensure that the curated data covers not only core Natural Language Processing (NLP) tasks but also culturally grounded and contextually relevant dimensions.
Table~\ref{tab:datasets} in the Appendix provides an overview of the corresponding candidate datasets associated with each task sub-category.


\begin{table}[htbp]
\centering
\resizebox{0.85\columnwidth}{!}{%
\begin{tabular}{llr}
\hline
\textbf{Setting} & \textbf{Configuration} & \textbf{\%} \\
\hline
\multirow{2}{*}{Complexity} & Easy & 62.32 \\
 & Hard & 37.68 \\
\hline
\multirow{3}{*}{Multi Turn} & 1-Turn & 91.66 \\
 & 3-Turn & 6.76 \\
 & 5-Turn & 1.58 \\
\hline
\multirow{3}{*}{Instruction Following} & Simple IF & 96.95 \\
 & Medium IF & 2.49 \\
 & Complex IF & 0.56 \\
\hline
\multirow{2}{*}{Safety} & Safe & 92.51 \\
 & Non-Safe & 7.49 \\
\hline
\multirow{3}{*}{Indian Context} & IC-1 & 32.04 \\
 & IC-2 & 10.16 \\
 & IC-3 & 57.80 \\
\hline
\multirow{3}{*}{Thinking Trails} & Normal & 99.98 \\
 & CoT & 0.01 \\
 & Self Thinking & 0.01 \\
\hline
\end{tabular}
}
\caption{Distribution of instances in \textit{Pragyaan-IT} across different task settings and configurations in Section \ref{subsec:settings}.}
\label{tab:sft-set-dist}
\end{table}

\begin{figure*}[htbp]
    \centering
    \begin{minipage}[t]{0.472\textwidth}
        \centering
        \includegraphics[width=\linewidth]{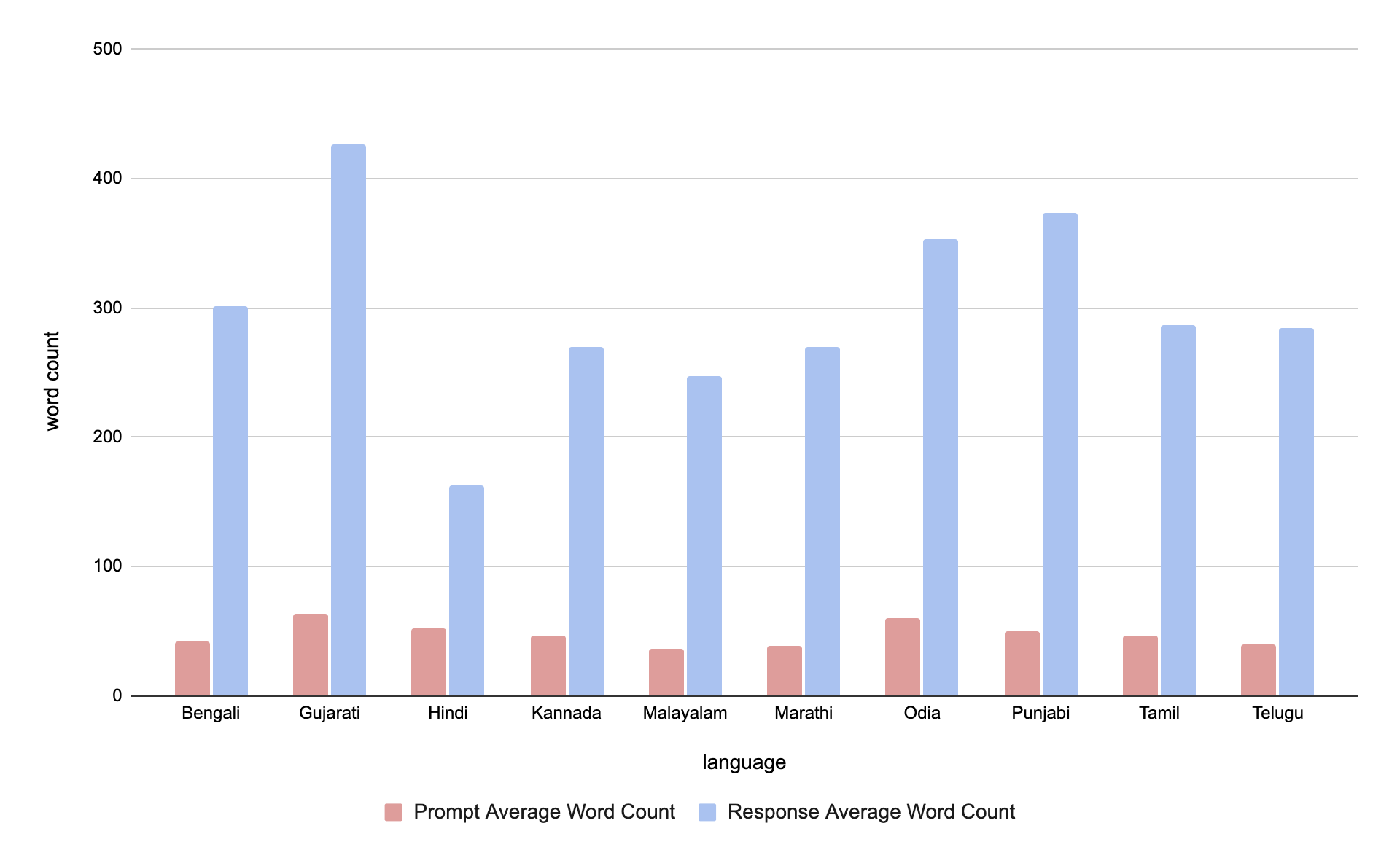}
    \end{minipage}%
    \hfill
    \begin{minipage}[t]{0.528\textwidth}
        \centering
        \includegraphics[width=.9\linewidth]{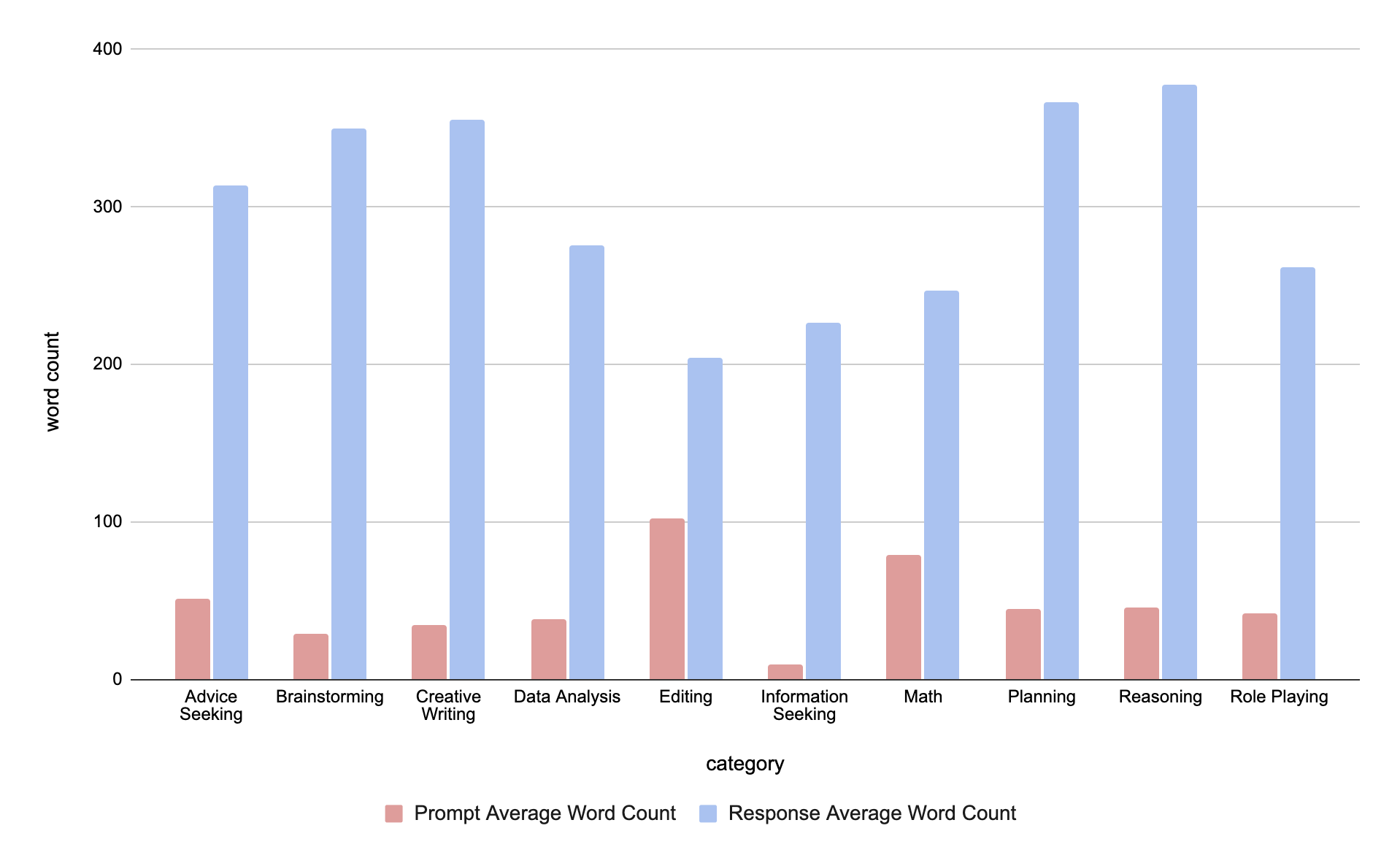}
    \end{minipage}
    \caption{Average word counts of \textit{Pragyaan-Align} alignment data across languages 
    (left) and categories (right).}
    \label{fig:dpo-lang-cat-dist}
\end{figure*}

\section{Pragyaan: Indic Post-training Datasets}

As part of this work, we construct high-quality post-training datasets that explicitly target 10 Indian languages (Bengali, Gujarati, Hindi, Kannada, Malayalam, Marathi, Oriya, Punjabi, Tamil, and Telugu),
yielding two complementary resources: 

i) \textit{Pragyaan-IT} (22.5K): an instruction-tuning dataset designed to enhance a model’s ability to follow diverse prompts across multiple domains, ensuring that models can generalize well to everyday user interactions.

ii) \textit{Pragyaan-Align} (100K): 
a preference dataset curated for Reinforcement Learning (RL)-based alignment methods,  
emphasizing preference learning, safety, and cultural grounding, allowing models 
to align more closely with the user intent.





\section{Analysis}
We begin with an assessment of raw synthetic generations refined through human annotation, followed by a broader dataset-level analysis.

\subsection{Human Annotation Refinement}

We evaluate the underlying LLM's performance for both generation and translation tasks across 5 dimensions, each on a scale of 1-5, for a small subset (see Section \ref{appendix:human} in the Appendix for details on evaluation).
Particularly, for English generations, grammatical accuracy remains slightly lower in comprehension (3.20) and creative writing (3.93) tasks, while receiving high scores for other tasks. For translation, the model shows the strongest performance for Hindi (Figure~\ref{fig:translation_evaluation}). Telugu and Gujarati exhibit moderate Lexical Diversity (3.43 and 3.30), while grammatical accuracy remains modest in Telugu (3.62), Hindi (3.60), and Punjabi (3.55). Thus, human refinements were critical for converting raw synthetic output into culturally grounded, linguistically accurate, and task-aligned data, directly underpinning the reliability of our data curation framework. 




\begin{figure*}[htbp]
    \centering
    \begin{minipage}[t]{0.472\textwidth}
        \centering
        \includegraphics[width=\linewidth]{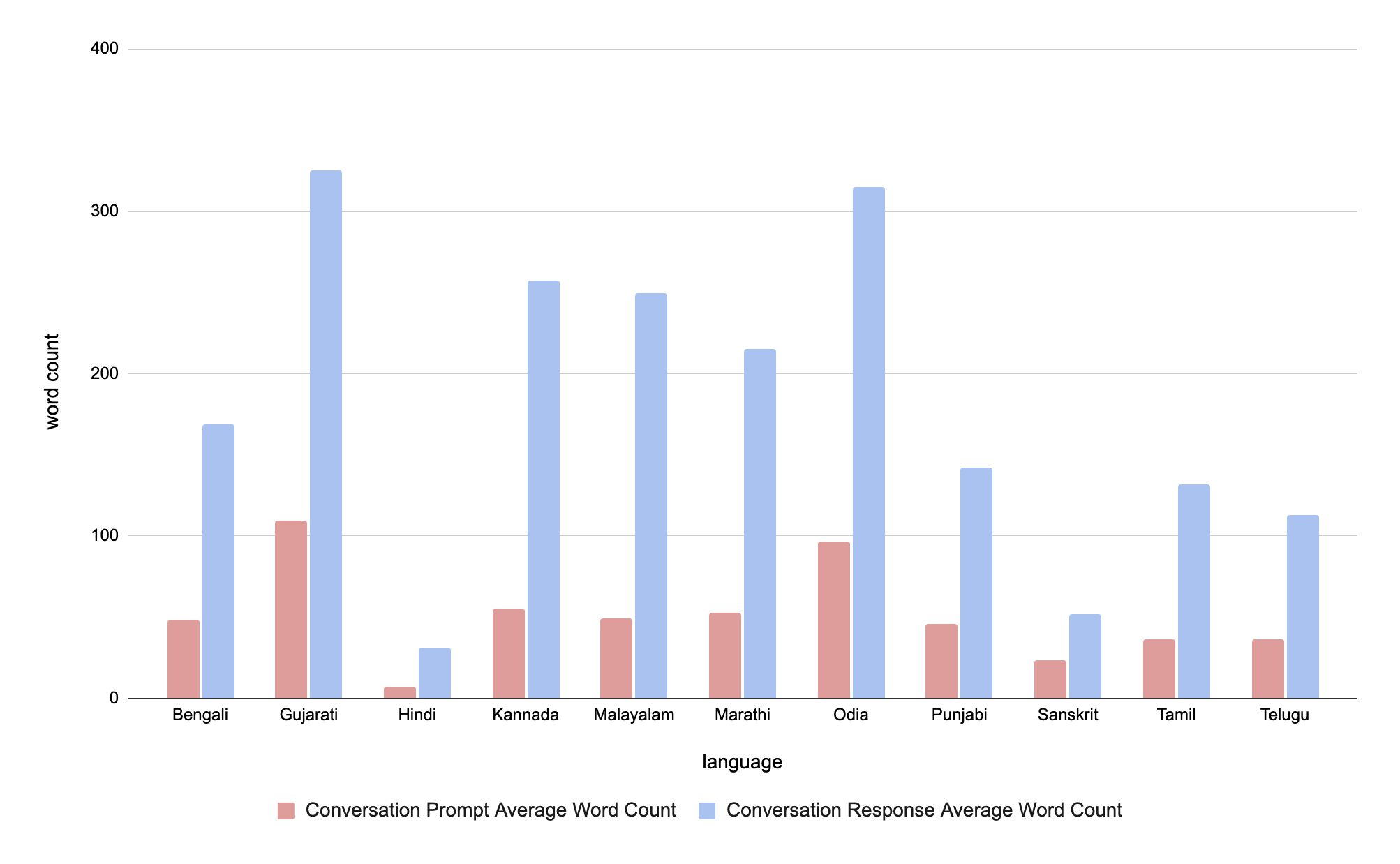}
    \end{minipage}%
    \hfill
    \begin{minipage}[t]{0.528\textwidth}
        \centering
        \includegraphics[width=.9\linewidth]{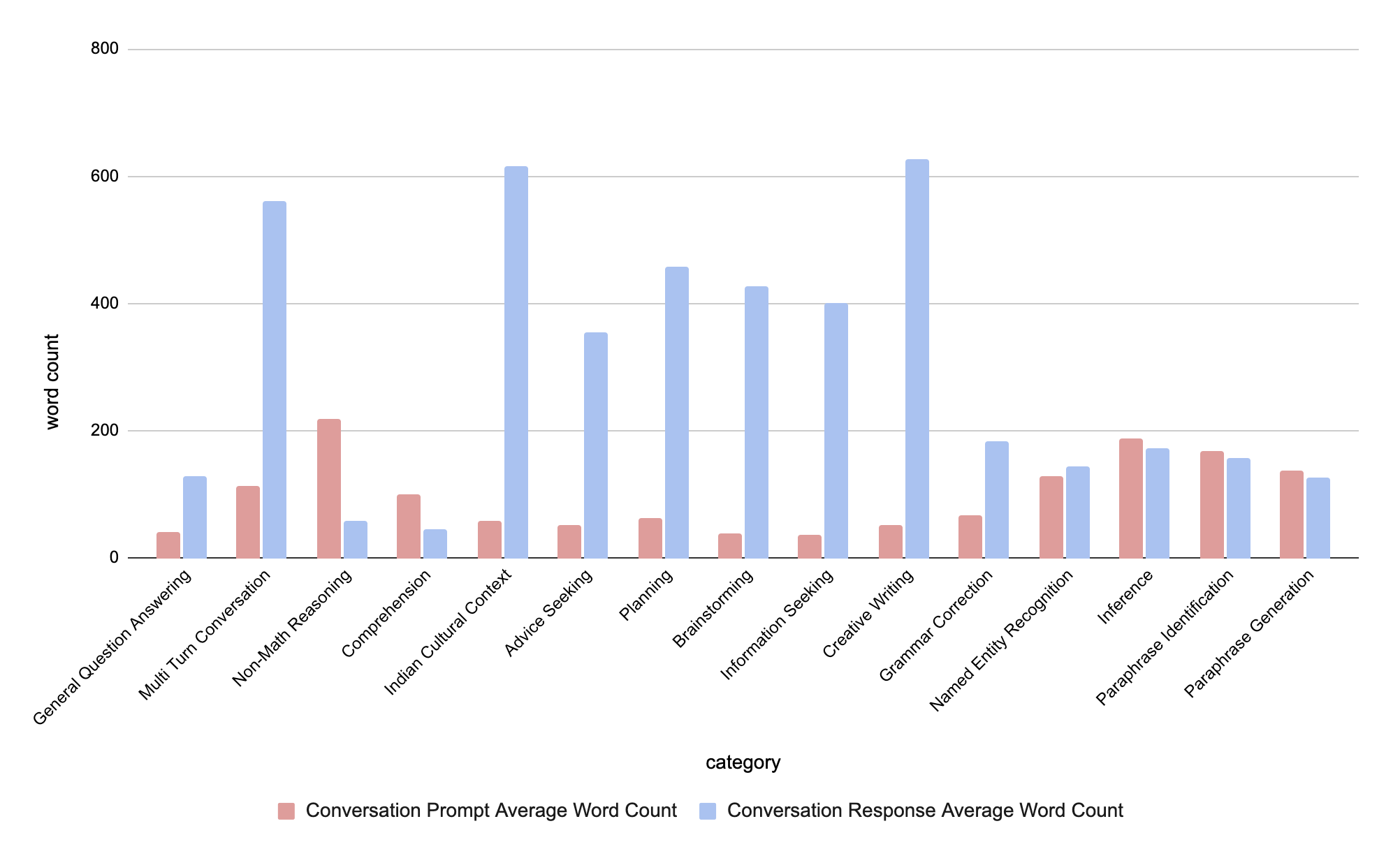}
    \end{minipage}
    \caption{Average word counts of \textit{Pragyaan-IT} data across languages 
    (left) and categories (right).}
    \label{fig:sft-lang-cat-word-counts}
\end{figure*}

\begin{figure*}[htbp]
    \centering
    \begin{minipage}[t]{0.472\textwidth}
        \centering
        \includegraphics[width=\linewidth]{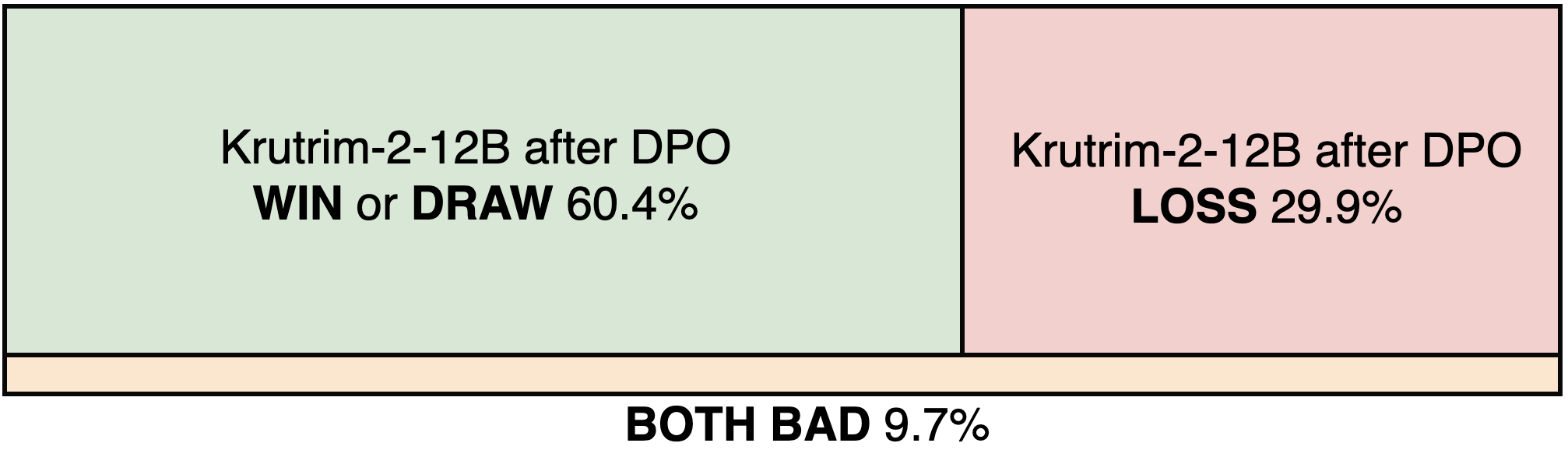}
    \end{minipage}%
    \hfill
    \begin{minipage}[t]{0.528\textwidth}
        \centering
        \includegraphics[width=.9\linewidth]{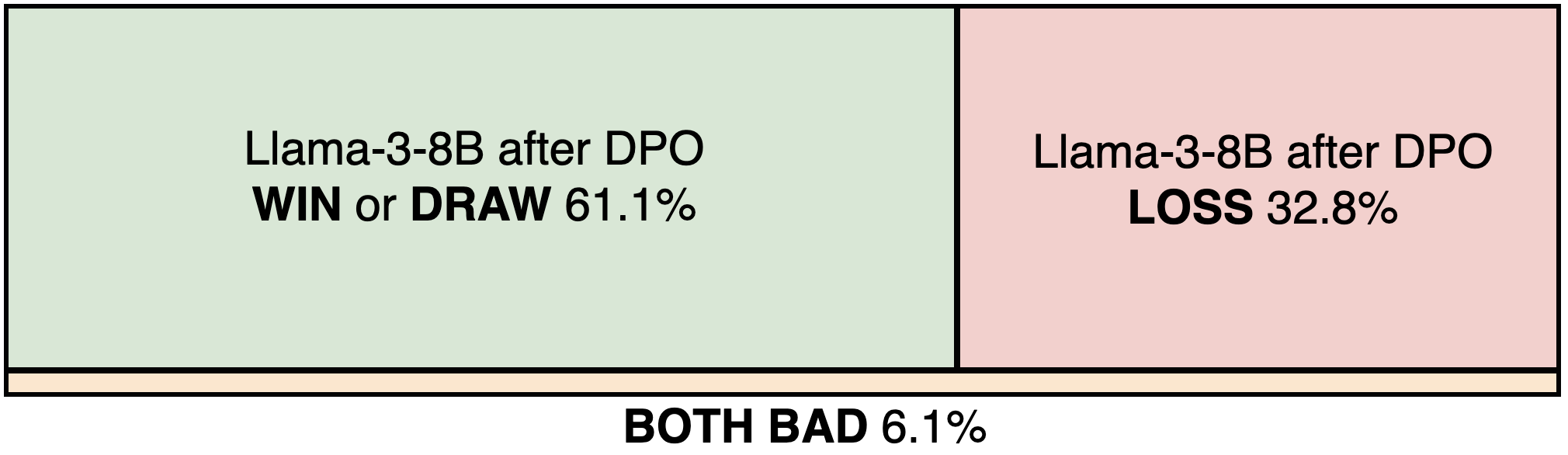}
    \end{minipage}
    \caption{Win rates of the \textit{Krutrim-2-12B} (left) and \textit{Llama-3-8B} (right) models after DPO, compared against their respective pre-DPO versions.}
    \label{fig:win-rates}
\end{figure*}

\subsection{Dataset analysis}

Figure~\ref{fig:sft-lang-cat-dist} shows the distribution of \textit{Pragyaan-IT} across 10 languages and 15 categories, while Table~\ref{tab:datasets} lists the candidate datasets used in its construction. As seen in Figure~\ref{fig:sft-lang-cat-dist}, Indian Cultural Context (26.2\%) and Multi-Turn Conversation (23.7\%) dominate, while reasoning and paraphrasing remain limited (1–3\%), forming targets for future expansion. Language coverage is led by Gujarati (17.7\%), Kannada (11.4\%), Marathi (11.1\%), and Odia (10.8\%). In the current setting (Table~\ref{tab:sft-set-dist}), we have 62.3\% `Easy' tasks, 
single-turn interactions (91.7\%), simple instruction-following (96.9\%), safe content (92.5\%) with Indian context well covered (IC-3: 57.8\%). While this analysis reflects the status of the \textit{Pragyaan-IT} dataset at the time of writing, the dataset is under active curation and will be more comprehensive across task categories and settings as described in earlier sections.

Word count analysis highlights linguistic and task-level variation (Figure~\ref{fig:sft-lang-cat-word-counts}). Gujarati and Odia are most verbose in both prompts ($\sim$110 words) and responses ($\sim$320 words), whereas Hindi mostly
remain concise. At the task level, Multi-turn conversation, Advice Seeking, and Creative Writing yield the longest responses (550–620 words).
Conversely, QA, Indian cultural context, and Comprehension tasks are consistently brief. Verbosity thus correlates with conversational and creative tasks, while simpler or context-specific settings produce shorter outputs.

\textit{Pragyaan-Align}, the preference dataset, 
has an equal representation across all languages and 10 categories that helps promote fairness and mitigate bias. All instances in \textit{Pragyaan-Align} follow a standardized configuration: single-turn interactions with instruction following and safe responses. 
Our analysis shows variation in text lengths across categories and languages. Average word counts for prompts range from 36–63 words, preferred responses 137–539, while rejected responses range from 154–466 words reflecting differences in task complexity and elaboration.
Detailed language-wise as well as category-wise trends are also presented in Figure \ref{fig:dpo-lang-cat-dist}.

Overall, our \textit{Pragyaan} datasets for post-training provide a broad task and Indian language coverage. The various task settings and our curation process leveraging LLMs along with human-in-the-loop help ensure high quality. 
Our future iterations will expand complex reasoning capabilities and enhance representation for under-covered low-resource languages. 

\subsection{Downstream performance}

To evaluate the quality of our curated dataset, we conduct a pilot study with Direct Preference Optimization (DPO) \cite{rafailov2023direct} based alignment on the \textit{Pragyaan-Align} dataset using two open-weight models which supports the languages under consideration: \textit{Krutrim-2-12B Instruct} \cite{kallappa2025krutrim} and \textit{Llama-3-8B Instruct} \cite{grattafiori2024llama}. 

For evaluation, we use the recently released \textit{Updesh} dataset\footnote{\url{https://huggingface.co/datasets/microsoft/Updesh_beta}}, which covers similar categories and languages. We sample around 100 examples from nine relevant categories, balanced across ten languages. Responses from the models
are scored against the ground truth on a scale of 1–5 using an LLM-as-a-judge. We provide more information about the hyperparameter configuration and other  experimental details in Section \ref{appendix:implementation} and the corresponding prompts used for evaluation in Section \ref{appendix:prompts} of the Appendix. 

As shown in Figure~\ref{fig:win-rates}, \textit{Krutrim-2-12B} after DPO either wins or draws in 60.4\% of cases, loses in 29.9\%, and both pre- and post-DPO responses score poorly (<2) in 9.7\%. A similar trend holds for \textit{Llama-3-8B} (61.1\%  wins or draws), confirming the promising potential of our curated dataset for alignment across different categories and multiple Indian languages.

\section{Conclusion}

This work addresses the scarcity of high-quality post-training data for multilingual LLMs by developing a human-in-the-loop pipeline to ensure diversity, quality and cultural grounding. 
Through this approach, we construct two datasets: \textit{Pragyaan-IT} and \textit{Pragyaan-Align} covering 10 Indian languages and multiple task categories. The datasets highlight inclusion of local cultural context, task diversity, multi-turn dialogue, and safety alignment, overcoming the limitations of naive translations and low-quality synthetic resources. Although designed for Indian languages, the pipeline is readily adaptable to other multilingual contexts. We present a comprehensive analysis of the dataset’s characteristics, covering language distribution and domain coverage, and further demonstrate its effectiveness for alignment through a small-scale pilot study.
Future efforts will expand language coverage and further annotation quality refinement. We aim for this work to support broader efforts in building 
culturally inclusive resources that strengthen LLM applicability in multilingual contexts.


\section*{Limitations}

Our dataset is part of an ongoing effort, with plans to continually expand post-training instances.
While our human-in-the-loop framework mitigates many issues, challenges such as minor linguistic inaccuracies, fluency variation across languages, and potential annotation subjectivity may still persist. Moreover, our research prioritized
Indian languages, and the generalization of findings to other multilingual settings remain currently unexplored. Extending the pipeline to new cultural and linguistic contexts will require additional validation. We view these limitations as avenues for future work towards broader applicability and refinement of our proposed framework.



\section*{Ethics Statement}

This study focuses on curating large-scale post-training datasets for Indian languages, encompassing diverse tasks and cultural contexts. The pipeline combines synthetic generation with human-in-the-loop refinement to ensure quality, safety, and cultural fidelity. 
We provide proper attribution to all source datasets and tools through citations. 
Human involvement was limited to annotation and quality control; no personally identifiable or sensitive information was collected. We engage a team of 50 in-house annotators for dataset creation. All contributors were clearly informed that their work supports LLM training and were compensated fairly at locally prevailing market rates. This study did not require formal IRB approval. Throughout the process, we prioritized preserving cultural nuances while avoiding harmful, biased, or unsafe content. The resulting dataset is designed to advance the development of multilingual and culturally inclusive LLMs.




\section*{Acknowledgements}


We thank the leadership at Krutrim for their support in carrying out this research. We also thank the Data Annotation Team for their meticulous efforts especially Sanmathi P N, Dr. Salman Alam, Rupkatha Mukherjee, Vinod Kumar, Shivaram Prasad Putta, Divyashree K, Arati V Thakur, Vaibhav M Sutariya, Mitali Pradhan, Payal Yadav, Sneha Aniyan, Hemant P Rajopadhye, Joel Johnson and Srinidhi S. Additionally, we also thank Guduru Manoj and Souvik Rana for the helpful discussions.

\bibliography{anthology,custom}

\begin{thebibliography}{88}
\expandafter\ifx\csname natexlab\endcsname\relax\def\natexlab#1{#1}\fi

\bibitem[{Abdin et~al.(2025)Abdin, Agarwal, Awadallah, Balachandran, Behl, Chen, de~Rosa, Gunasekar, Javaheripi, Joshi et~al.}]{abdin2025phi}
Marah Abdin, Sahaj Agarwal, Ahmed Awadallah, Vidhisha Balachandran, Harkirat Behl, Lingjiao Chen, Gustavo de~Rosa, Suriya Gunasekar, Mojan Javaheripi, Neel Joshi, et~al. 2025.
\newblock Phi-4-reasoning technical report.
\newblock \emph{arXiv preprint arXiv:2504.21318}.

\bibitem[{Bai et~al.(2022{\natexlab{a}})Bai, Jones, Ndousse, Askell, Chen, DasSarma, Drain, Fort, Ganguli, Henighan, Joseph, Kadavath, Kernion, Conerly, El-Showk, Elhage, Hatfield-Dodds, Hernandez, Hume, Johnston, Kravec, Lovitt, Nanda, Olsson, Amodei, Brown, Clark, McCandlish, Olah, Mann, and Kaplan}]{bai2022training}
Yuntao Bai, Andy Jones, Kamal Ndousse, Amanda Askell, Anna Chen, Nova DasSarma, Dawn Drain, Stanislav Fort, Deep Ganguli, Tom Henighan, Nicholas Joseph, Saurav Kadavath, Jackson Kernion, Tom Conerly, Sheer El-Showk, Nelson Elhage, Zac Hatfield-Dodds, Danny Hernandez, Tristan Hume, Scott Johnston, Shauna Kravec, Liane Lovitt, Neel Nanda, Catherine Olsson, Dario Amodei, Tom Brown, Jack Clark, Sam McCandlish, Chris Olah, Ben Mann, and Jared Kaplan. 2022{\natexlab{a}}.
\newblock \href {http://arxiv.org/abs/2204.05862} {Training a helpful and harmless assistant with reinforcement learning from human feedback}.

\bibitem[{Bai et~al.(2022{\natexlab{b}})Bai, Kadavath, Kundu, Askell et~al.}]{bai2022constitutional}
Yuntao Bai, Saurav Kadavath, Sandhini Kundu, Amanda Askell, et~al. 2022{\natexlab{b}}.
\newblock Constitutional ai: Harmlessness from ai feedback.
\newblock \emph{arXiv preprint arXiv:2212.08073}.

\bibitem[{Barham et~al.(2023)Barham, , Weller, Yuan, Murray, Yarmohammadi, Jiang, Vashishtha, Martin, Liu, White, Boyd-Graber, and Durme}]{barham2023megawika}
Samuel Barham, , Weller, Michelle Yuan, Kenton Murray, Mahsa Yarmohammadi, Zhengping Jiang, Siddharth Vashishtha, Alexander Martin, Anqi Liu, Aaron~Steven White, Jordan Boyd-Graber, and Benjamin~Van Durme. 2023.
\newblock \href {http://arxiv.org/abs/2307.07049} {Megawika: Millions of reports and their sources across 50 diverse languages}.

\bibitem[{Bercovich et~al.(2025{\natexlab{a}})Bercovich, Levy, Golan, Dabbah, El-Yaniv, Puny, Galil, Moshe, Ronen, Nabwani et~al.}]{bercovich2025llamanemotronefficientreasoningmodels}
Akhiad Bercovich, Itay Levy, Izik Golan, Mohammad Dabbah, Ran El-Yaniv, Omri Puny, Ido Galil, Zach Moshe, Tomer Ronen, Najeeb Nabwani, et~al. 2025{\natexlab{a}}.
\newblock Llama-nemotron: Efficient reasoning models.
\newblock \emph{arXiv preprint arXiv:2505.00949}.

\bibitem[{Bercovich et~al.(2025{\natexlab{b}})Bercovich, Levy, Golan, Dabbah, El-Yaniv, Puny, Galil, Moshe, Ronen, Nabwani et~al.}]{bercovich2025llama}
Akhiad Bercovich, Itay Levy, Izik Golan, Mohammad Dabbah, Ran El-Yaniv, Omri Puny, Ido Galil, Zach Moshe, Tomer Ronen, Najeeb Nabwani, et~al. 2025{\natexlab{b}}.
\newblock Llama-nemotron: Efficient reasoning models.
\newblock \emph{arXiv preprint arXiv:2505.00949}.

\bibitem[{Bian et~al.(2023)Bian, Lin, Lu, Han, Sun, and He}]{ChatAlpaca}
Ning Bian, Hongyu Lin, Yaojie Lu, Xianpei Han, Le~Sun, and Ben He. 2023.
\newblock Chatalpaca: A multi-turn dialogue corpus based on alpaca instructions.
\newblock \url{https://github.com/cascip/ChatAlpaca}.

\bibitem[{Borkan et~al.(2019)Borkan, Dixon, Sorensen, Thain, and Vasserman}]{borkan2019nuanced}
Daniel Borkan, Lucas Dixon, Jeffrey Sorensen, Nithum Thain, and Lucy Vasserman. 2019.
\newblock Nuanced metrics for measuring unintended bias with real data for text classification.
\newblock In \emph{Companion Proceedings of The 2019 World Wide Web Conference}.

\bibitem[{Chung et~al.(2022)Chung, Hou, Longpre, Zoph et~al.}]{chung2022scaling}
Hyung~Won Chung, Le~Hou, Shayne Longpre, Barret Zoph, et~al. 2022.
\newblock Scaling instruction-finetuned language models.
\newblock \emph{arXiv preprint arXiv:2210.11416}.

\bibitem[{Cobbe et~al.(2021)Cobbe, Kosaraju, Bavarian, Chen, Jun, Kaiser, Plappert, Tworek, Hilton, Nakano, Hesse, and Schulman}]{cobbe2021gsm8k}
Karl Cobbe, Vineet Kosaraju, Mohammad Bavarian, Mark Chen, Heewoo Jun, Lukasz Kaiser, Matthias Plappert, Jerry Tworek, Jacob Hilton, Reiichiro Nakano, Christopher Hesse, and John Schulman. 2021.
\newblock Training verifiers to solve math word problems.
\newblock \emph{arXiv preprint arXiv:2110.14168}.

\bibitem[{Cohan et~al.(2018)Cohan, Dernoncourt, Kim, Bui, Kim, Chang, and Goharian}]{cohan-etal-2018-discourse}
Arman Cohan, Franck Dernoncourt, Doo~Soon Kim, Trung Bui, Seokhwan Kim, Walter Chang, and Nazli Goharian. 2018.
\newblock A discourse-aware attention model for abstractive summarization of long documents.
\newblock In \emph{Proceedings of the 2018 Conference of the North American Chapter of the Association for Computational Linguistics: Human Language Technologies, Volume 2 (Short Papers)}.

\bibitem[{Conneau et~al.(2018)Conneau, Rinott, Lample, Williams, Bowman, Schwenk, and Stoyanov}]{conneau2018xnli}
Alexis Conneau, Ruty Rinott, Guillaume Lample, Adina Williams, Samuel~R. Bowman, Holger Schwenk, and Veselin Stoyanov. 2018.
\newblock Xnli: Evaluating cross-lingual sentence representations.
\newblock In \emph{Proceedings of the 2018 Conference on Empirical Methods in Natural Language Processing}. Association for Computational Linguistics.

\bibitem[{Conover et~al.(2023)Conover, Hayes, Mathur, Xie, Wan, Shah, Ghodsi, Wendell, Zaharia, and Xin}]{DatabricksBlog2023DollyV2}
Mike Conover, Matt Hayes, Ankit Mathur, Jianwei Xie, Jun Wan, Sam Shah, Ali Ghodsi, Patrick Wendell, Matei Zaharia, and Reynold Xin. 2023.
\newblock \href {https://www.databricks.com/blog/2023/04/12/dolly-first-open-commercially-viable-instruction-tuned-llm} {{Free Dolly: Introducing the World's First Truly Open Instruction-Tuned LLM}}.

\bibitem[{Cui et~al.(2023)Cui, Yuan, Ding, Yao, He, Zhu, Ni, Xie, Xie, Lin et~al.}]{cui2023ultrafeedback}
Ganqu Cui, Lifan Yuan, Ning Ding, Guanming Yao, Bingxiang He, Wei Zhu, Yuan Ni, Guotong Xie, Ruobing Xie, Yankai Lin, et~al. 2023.
\newblock Ultrafeedback: Boosting language models with scaled ai feedback.
\newblock \emph{arXiv preprint arXiv:2310.01377}.

\bibitem[{Doddapaneni et~al.(2023)Doddapaneni, Aralikatte, Ramesh, Goyal, Khapra, Kunchukuttan, and Kumar}]{doddapaneni-etal-2023-towards}
Sumanth Doddapaneni, Rahul Aralikatte, Gowtham Ramesh, Shreya Goyal, Mitesh~M. Khapra, Anoop Kunchukuttan, and Pratyush Kumar. 2023.
\newblock \href {https://aclanthology.org/2023.acl-long.693} {Towards leaving no {I}ndic language behind: Building monolingual corpora, benchmark and models for {I}ndic languages}.
\newblock In \emph{Proceedings of the 61st Annual Meeting of the Association for Computational Linguistics (Volume 1: Long Papers)}, pages 12402--12426, Toronto, Canada. Association for Computational Linguistics.

\bibitem[{Grattafiori et~al.(2024)Grattafiori, Dubey, Jauhri, Pandey, Kadian, Al-Dahle, Letman, Mathur, Schelten, Vaughan et~al.}]{grattafiori2024llama}
Aaron Grattafiori, Abhimanyu Dubey, Abhinav Jauhri, Abhinav Pandey, Abhishek Kadian, Ahmad Al-Dahle, Aiesha Letman, Akhil Mathur, Alan Schelten, Alex Vaughan, et~al. 2024.
\newblock The llama 3 herd of models.
\newblock \emph{arXiv preprint arXiv:2407.21783}.

\bibitem[{Gunasekar et~al.(2023)Gunasekar, Zhang, Aneja, Mendes, Giorno, Gopi, Javaheripi, Kauffmann, de~Rosa, Saarikivi, Salim, Shah, Behl, Wang, Bubeck, Eldan, Kalai, Lee, and Li}]{gunasekar2023textbooks}
Suriya Gunasekar, Yi~Zhang, Jyoti Aneja, Caio César~Teodoro Mendes, Allie~Del Giorno, Sivakanth Gopi, Mojan Javaheripi, Piero Kauffmann, Gustavo de~Rosa, Olli Saarikivi, Adil Salim, Shital Shah, Harkirat~Singh Behl, Xin Wang, Sébastien Bubeck, Ronen Eldan, Adam~Tauman Kalai, Yin~Tat Lee, and Yuanzhi Li. 2023.
\newblock \href {http://arxiv.org/abs/2306.11644} {Textbooks are all you need}.

\bibitem[{Guo et~al.(2025)Guo, Yang, Zhang, Song, Zhang, Xu, Zhu, Ma, Wang, Bi et~al.}]{guo2025deepseek}
Daya Guo, Dejian Yang, Haowei Zhang, Junxiao Song, Ruoyu Zhang, Runxin Xu, Qihao Zhu, Shirong Ma, Peiyi Wang, Xiao Bi, et~al. 2025.
\newblock Deepseek-r1: Incentivizing reasoning capability in llms via reinforcement learning.
\newblock \emph{arXiv preprint arXiv:2501.12948}.

\bibitem[{Gupta et~al.(2023)Gupta, Scaria, Anantheswaran, Verma, Parmar, Sawant, Baral, and Mishra}]{gupta2023targen}
Himanshu Gupta, Kevin Scaria, Ujjwala Anantheswaran, Shreyas Verma, Mihir Parmar, Saurabh~Arjun Sawant, Chitta Baral, and Swaroop Mishra. 2023.
\newblock \href {http://arxiv.org/abs/2310.17876} {Targen: Targeted data generation with large language models}.

\bibitem[{Hartung et~al.(2023)Hartung, Herygers, Kurlekar, Zakaria, Volkan, Gröttrup, and Georges}]{hartung2023measuring}
Kai Hartung, Aaricia Herygers, Shubham Kurlekar, Khabbab Zakaria, Taylan Volkan, Sören Gröttrup, and Munir Georges. 2023.
\newblock \href {http://arxiv.org/abs/2306.07152} {Measuring sentiment bias in machine translation}.

\bibitem[{Hendrycks et~al.(2020)Hendrycks, Burns, Basart, Zou, Mazeika, Song, and Steinhardt}]{hendrycks2020measuring}
Dan Hendrycks, Collin Burns, Steven Basart, Andy Zou, Mantas Mazeika, Dawn Song, and Jacob Steinhardt. 2020.
\newblock Measuring massive multitask language understanding.
\newblock \emph{arXiv preprint arXiv:2009.03300}.

\bibitem[{Hendrycks et~al.(2021)Hendrycks, Burns, Kadavath, Arora, Basart, Tang, Song, and Steinhardt}]{hendrycksmath2021}
Dan Hendrycks, Collin Burns, Saurav Kadavath, Akul Arora, Steven Basart, Eric Tang, Dawn Song, and Jacob Steinhardt. 2021.
\newblock Measuring mathematical problem solving with the math dataset.
\newblock \emph{NeurIPS}.

\bibitem[{Hinton et~al.(2015)Hinton, Vinyals, and Dean}]{hinton2015distilling}
Geoffrey Hinton, Oriol Vinyals, and Jeff Dean. 2015.
\newblock Distilling the knowledge in a neural network.
\newblock \emph{arXiv preprint arXiv:1503.02531}.

\bibitem[{Honovich et~al.(2022)Honovich, Scialom, Levy, and Schick}]{honovich2022unnatural}
Or~Honovich, Thomas Scialom, Omer Levy, and Timo Schick. 2022.
\newblock \href {https://arxiv.org/abs/2212.09689} {Unnatural instructions: Tuning language models with (almost) no human labor}.
\newblock \emph{arXiv preprint arXiv:2212.09689}.

\bibitem[{{Joseph Cheung}(2023)}]{joseph_cheung_2023}
{Joseph Cheung}. 2023.
\newblock \href {https://doi.org/10.57967/hf/0570} {{ GuanacoDataset (Revision 8cf0d29) }}.

\bibitem[{Joshi et~al.(2017)Joshi, Choi, Weld, and Zettlemoyer}]{joshi2017triviaqa}
Mandar Joshi, Eunsol Choi, Daniel~S Weld, and Luke Zettlemoyer. 2017.
\newblock Triviaqa: A large scale distantly supervised challenge dataset for reading comprehension.
\newblock \emph{arXiv preprint arXiv:1705.03551}.

\bibitem[{Joshi et~al.(2020)Joshi, Santy, Budhiraja, Bali, and Choudhury}]{joshi2020state}
Pratik Joshi, Sebastin Santy, Amar Budhiraja, Kalika Bali, and Monojit Choudhury. 2020.
\newblock The state and fate of linguistic diversity and inclusion in the nlp world.
\newblock \emph{arXiv preprint arXiv:2004.09095}.

\bibitem[{Kallappa et~al.(2025)Kallappa, Kamble, Ravi, Patidar, Dhruv, Kumar, Awasthi, Manjunath, Agarwal, Ashish, Bhargava, and Khatri}]{kallappa2025krutrim}
Aditya Kallappa, Palash Kamble, Abhinav Ravi, Akshat Patidar, Vinayak Dhruv, Deepak Kumar, Raghav Awasthi, Arveti Manjunath, Shubham Agarwal, Kumar Ashish, Gautam Bhargava, and Chandra Khatri. 2025.
\newblock \href {http://arxiv.org/abs/2502.09642} {Krutrim llm: Multilingual foundational model for over a billion people}.

\bibitem[{Khan et~al.(2024)Khan, Mehta, Sankar, Kumaravelan, Doddapaneni, Jain, Kunchukuttan, Kumar, Dabre, Khapra et~al.}]{khan2024indicllmsuite}
Mohammed Safi Ur~Rahman Khan, Priyam Mehta, Ananth Sankar, Umashankar Kumaravelan, Sumanth Doddapaneni, Sparsh Jain, Anoop Kunchukuttan, Pratyush Kumar, Raj Dabre, Mitesh~M Khapra, et~al. 2024.
\newblock Indicllmsuite: A blueprint for creating pre-training and fine-tuning datasets for indian languages.
\newblock \emph{arXiv preprint arXiv:2403.06350}.

\bibitem[{Kim et~al.(2023)Kim, Joo, Kim, Jang, Ye, Shin, and Seo}]{kim2023cot}
Seungone Kim, Se~June Joo, Doyoung Kim, Joel Jang, Seonghyeon Ye, Jamin Shin, and Minjoon Seo. 2023.
\newblock The cot collection: Improving zero-shot and few-shot learning of language models via chain-of-thought fine-tuning.
\newblock \emph{arXiv preprint arXiv:2305.14045}.

\bibitem[{K{\"o}pf et~al.(2023)K{\"o}pf, Kilcher, von R{\"u}tte, Anagnostidis, Tam, Stevens, Barhoum, Duc, Stanley, Nagyfi et~al.}]{kopf2023openassistant}
Andreas K{\"o}pf, Yannic Kilcher, Dimitri von R{\"u}tte, Sotiris Anagnostidis, Zhi-Rui Tam, Keith Stevens, Abdullah Barhoum, Nguyen~Minh Duc, Oliver Stanley, Rich{\'a}rd Nagyfi, et~al. 2023.
\newblock Openassistant conversations--democratizing large language model alignment.
\newblock \emph{arXiv preprint arXiv:2304.07327}.

\bibitem[{Kumar et~al.(2022)Kumar, Shrotriya, Sahu, Mishra, Dabre, Puduppully, Kunchukuttan, Khapra, and Kumar}]{Kumar2022IndicNLGSM}
Aman Kumar, Himani Shrotriya, Prachi Sahu, Amogh Mishra, Raj Dabre, Ratish Puduppully, Anoop Kunchukuttan, Mitesh~M. Khapra, and Pratyush Kumar. 2022.
\newblock \href {https://doi.org/10.18653/v1/2022.emnlp-main.360} {{I}ndic{NLG} benchmark: Multilingual datasets for diverse {NLG} tasks in {I}ndic languages}.
\newblock In \emph{Proceedings of the 2022 Conference on Empirical Methods in Natural Language Processing}, pages 5363--5394, Abu Dhabi, United Arab Emirates. Association for Computational Linguistics.

\bibitem[{Kwiatkowski et~al.(2019)Kwiatkowski, Palomaki, Redfield, Collins, Parikh, Alberti, Epstein, Polosukhin, Kelcey, Devlin, Toutanova, Jones, Chang, Dai, Uszkoreit, Le, and Petrov}]{Kwiatkowski2019Nq}
Tom Kwiatkowski, Jennimaria Palomaki, Olivia Redfield, Michael Collins, Ankur Parikh, Chris Alberti, Danielle Epstein, Illia Polosukhin, Matthew Kelcey, Jacob Devlin, Kenton~Lee Toutanova, Llion Jones, Ming-Wei Chang, Andrew Dai, Jakob Uszkoreit, Quoc Le, and Slav Petrov. 2019.
\newblock Natural questions: a benchmark for question answering research.
\newblock \emph{TACL}.

\bibitem[{Lambert et~al.(2025)Lambert, Morrison, Pyatkin, Huang, Ivison, Brahman, Miranda, Liu, Dziri, Lyu, Gu, Malik, Graf, Hwang, Yang, Bras, Tafjord, Wilhelm, Soldaini, Smith, Wang, Dasigi, and Hajishirzi}]{lambert2024tulu3}
Nathan Lambert, Jacob Morrison, Valentina Pyatkin, Shengyi Huang, Hamish Ivison, Faeze Brahman, Lester James~V. Miranda, Alisa Liu, Nouha Dziri, Shane Lyu, Yuling Gu, Saumya Malik, Victoria Graf, Jena~D. Hwang, Jiangjiang Yang, Ronan~Le Bras, Oyvind Tafjord, Chris Wilhelm, Luca Soldaini, Noah~A. Smith, Yizhong Wang, Pradeep Dasigi, and Hannaneh Hajishirzi. 2025.
\newblock \href {http://arxiv.org/abs/2411.15124} {Tulu 3: Pushing frontiers in open language model post-training}.

\bibitem[{Li et~al.(2023{\natexlab{a}})Li, Koto, Wu, Aji, and Baldwin}]{li2023bactrian}
Haonan Li, Fajri Koto, Minghao Wu, Alham~Fikri Aji, and Timothy Baldwin. 2023{\natexlab{a}}.
\newblock Bactrian-x: A multilingual replicable instruction-following model with low-rank adaptation.
\newblock \emph{arXiv preprint arXiv:2305.15011}.

\bibitem[{Li et~al.(2023{\natexlab{b}})Li, Yin, Li, Chen, Wang, Ren, Li, Yang, Xu, Sun, Kong, and Liu}]{li2023m3it}
Lei Li, Yuwei Yin, Shicheng Li, Liang Chen, Peiyi Wang, Shuhuai Ren, Mukai Li, Yazheng Yang, Jingjing Xu, Xu~Sun, Lingpeng Kong, and Qi~Liu. 2023{\natexlab{b}}.
\newblock \href {http://arxiv.org/abs/2306.04387} {M$^3$it: A large-scale dataset towards multi-modal multilingual instruction tuning}.

\bibitem[{Liu et~al.(2024)Liu, Zhang, Sun, Huang, Wu, Li, Yu, Liu, Li, Guo, Li, Huang, Li, Zhou, Chen, Jia, and Jiang}]{liu2024mmul-pro}
Yang Liu, Xiaotian Zhang, Haoze Sun, Yuzhen Huang, Wei Wu, Yiliang Li, Hao Yu, Jun Liu, Yueqi Li, Zhicheng Guo, Xiaoguang Li, Yongfeng Huang, Guilin Li, Yujun Zhou, Yufeng Chen, Chenyan Jia, and Xing-Jian Jiang. 2024.
\newblock Mmlu-pro: a more advanced and challenging multi-task evaluation for llms.
\newblock \emph{arXiv preprint arXiv:2406.01574}.

\bibitem[{Longpre et~al.(2023)Longpre, Hou, Vu, Webson et~al.}]{longpre2023flan}
Shayne Longpre, Le~Hou, Tu~Vu, Albert Webson, et~al. 2023.
\newblock The flan collection: Designing data and methods for effective instruction tuning.
\newblock \emph{arXiv preprint arXiv:2301.13688}.

\bibitem[{Luo et~al.(2023)Luo, Xu, Zhao, Sun, Geng, Hu, Tao, Ma, Lin, and Jiang}]{luo2023wizardcoder}
Ziyang Luo, Can Xu, Pu~Zhao, Qingfeng Sun, Xiubo Geng, Wenxiang Hu, Chongyang Tao, Jing Ma, Qingwei Lin, and Daxin Jiang. 2023.
\newblock \href {http://arxiv.org/abs/2306.08568} {Wizardcoder: Empowering code large language models with evol-instruct}.

\bibitem[{Mhaske et~al.(2022)Mhaske, Kedia, Doddapaneni, Khapra, Kumar, Murthy, and Kunchukuttan}]{mhaske2022naamapadam}
Arnav Mhaske, Harshit Kedia, Sumanth Doddapaneni, Mitesh~M. Khapra, Pratyush Kumar, Rudra Murthy, and Anoop Kunchukuttan. 2022.
\newblock \href {https://doi.org/10.48550/ARXIV.2212.10168} {Naamapadam: A large-scale named entity annotated data for indic languages}.

\bibitem[{Mihaylov et~al.(2018)Mihaylov, Clark, Khot, and Sabharwal}]{mihaylov-etal-2018-openbookqa}
Todor Mihaylov, Peter Clark, Tushar Khot, and Ashish Sabharwal. 2018.
\newblock A new dataset for open book question answering.
\newblock In \emph{Proceedings of the 2018 Conference on Empirical Methods in Natural Language Processing}.

\bibitem[{Muennighoff et~al.(2023)Muennighoff, Liu, Zebaze, Zheng, Hui, Zhuo, Singh, Tang, von Werra, and Longpre}]{muennighoff2023octopack}
Niklas Muennighoff, Qian Liu, Armel Zebaze, Qinkai Zheng, Binyuan Hui, Terry~Yue Zhuo, Swayam Singh, Xiangru Tang, Leandro von Werra, and Shayne Longpre. 2023.
\newblock Octopack: Instruction tuning code large language models.
\newblock \emph{arXiv preprint arXiv:2308.07124}.

\bibitem[{Muennighoff et~al.(2022)Muennighoff, Wang, Sutawika, Roberts, Biderman, Scao, Bari, Shen, Yong, Schoelkopf et~al.}]{muennighoff2022crosslingual}
Niklas Muennighoff, Thomas Wang, Lintang Sutawika, Adam Roberts, Stella Biderman, Teven~Le Scao, M~Saiful Bari, Sheng Shen, Zheng-Xin Yong, Hailey Schoelkopf, et~al. 2022.
\newblock Crosslingual generalization through multitask finetuning.
\newblock \emph{arXiv preprint arXiv:2211.01786}.

\bibitem[{Muennighoff et~al.(2025)Muennighoff, Yang, Shi, Li, Fei-Fei, Hajishirzi, Zettlemoyer, Liang, and Emmanuel}]{muennighoff2025s1simpletesttimescaling}
Niklas Muennighoff, Zitong Yang, Weijia Shi, Xiang~Lisa Li, Li~Fei-Fei, Hannaneh Hajishirzi, Luke Zettlemoyer, Percy Liang, and Emmanuel. 2025.
\newblock \href {https://huggingface.co/datasets/simplescaling/s1K-1.1} {s1: Simple test-time scaling}.

\bibitem[{Nguyen et~al.(2023)Nguyen, Suri, Tsui, and Schuhmann}]{oig2023}
Huu Nguyen, Sameer Suri, Ken Tsui, and Christoph Schuhmann. 2023.
\newblock The open instruction generalist (oig) dataset.
\newblock \url{https://laion.ai/blog/oig-dataset/}.

\bibitem[{Ni et~al.(2023)Ni, Xue, Jain, Shah, Zheng, and You}]{instructionwild}
Jinjie Ni, Fuzhao Xue, Kabir Jain, Mahir~Hitesh Shah, Zangwei Zheng, and Yang You. 2023.
\newblock Instruction in the wild: A user-based instruction dataset.
\newblock \url{https://github.com/XueFuzhao/InstructionWild}.

\bibitem[{NVIDIA(2025)}]{nvidia2025llama}
NVIDIA. 2025.
\newblock Llama-nemotron-post-training dataset: A comprehensive collection of instruction tuning and alignment data.
\newblock \emph{arXiv preprint arXiv:2505.00949}.

\bibitem[{Ouyang et~al.(2022)Ouyang, Wu, Jiang, Almeida et~al.}]{ouyang2022training}
Long Ouyang, Jeff Wu, Xu~Jiang, Diogo Almeida, et~al. 2022.
\newblock Training language models to follow instructions with human feedback.
\newblock \emph{arXiv preprint arXiv:2203.02155}.

\bibitem[{Penedo et~al.(2025)Penedo, Lozhkov, Kydlíček, Allal, Beeching, Lajarín, Gallouédec, Habib, Tunstall, and von Werra}]{penedo2025codeforces}
Guilherme Penedo, Anton Lozhkov, Hynek Kydlíček, Loubna~Ben Allal, Edward Beeching, Agustín~Piqueres Lajarín, Quentin Gallouédec, Nathan Habib, Lewis Tunstall, and Leandro von Werra. 2025.
\newblock Codeforces cots.
\newblock \url{https://huggingface.co/datasets/open-r1/codeforces-cots}.

\bibitem[{Pudjiati et~al.(2022)Pudjiati, Lustyantie, Iskandar, and Fitria}]{pudjiati2022post}
Danti Pudjiati, Ninuk Lustyantie, Ifan Iskandar, and Tira~Nur Fitria. 2022.
\newblock Post-editing of machine translation: Creating a better translation of cultural specific terms.
\newblock \emph{Language Circle: Journal of Language and Literature}, 17(1):61--73.

\bibitem[{Puduppully et~al.(2024)Puduppully, Shrotriya, Kunchukuttan, and Khapra}]{puduppully2024mathinstruct}
Ratish Puduppully, Himani Shrotriya, Anoop Kunchukuttan, and Mitesh~M. Khapra. 2024.
\newblock \href {https://huggingface.co/datasets/jiuhai/MathInstruct} {Mathinstruct: A high-quality math instruction dataset}.

\bibitem[{Rafailov et~al.(2023)Rafailov, Sharma, Mitchell, Manning, Ermon, and Finn}]{rafailov2023direct}
Rafael Rafailov, Archit Sharma, Eric Mitchell, Christopher~D Manning, Stefano Ermon, and Chelsea Finn. 2023.
\newblock Direct preference optimization: Your language model is secretly a reward model.
\newblock \emph{Advances in neural information processing systems}, 36:53728--53741.

\bibitem[{Rajani et~al.(2023)Rajani, Tunstall, Beeching, Lambert, Rush, and Wolf}]{no_robots}
Nazneen Rajani, Lewis Tunstall, Edward Beeching, Nathan Lambert, Alexander~M. Rush, and Thomas Wolf. 2023.
\newblock No robots.
\newblock \url{https://huggingface.co/datasets/HuggingFaceH4/no_robots}.

\bibitem[{Rajbhandari et~al.(2020)Rajbhandari, Rasley, Ruwase, and He}]{rajbhandari2020zero}
Samyam Rajbhandari, Jeff Rasley, Olatunji Ruwase, and Yuxiong He. 2020.
\newblock Zero: Memory optimizations toward training trillion parameter models.
\newblock In \emph{Proceedings of the International Conference for High Performance Computing, Networking, Storage and Analysis (SC)}.

\bibitem[{Rajpurkar et~al.(2018)Rajpurkar, Jia, and Liang}]{rajpurkar2018know}
Pranav Rajpurkar, Robin Jia, and Percy Liang. 2018.
\newblock Know what you don't know: Unanswerable questions for squad.
\newblock \emph{arXiv preprint arXiv:1806.03822}.

\bibitem[{Rein et~al.(2024)Rein, Hou, Stickland, Petty, Pang, Dirani, Michael, and Bowman}]{rein2024gpqa}
David Rein, Betty~Li Hou, Asa~Cooper Stickland, Jackson Petty, Richard~Yuanzhe Pang, Julien Dirani, Julian Michael, and Samuel~R. Bowman. 2024.
\newblock \href {https://openreview.net/forum?id=Ti67584b98} {{GPQA}: A graduate-level google-proof q\&a benchmark}.
\newblock In \emph{First Conference on Language Modeling}.

\bibitem[{Research(2024)}]{servicenow2024r1}
ServiceNow~AI Research. 2024.
\newblock R1-distill-sft: A dataset for instruction tuning with a focus on distillation from large language models.
\newblock \emph{arXiv preprint arXiv:2403.06350}.

\bibitem[{Rohera et~al.(2024)Rohera, Ginimav, Salunke, Sawant, and Joshi}]{rohera2024l3cube}
Pritika Rohera, Chaitrali Ginimav, Akanksha Salunke, Gayatri Sawant, and Raviraj Joshi. 2024.
\newblock L3cube-indicquest: A benchmark question answering dataset for evaluating knowledge of llms in indic context.
\newblock \emph{arXiv preprint arXiv:2405.18789}.

\bibitem[{Rothermund et~al.(2025)Rothermund, Vögele, Roth, Schmutz, Wiegand, and Aschenbrenner}]{rothermund2025finemedlm}
Christian Rothermund, David Vögele, Lucas Roth, Philipp Schmutz, Tobias Wiegand, and Sebastian Aschenbrenner. 2025.
\newblock Finemedlm-o1: Enhancing medical knowledge reasoning ability of llm from supervised fine-tuning to test-time training.
\newblock \emph{arXiv preprint arXiv:2501.09213}.

\bibitem[{Sanh et~al.(2022)Sanh, Webson, Raffel, Bach, Sutawika, Alyafeai, Chaffin, Stiegler, Raja, Dey, Bari, Xu, Thakker, Sharma, Szczechla, Kim, Chhablani, Nayak, Datta, Chang, Jiang, Wang, Manica, Shen, Yong, Pandey, Bawden, Wang, Neeraj, Rozen, Sharma, Santilli, Fevry, Fries, Teehan, Scao, Biderman, Gao, Wolf, and Rush}]{sanh2022multitask}
Victor Sanh, Albert Webson, Colin Raffel, Stephen Bach, Lintang Sutawika, Zaid Alyafeai, Antoine Chaffin, Arnaud Stiegler, Arun Raja, Manan Dey, M~Saiful Bari, Canwen Xu, Urmish Thakker, Shanya~Sharma Sharma, Eliza Szczechla, Taewoon Kim, Gunjan Chhablani, Nihal Nayak, Debajyoti Datta, Jonathan Chang, Mike Tian-Jian Jiang, Han Wang, Matteo Manica, Sheng Shen, Zheng~Xin Yong, Harshit Pandey, Rachel Bawden, Thomas Wang, Trishala Neeraj, Jos Rozen, Abheesht Sharma, Andrea Santilli, Thibault Fevry, Jason~Alan Fries, Ryan Teehan, Teven~Le Scao, Stella Biderman, Leo Gao, Thomas Wolf, and Alexander~M Rush. 2022.
\newblock \href {https://openreview.net/forum?id=9Vrb9D0WI4} {Multitask prompted training enables zero-shot task generalization}.
\newblock In \emph{International Conference on Learning Representations}.

\bibitem[{Savoldi et~al.(2021)Savoldi, Gaido, Bentivogli, Negri, and Turchi}]{savoldi2021gender}
Beatrice Savoldi, Marco Gaido, Luisa Bentivogli, Matteo Negri, and Marco Turchi. 2021.
\newblock \href {http://arxiv.org/abs/2104.06001} {Gender bias in machine translation}.

\bibitem[{Singh et~al.(2024{\natexlab{a}})Singh, Gupta, Bharadwaj, Tewari, and Talukdar}]{singh2024indicgenbench}
Harman Singh, Nitish Gupta, Shikhar Bharadwaj, Dinesh Tewari, and Partha Talukdar. 2024{\natexlab{a}}.
\newblock \href {http://arxiv.org/abs/2404.16816} {Indicgenbench: A multilingual benchmark to evaluate generation capabilities of llms on indic languages}.

\bibitem[{Singh et~al.(2024{\natexlab{b}})Singh, Vargus, Dsouza, Karlsson, Mahendiran, Ko, Shandilya, Patel, Mataciunas, OMahony, Zhang, Hettiarachchi, Wilson, Machado, Moura, Krzemiński, Fadaei, Ergün, Okoh, Alaagib, Mudannayake, Alyafeai, Chien, Ruder, Guthikonda, Alghamdi, Gehrmann, Muennighoff, Bartolo, Kreutzer, Üstün, Fadaee, and Hooker}]{singh2024aya}
Shivalika Singh, Freddie Vargus, Daniel Dsouza, Börje~F. Karlsson, Abinaya Mahendiran, Wei-Yin Ko, Herumb Shandilya, Jay Patel, Deividas Mataciunas, Laura OMahony, Mike Zhang, Ramith Hettiarachchi, Joseph Wilson, Marina Machado, Luisa~Souza Moura, Dominik Krzemiński, Hakimeh Fadaei, Irem Ergün, Ifeoma Okoh, Aisha Alaagib, Oshan Mudannayake, Zaid Alyafeai, Vu~Minh Chien, Sebastian Ruder, Surya Guthikonda, Emad~A. Alghamdi, Sebastian Gehrmann, Niklas Muennighoff, Max Bartolo, Julia Kreutzer, Ahmet Üstün, Marzieh Fadaee, and Sara Hooker. 2024{\natexlab{b}}.
\newblock \href {http://arxiv.org/abs/2402.06619} {Aya dataset: An open-access collection for multilingual instruction tuning}.

\bibitem[{Stojanovski et~al.(2025)Stojanovski, Stanley, Sharratt, Jones, Adefioye, Kaddour, and Köpf}]{stojanovski2025reasoninggymreasoningenvironments}
Zafir Stojanovski, Oliver Stanley, Joe Sharratt, Richard Jones, Abdulhakeem Adefioye, Jean Kaddour, and Andreas Köpf. 2025.
\newblock \href {http://arxiv.org/abs/2505.24760} {Reasoning gym: Reasoning environments for reinforcement learning with verifiable rewards}.

\bibitem[{Talmor et~al.(2019)Talmor, Herzig, Lourie, and Berant}]{talmor-etal-2019-commonsenseqa}
Alon Talmor, Jonathan Herzig, Nicholas Lourie, and Jonathan Berant. 2019.
\newblock {C}ommonsense{QA}: A question answering challenge targeting commonsense knowledge.
\newblock In \emph{Proceedings of the 2019 Conference of the North American Chapter of the Association for Computational Linguistics: Human Language Technologies, Volume 1 (Long and Short Papers)}.

\bibitem[{Taori et~al.(2023)Taori, Gulrajani, Zhang, Dubois, Li, Guestrin, Liang, and Hashimoto}]{alpaca}
Rohan Taori, Ishaan Gulrajani, Tianyi Zhang, Yann Dubois, Xuechen Li, Carlos Guestrin, Percy Liang, and Tatsunori~B Hashimoto. 2023.
\newblock Stanford alpaca: An instruction-following llama model.

\bibitem[{Toshniwal et~al.(2024)Toshniwal, Moshkov, Narenthiran, Gitman, Jia, and Gitman}]{toshniwal2024openmath}
Shubham Toshniwal, Ivan Moshkov, Sean Narenthiran, Daria Gitman, Fei Jia, and Igor Gitman. 2024.
\newblock Openmathinstruct-1: A 1.8 million math instruction tuning dataset.
\newblock \emph{arXiv preprint arXiv: Arxiv-2402.10176}.

\bibitem[{Touileb et~al.(2024)Touileb, Longpre, L'Heureux, Houlsby, Le, Aribaud, Le~Scao, Pasquini, Miaschi, and Muennighoff}]{touileb2024magpie}
Foutse Touileb, Xavier Longpre, Adrien L'Heureux, Tom Houlsby, Duy-Kien Le, Vincent Aribaud, Hugo Le~Scao, Tommaso Pasquini, Marco Miaschi, and Patrick Muennighoff. 2024.
\newblock Magpie: A broad-coverage, fine-grained multitask instruction dataset.
\newblock \emph{arXiv preprint arXiv:2403.00010}.

\bibitem[{Touvron et~al.(2023)Touvron, Martin, Stone, Albert, Almahairi, Babaei, Bashlykov, Batra, Bhargava, Bhosale, Bikel, Blecher, Ferrer, Chen, Cucurull, Esiobu, Fernandes, Fu, Fu, Fuller, Gao, Goswami, Goyal, Hartshorn, Hosseini, Hou, Inan, Kardas, Kerkez, Khabsa, Kloumann, Korenev, Koura, Lachaux, Lavril, Lee, Liskovich, Lu, Mao, Martinet, Mihaylov, Mishra, Molybog, Nie, Poulton, Reizenstein, Rungta, Saladi, Schelten, Silva, Smith, Subramanian, Tan, Tang, Taylor, Williams, Kuan, Xu, Yan, Zarov, Zhang, Fan, Kambadur, Narang, Rodriguez, Stojnic, Edunov, and Scialom}]{touvron2023llama}
Hugo Touvron, Louis Martin, Kevin Stone, Peter Albert, Amjad Almahairi, Yasmine Babaei, Nikolay Bashlykov, Soumya Batra, Prajjwal Bhargava, Shruti Bhosale, Dan Bikel, Lukas Blecher, Cristian~Canton Ferrer, Moya Chen, Guillem Cucurull, David Esiobu, Jude Fernandes, Jeremy Fu, Wenyin Fu, Brian Fuller, Cynthia Gao, Vedanuj Goswami, Naman Goyal, Anthony Hartshorn, Saghar Hosseini, Rui Hou, Hakan Inan, Marcin Kardas, Viktor Kerkez, Madian Khabsa, Isabel Kloumann, Artem Korenev, Punit~Singh Koura, Marie-Anne Lachaux, Thibaut Lavril, Jenya Lee, Diana Liskovich, Yinghai Lu, Yuning Mao, Xavier Martinet, Todor Mihaylov, Pushkar Mishra, Igor Molybog, Yixin Nie, Andrew Poulton, Jeremy Reizenstein, Rashi Rungta, Kalyan Saladi, Alan Schelten, Ruan Silva, Eric~Michael Smith, Ranjan Subramanian, Xiaoqing~Ellen Tan, Binh Tang, Ross Taylor, Adina Williams, Jian~Xiang Kuan, Puxin Xu, Zheng Yan, Iliyan Zarov, Yuchen Zhang, Angela Fan, Melanie Kambadur, Sharan Narang, Aurelien Rodriguez, Robert Stojnic, Sergey Edunov, and Thomas
  Scialom. 2023.
\newblock \href {http://arxiv.org/abs/2307.09288} {Llama 2: Open foundation and fine-tuned chat models}.

\bibitem[{Wang et~al.(2022{\natexlab{a}})Wang, Rubinstein, and Cohn}]{wang2022measuring}
Jun Wang, Benjamin Rubinstein, and Trevor Cohn. 2022{\natexlab{a}}.
\newblock Measuring and mitigating name biases in neural machine translation.
\newblock In \emph{ACL}.

\bibitem[{Wang et~al.(2022{\natexlab{b}})Wang, Kordi, Mishra, Liu, Smith, Khashabi, and Hajishirzi}]{selfinstruct}
Yizhong Wang, Yeganeh Kordi, Swaroop Mishra, Alisa Liu, Noah~A. Smith, Daniel Khashabi, and Hannaneh Hajishirzi. 2022{\natexlab{b}}.
\newblock Self-instruct: Aligning language model with self generated instructions.

\bibitem[{Wang et~al.(2022{\natexlab{c}})Wang, Kordi, Mishra, Liu, Smith, Khashabi, and Hajishirzi}]{wang2022self}
Yizhong Wang, Yeganeh Kordi, Swaroop Mishra, Alisa Liu, Noah~A Smith, Daniel Khashabi, and Hannaneh Hajishirzi. 2022{\natexlab{c}}.
\newblock \href {https://arxiv.org/abs/2212.10560} {Self-instruct: Aligning language model with self generated instructions}.
\newblock \emph{ArXiv preprint}, abs/2212.10560.

\bibitem[{Wang et~al.(2022{\natexlab{d}})Wang, Li, Michael, Gonen, He, Zhao, Lin, Shafi, Singh, Touileb, and et~al.}]{wang2022supernaturalinstructions}
Yizhong Wang, Yeganeh Li, Niklas Michael, Hila Gonen, Jesse He, Yi~Zhao, Ziyi Lin, Yejin Shafi, Karan Singh, Foutse Touileb, and et~al. 2022{\natexlab{d}}.
\newblock Super-naturalinstructions: A generalization beyond english language.
\newblock \emph{arXiv preprint arXiv:2204.05367}.

\bibitem[{Wei et~al.(2022{\natexlab{a}})Wei, Bosma, Zhao, Guu, Yu, Lester, Du, Dai, and Le}]{wei2022finetuned}
Jason Wei, Maarten Bosma, Vincent Zhao, Kelvin Guu, Adams~Wei Yu, Brian Lester, Nan Du, Andrew~M. Dai, and Quoc~V Le. 2022{\natexlab{a}}.
\newblock \href {https://openreview.net/forum?id=gEZrGCozdqR} {Finetuned language models are zero-shot learners}.
\newblock In \emph{International Conference on Learning Representations}.

\bibitem[{Wei et~al.(2022{\natexlab{b}})Wei, Wang, Schuurmans, Bosma, Xia, Chi, Le, Zhou et~al.}]{wei2022chain}
Jason Wei, Xuezhi Wang, Dale Schuurmans, Maarten Bosma, Fei Xia, Ed~Chi, Quoc~V Le, Denny Zhou, et~al. 2022{\natexlab{b}}.
\newblock Chain-of-thought prompting elicits reasoning in large language models.
\newblock \emph{Advances in neural information processing systems}, 35:24824--24837.

\bibitem[{Welleck et~al.(2021)Welleck, Le~Bras, Hajishirzi, and Choi}]{welleck2021naturalproofs}
Sean Welleck, Ronan Le~Bras, Hannaneh Hajishirzi, and Yejin Choi. 2021.
\newblock Naturalproofs: Mathematical theorem proving in natural language.
\newblock \emph{arXiv preprint arXiv:2104.01112}.

\bibitem[{Xie et~al.(2022)Xie, Wu, Shi, Zhong, Scholak, Yasunaga, Wu, Zhong, Yin, Wang et~al.}]{xie2022unifiedskg}
Tianbao Xie, Chen~Henry Wu, Peng Shi, Ruiqi Zhong, Torsten Scholak, Michihiro Yasunaga, Chien-Sheng Wu, Ming Zhong, Pengcheng Yin, Sida~I Wang, et~al. 2022.
\newblock Unifiedskg: Unifying and multi-tasking structured knowledge grounding with text-to-text language models.
\newblock \emph{arXiv preprint arXiv:2201.05966}.

\bibitem[{Xu et~al.(2023)Xu, Sun, Zheng, Geng, Zhao, Feng, Tao, and Jiang}]{xu2023wizardlm}
Can Xu, Qingfeng Sun, Kai Zheng, Xiubo Geng, Pu~Zhao, Jiazhan Feng, Chongyang Tao, and Daxin Jiang. 2023.
\newblock \href {http://arxiv.org/abs/2304.12244} {Wizardlm: Empowering large language models to follow complex instructions}.

\bibitem[{Xu et~al.(2024)Xu, Jiang, Niu, Deng, Poovendran, Choi, and Lin}]{argilla2024magpie}
Zhangchen Xu, Fengqing Jiang, Luyao Niu, Yuntian Deng, Radha Poovendran, Yejin Choi, and Bill~Yuchen Lin. 2024.
\newblock \href {http://arxiv.org/abs/2406.08464} {Magpie: Alignment data synthesis from scratch by prompting aligned llms with nothing}.

\bibitem[{Xu et~al.(2025)Xu, Liu, Yin, Zhou, and Poovendran}]{xu-etal-2025-kodcode}
Zhangchen Xu, Yang Liu, Yueqin Yin, Mingyuan Zhou, and Radha Poovendran. 2025.
\newblock {K}od{C}ode: A diverse, challenging, and verifiable synthetic dataset for coding.
\newblock In \emph{Findings of the Association for Computational Linguistics: ACL 2025}, pages 6980--7008.

\bibitem[{Yang et~al.(2015)Yang, Yih, and Meek}]{yang2015wikiqa}
Yi~Yang, Wen-tau Yih, and Christopher Meek. 2015.
\newblock Wikiqa: A challenge dataset for open-domain question answering.
\newblock In \emph{Proceedings of the 2015 Conference on Empirical Methods in Natural Language Processing}.

\bibitem[{Yang et~al.(2018)Yang, Qi, Zhang, Bengio, Cohen, Salakhutdinov, and Manning}]{yang-etal-2018-hotpotqa}
Zhilin Yang, Peng Qi, Saizheng Zhang, Yoshua Bengio, William Cohen, Ruslan Salakhutdinov, and Christopher~D. Manning. 2018.
\newblock \href {https://doi.org/10.18653/v1/D18-1259} {{H}otpot{QA}: A dataset for diverse, explainable multi-hop question answering}.
\newblock In \emph{Proceedings of the 2018 Conference on Empirical Methods in Natural Language Processing}, pages 2369--2380, Brussels, Belgium. Association for Computational Linguistics.

\bibitem[{Ye et~al.(2025)Ye, Huang, Xiao, Chern, Xia, and Liu}]{ye2025limoreasoning}
Yixin Ye, Zhen Huang, Yang Xiao, Ethan Chern, Shijie Xia, and Pengfei Liu. 2025.
\newblock \href {http://arxiv.org/abs/2502.03387} {Limo: Less is more for reasoning}.

\bibitem[{Yu et~al.(2023{\natexlab{a}})Yu, Liu, Yang, Wang, Wang, Li, Zhang, and Zhang}]{yu2023metamath}
Xiang Yu, Haidong Liu, Ning Yang, Qiaoli Wang, Jianye Wang, Jia Li, Li~Zhang, and Jian Zhang. 2023{\natexlab{a}}.
\newblock Metamath: Bootstrap your own mathematical reasoning dataset.
\newblock \emph{arXiv preprint arXiv:2309.12284}.

\bibitem[{Yu et~al.(2023{\natexlab{b}})Yu, Zhuang, Zhang, Meng, Ratner, Krishna, Shen, and Zhang}]{yu2023large}
Yue Yu, Yuchen Zhuang, Jieyu Zhang, Yu~Meng, Alexander Ratner, Ranjay Krishna, Jiaming Shen, and Chao Zhang. 2023{\natexlab{b}}.
\newblock \href {http://arxiv.org/abs/2306.15895} {Large language model as attributed training data generator: A tale of diversity and bias}.

\bibitem[{Zhang et~al.(2024)Zhang, Liu, Xu, Wang, Chen, Li, Li, Xie, and Zhang}]{zhang2024open-thoughts}
Tianyu Zhang, Yang Liu, Yuxuan Xu, Yujia Wang, Haoyu Chen, Hongwei Li, Chen Li, Jiawei Xie, and Chen Zhang. 2024.
\newblock Open-thoughts: A large-scale dataset for learning and evaluating llm thinking processes.
\newblock \emph{arXiv preprint arXiv:2406.04178}.

\bibitem[{Zhou et~al.(2023)Zhou, Liu, Xu, Iyer, Sun, Mao, Ma, Efrat, Yu, Yu et~al.}]{zhou2023lima}
Chunting Zhou, Pengfei Liu, Puxin Xu, Srini Iyer, Jiao Sun, Yuning Mao, Xuezhe Ma, Avia Efrat, Ping Yu, Lili Yu, et~al. 2023.
\newblock Lima: Less is more for alignment.
\newblock \emph{arXiv preprint arXiv:2305.11206}.

\bibitem[{Zhuo et~al.(2024)Zhuo, Zebaze, Suppattarachai, von Werra, de~Vries, Liu, and Muennighoff}]{zhuo2024astraios}
Terry~Yue Zhuo, Armel Zebaze, Nitchakarn Suppattarachai, Leandro von Werra, Harm de~Vries, Qian Liu, and Niklas Muennighoff. 2024.
\newblock Astraios: Parameter-efficient instruction tuning code large language models.
\newblock \emph{arXiv preprint arXiv:2401.00788}.

\end{thebibliography}
\bibliographystyle{acl_natbib}


\appendix

\section{Appendix}
\label{sec:appendix}

\subsection{Dataset Curation}
For each sub-category, we curated multiple publicly available candidate datasets and evaluated their applicability to our data collection objectives, assessing how they align with the chosen curation approaches. Table \ref{tab:datasets} summarizes representative categories along with corresponding datasets incorporated into our pipeline. It is worth noting that data collection for certain sub-categories remains an ongoing effort.

\begin{table*}[htbp]
\scriptsize
\centering
\begin{tabular}{p{2.7cm}p{5.4cm}p{2.9cm}p{4.3cm}}
\hline
\textbf{Sub Category} & \textbf{Candidate Datasets} & \textbf{Sub Category} & \textbf{Candidate Datasets} \\ \hline

Non-Math Reasoning & \begin{tabular}[c]{@{}l@{}} MMLU \citep{hendrycks2020measuring} \\ MMLU-Pro \citep{liu2024mmul-pro} \\ GPQA \citep{rein2024gpqa} \\ medical-o1-reasoning-SFT \citep{rothermund2025finemedlm} \\ OpenThoughts \citep{zhang2024open-thoughts} \end{tabular} 
& Sentiment Analysis & \begin{tabular}[c]{@{}l@{}} IndicSentiment \citep{doddapaneni-etal-2023-towards} \\ pietrolesci/imdb \end{tabular} \\ \hline

Math Reasoning & \begin{tabular}[c]{@{}l@{}} OpenThoughts \citep{zhang2024open-thoughts} \\ Llama-Nemotron-Posttraining-Dataset \citep{nvidia2025llama} \\ ServiceNow-AI/R1-Distil-SFT \citep{servicenow2024r1} \\ LIMO \citep{ye2025limoreasoning} \\ s1K \citep{muennighoff2025s1simpletesttimescaling} \end{tabular}
& General Question Answering & \begin{tabular}[c]{@{}l@{}} OpenBookQA \citep{mihaylov-etal-2018-openbookqa} \\ TriviaQA \citep{joshi2017triviaqa} \\ CommonSenseQA \citep{talmor-etal-2019-commonsenseqa} \\ WikiQA \citep{yang2015wikiqa} \\ HotPotQA \citep{yang-etal-2018-hotpotqa} \\ MegaWika \citep{barham2023megawika} \end{tabular} \\ \hline

Code Reasoning & \begin{tabular}[c]{@{}l@{}} OpenThoughts \citep{zhang2024open-thoughts} \\ Llama-Nemotron-Posttraining-Dataset \citep{nvidia2025llama} \\ KodCode \citep{xu-etal-2025-kodcode} \\ open-r1/codeforces-cots \citep{penedo2025codeforces} \end{tabular}
& Fact Check & \begin{tabular}[c]{@{}l@{}} NaturalQuestions \citep{Kwiatkowski2019Nq} \\ TriviaQA \citep{joshi2017triviaqa} \\ Indic Quest \citep{rohera2024l3cube} \end{tabular} \\ \hline

Indian Relationships & \begin{tabular}[c]{@{}l@{}} Reasoning Gym \citep{stojanovski2025reasoninggymreasoningenvironments} \end{tabular}
& Multi Turn Conversation & \begin{tabular}[c]{@{}l@{}} ChatAlpaca \citep{ChatAlpaca} \\ NoRobots \citep{no_robots} \\ Opus Samantha \end{tabular} \\ \hline

Inference & \begin{tabular}[c]{@{}l@{}} XNLI \citep{conneau2018xnli} \end{tabular}
& Role Playing & \begin{tabular}[c]{@{}l@{}} Public Domain Alpaca \\ LimaRP \\ Magpie \citep{touileb2024magpie} \end{tabular} \\ \hline

Data Analysis & \begin{tabular}[c]{@{}l@{}} Magpie \citep{touileb2024magpie} \end{tabular}
& Advice Seeking & \begin{tabular}[c]{@{}l@{}} Magpie \citep{touileb2024magpie} \end{tabular} \\ \hline

Named Entity Recognition & \begin{tabular}[c]{@{}l@{}} Naamapadam \citep{mhaske2022naamapadam} \\ Bharath Bench \end{tabular}
& Planning & \begin{tabular}[c]{@{}l@{}} Magpie \citep{touileb2024magpie} \end{tabular} \\ \hline

Text Classification & \begin{tabular}[c]{@{}l@{}} pietrolesci/civilcomments-wilds  \citep{borkan2019nuanced} \\ pietrolesci/wikitoxic \\ pietrolesci/hyperpartisan\_news\_detection \\ Bharath Bench \\ Aya Collection \citep{singh2024aya} \end{tabular}
& Brainstorming & \begin{tabular}[c]{@{}l@{}} Magpie \citep{touileb2024magpie} \end{tabular} \\ \hline

Grammar Correction & \begin{tabular}[c]{@{}l@{}} Bharath Bench \end{tabular}
& Information Seeking & \begin{tabular}[c]{@{}l@{}} Magpie \citep{touileb2024magpie} \end{tabular} \\ \hline

Translation & \begin{tabular}[c]{@{}l@{}} Flores-IN \citep{singh2024indicgenbench} \\ IN-22 \end{tabular}
& Indian Cultural Context & \begin{tabular}[c]{@{}l@{}} Bharath Bench \end{tabular} \\ \hline

Creative Writing & \begin{tabular}[c]{@{}l@{}} Magpie \citep{touileb2024magpie} \\ Poetry \end{tabular}
& Comprehension & \begin{tabular}[c]{@{}l@{}} TriviaQA \citep{joshi2017triviaqa} \\ SQUAD \citep{rajpurkar2018know} \\ SQUAD 2.0 \citep{rajpurkar2018know} \\ IndicQA \end{tabular} \\ \hline

Paraphrase Identification & \begin{tabular}[c]{@{}l@{}} IndicXParaphrase \citep{doddapaneni-etal-2023-towards} \end{tabular}
& Math QA & \begin{tabular}[c]{@{}l@{}} OpenMathInstruct 2 \citep{toshniwal2024openmath} \\ GSM8K \citep{cobbe2021gsm8k} \\ MATH \citep{hendrycksmath2021} \\ Tulu3 Persona Math \citep{lambert2024tulu3} \\ Tulu3 Persona GSM8K \citep{lambert2024tulu3} \end{tabular} \\ \hline

Text Summarization & \begin{tabular}[c]{@{}l@{}} CrossSumIN \citep{singh2024indicgenbench} \\ Indic Sentence Summarization \citep{Kumar2022IndicNLGSM} \\ arxiv-summarization \citep{cohan-etal-2018-discourse} \\ news-summarization \end{tabular}
& Math Instruction Tuning & \begin{tabular}[c]{@{}l@{}} MetaMathQA \citep{yu2023metamath} \\ Math Instruct \citep{puduppully2024mathinstruct} \end{tabular} \\ \hline

Headline Generation & \begin{tabular}[c]{@{}l@{}} Indic Headline Generation \citep{Kumar2022IndicNLGSM} \\ NewSHead \end{tabular}
& Math Proofs & \begin{tabular}[c]{@{}l@{}} Natural Proofs \citep{welleck2021naturalproofs} \end{tabular} \\ \hline

Question Generation & \begin{tabular}[c]{@{}l@{}} Indic Question Generation \citep{Kumar2022IndicNLGSM} \end{tabular} & & \\ \hline

\end{tabular}
\caption{Representative datasets curated for different sub-categories that are used as candidate datasets in our data curation approaches.}
\label{tab:datasets}
\end{table*}

\subsection{Data Construction Approaches}

We further provide more details about the prompt and responses for both the approaches in Table \ref{tab:data_approaches}. 


\begin{table*}[htbp]
\centering
\small
\begin{tabular}{p{3.0cm}p{6.2cm}p{5.2cm}}
\hline
\textbf{Approaches} & \textbf{Prompts} & \textbf{Responses} \\ \hline
\textbf{Approach 1: Translation + Human Refinement} & 
English prompts $\rightarrow$ LLM translation $\rightarrow$ human verification/adaptation. 
\newline 
\textit{Outputs:} Indic Generic Prompts, Indic Context Prompts. &
English responses $\rightarrow$ LLM translation $\rightarrow$ human verification/adaptation. 
\newline 
\textit{Outputs:} Indic Generic Responses, Indic Context Responses. \\ \hline

\textbf{Approach 2: Synthetic Expansion + Human Refinement} & 
Seed English prompts $\rightarrow$ LLM expansion (Self-Instruct) $\rightarrow$ LLM translation $\rightarrow$ human verification/adaptation. 
\newline 
\textit{Outputs:} Synthetic Indic Generic Prompts, Synthetic Indic Context Prompts. & 
Synthetic English responses $\rightarrow$ LLM translation $\rightarrow$ human verification/adaptation. 
\newline 
\textit{Outputs:} Synthetic Indic Generic Responses, Synthetic Indic Context Responses. \\ \hline
\end{tabular}
\caption{Comparison of data construction approaches, showing pipelines for prompts and responses with resulting categories.}
\label{tab:data_approaches}
\end{table*}

\subsection{Complexity Definitions}
\begin{figure*}[htbp]
    \centering
\includegraphics[width=1\linewidth]{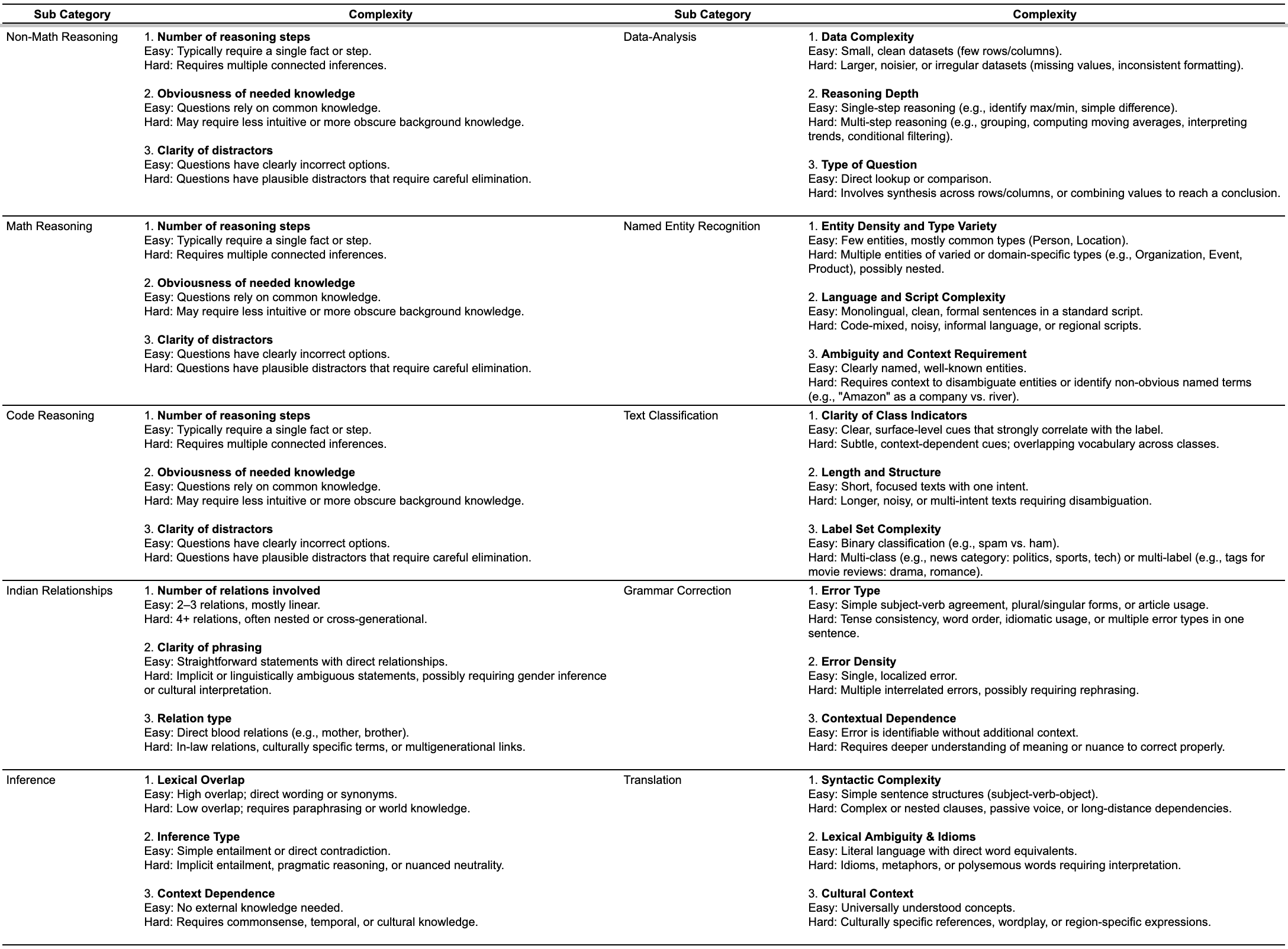} \caption{Complexity Definitions for each sub-category.}
    \label{fig:complexities_1}
\end{figure*}

\begin{figure*}[htbp]
    \centering
\includegraphics[width=1\linewidth]{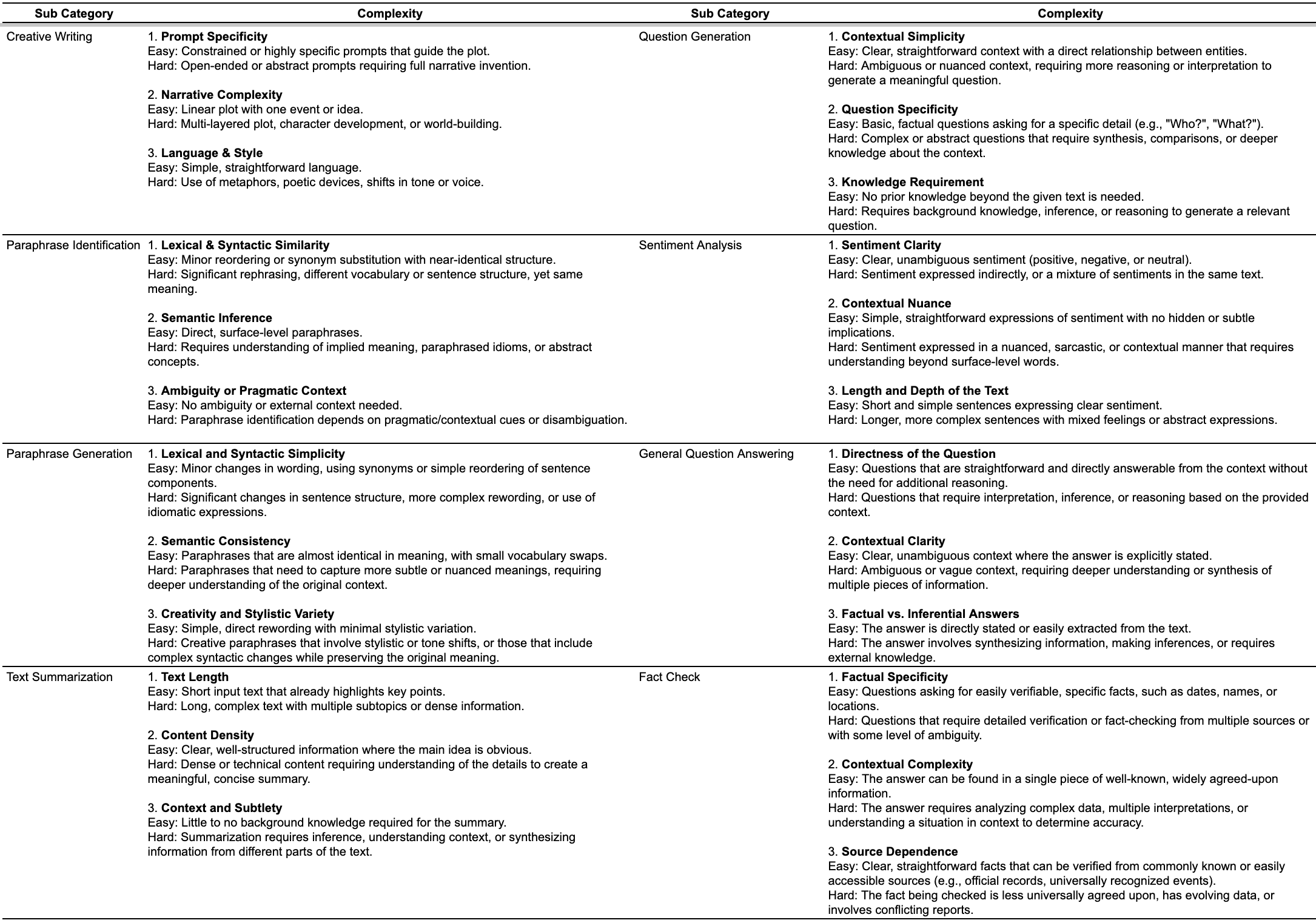} \caption{Complexity Definitions for each sub-category.}
    \label{fig:complexities_2}
\end{figure*}

\begin{figure*}[htbp]
    \centering
\includegraphics[width=1\linewidth]{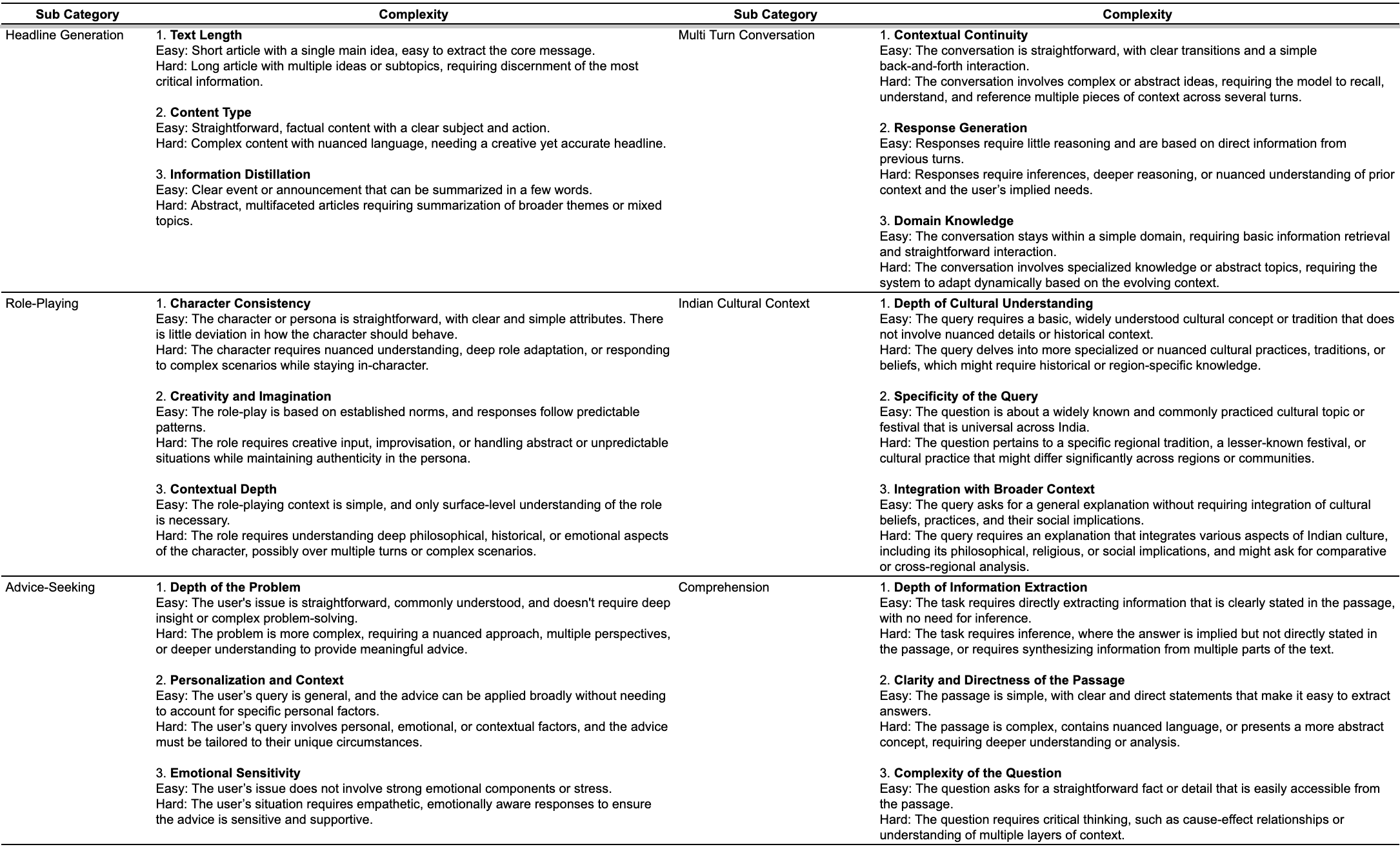} \caption{Complexity Definitions for each sub-category.}
    \label{fig:complexities_3}
\end{figure*}

\begin{figure*}[htbp]
    \centering
\includegraphics[width=1\linewidth]{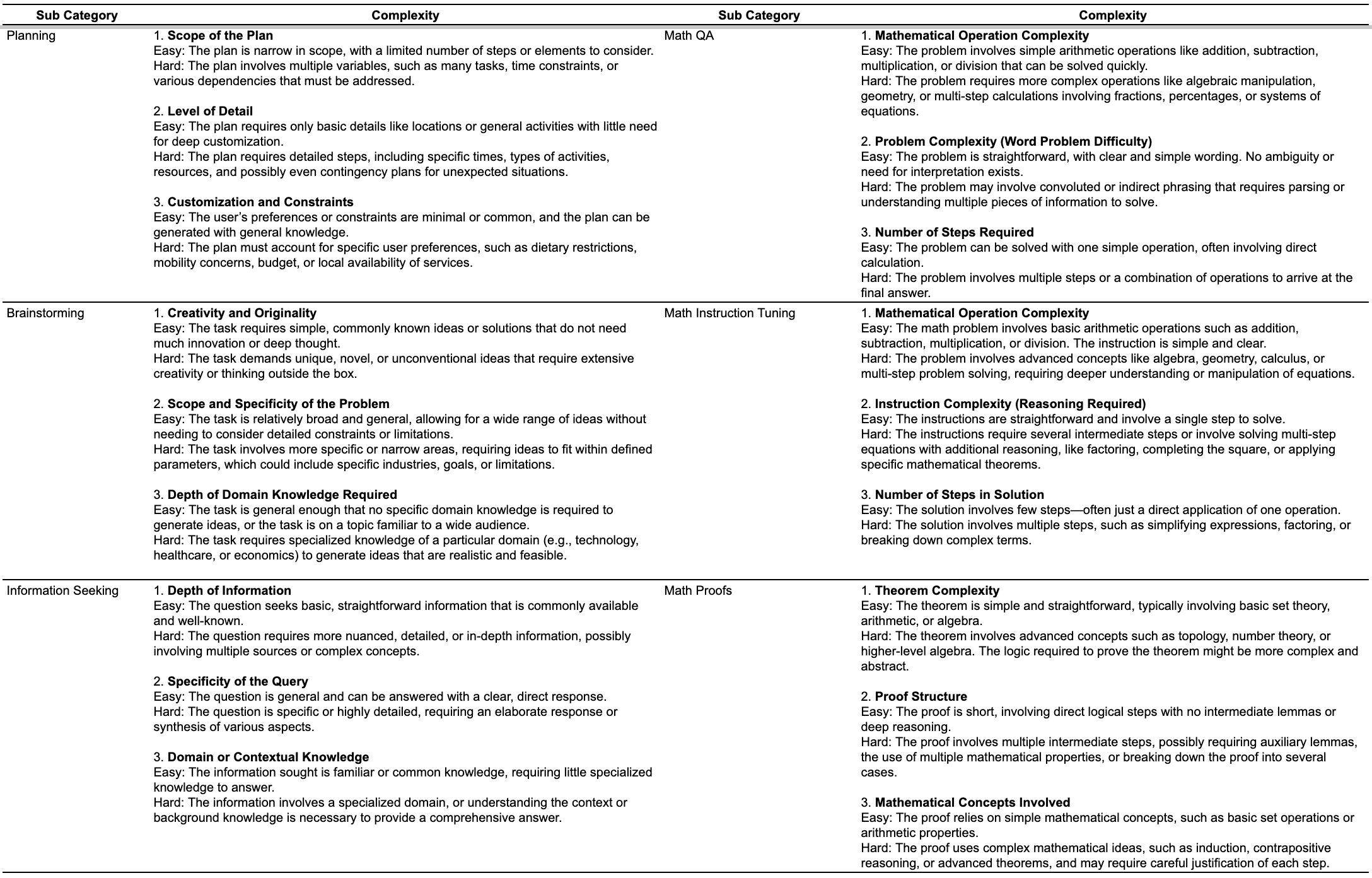} \caption{Complexity Definitions for each sub-category.}
    \label{fig:complexities_4}
\end{figure*}

\begin{figure*}[htbp]
    \centering
\includegraphics[width=1\linewidth]{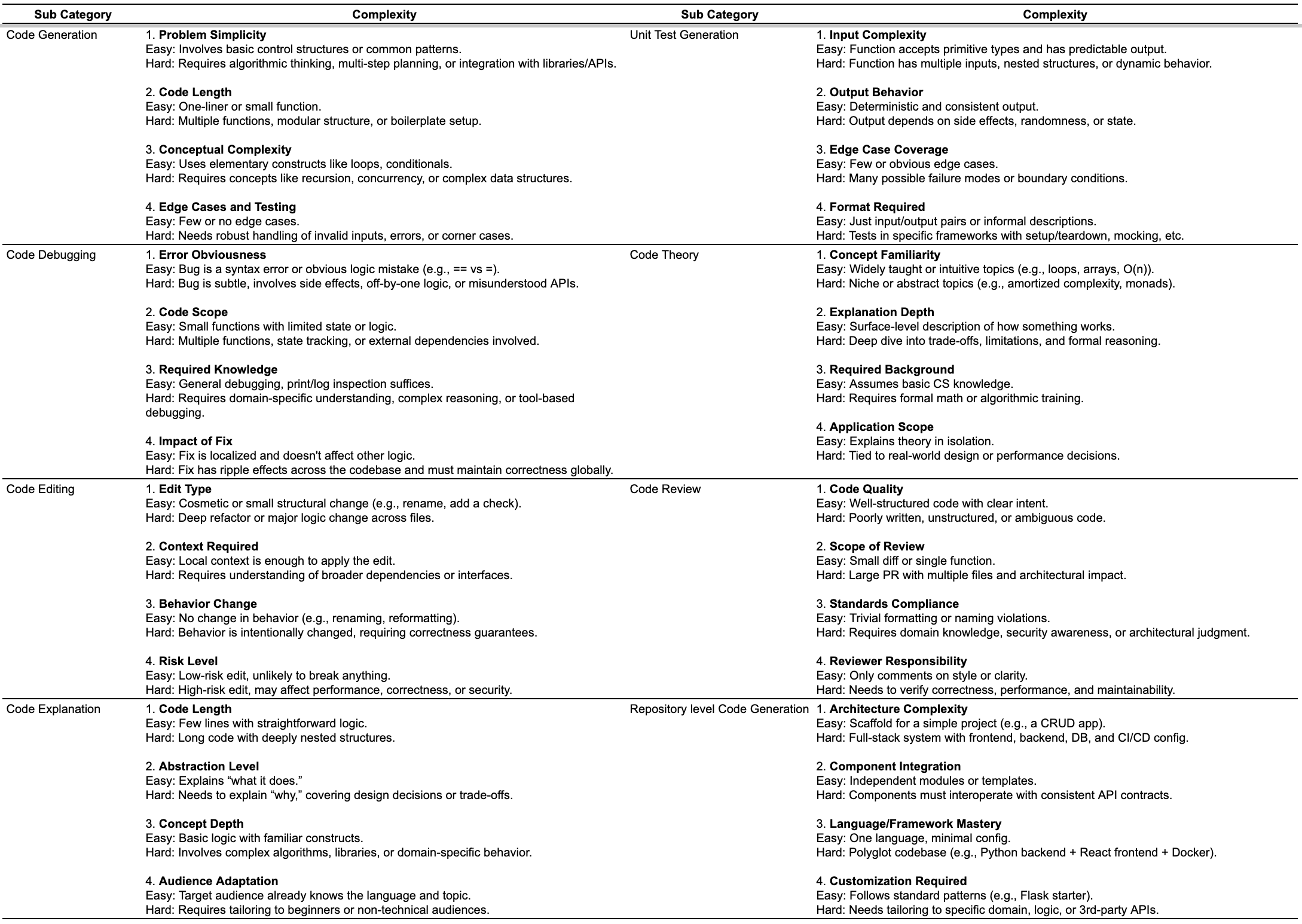} \caption{Complexity Definitions for each sub-category.}
    \label{fig:complexities_5}
\end{figure*}

\begin{figure*}[htbp]
    \centering
\includegraphics[width=1\linewidth]{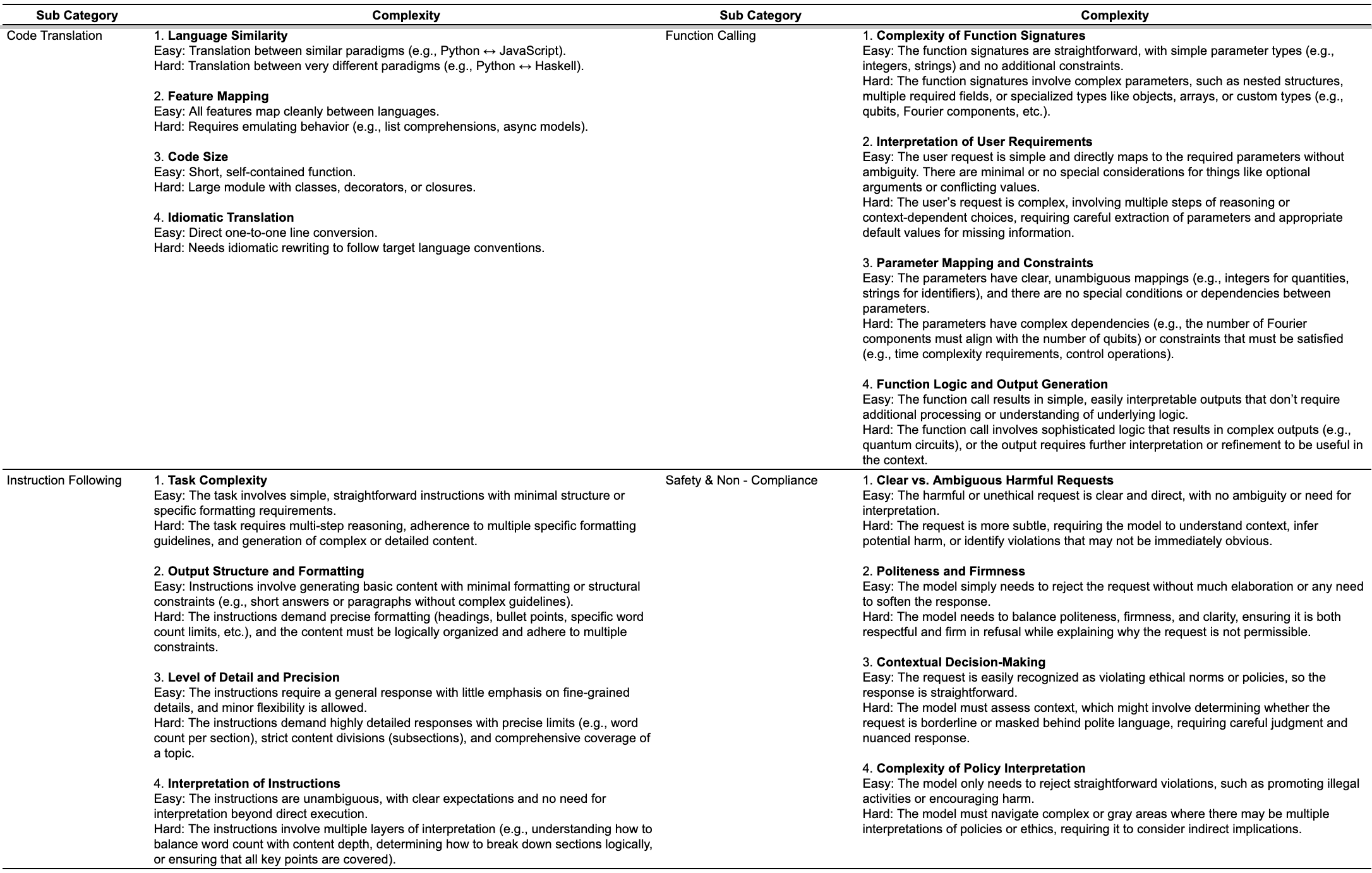} \caption{Complexity Definitions for each sub-category.}
    \label{fig:complexities_6}
\end{figure*}

Notably, complexity is evaluated relative to each task category, ensuring that difficulty levels are interpreted within the specific context of that category. Figures \ref{fig:complexities_1}–\ref{fig:complexities_6} illustrate how complexity is defined across different task categories.

\subsection{Instruction Following}
\begin{figure*}[htbp]
    \centering
\includegraphics[width=1\linewidth]{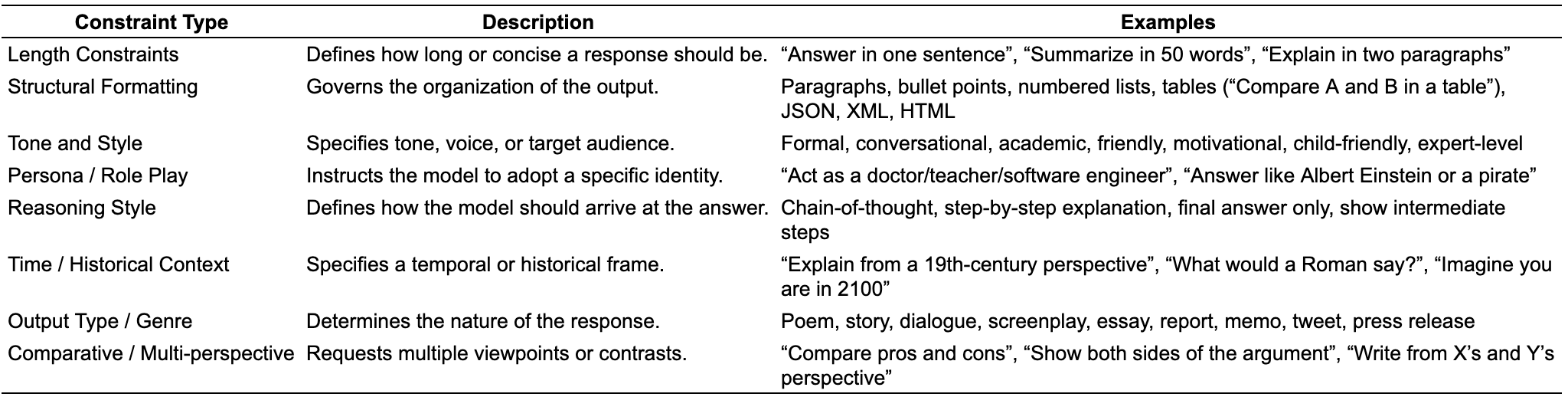} \caption{Instruction following: constraint types and examples}
    \label{fig:constraints}
\end{figure*}

Various configurations of Instruction Following incorporate different combinations of constraints within the prompt. Figure \ref{fig:constraints} outlines the types of constraints considered, accompanied by illustrative examples for clarity while human annotation.

\subsection{Human Annotation Refinement}
\label{appendix:human}
Data annotators initially assessed the performance of the underlying LLM on both generation and translation tasks across multiple dimensions. 

\begin{figure*}[htbp]
    \centering
    \includegraphics[width=1\linewidth]{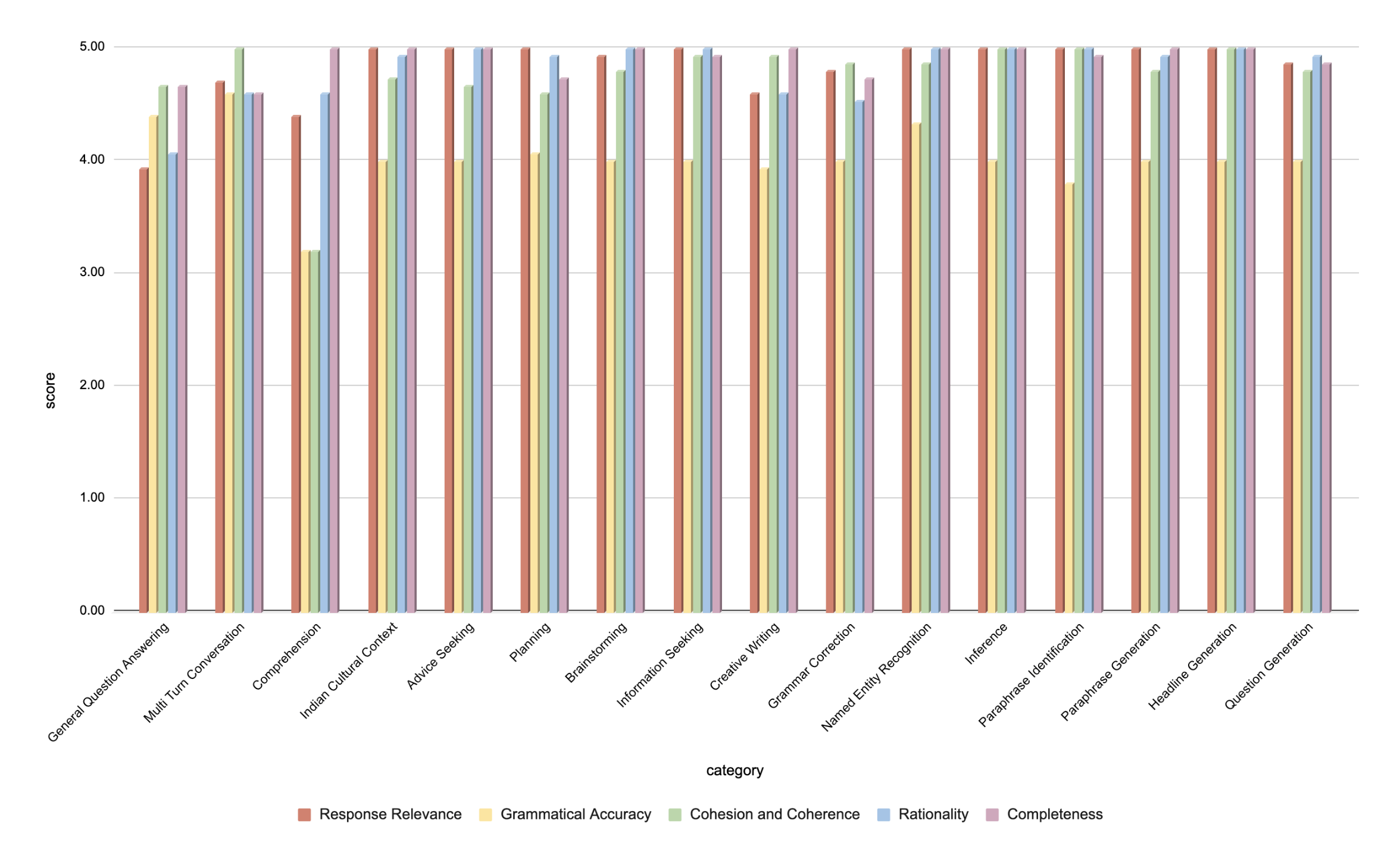}
    \caption{LLM generation quality evaluation scores across various categories.}
    \label{fig:generation_evaluation}
\end{figure*}

For generation, raw LLM responses were assessed along dimensions such as Response Relevance, Grammatical Accuracy, Cohesion and Coherence, Rationality, and Completeness (see Table~\ref{tab:gen-eval-dimensions} for the guidelines). As shown in Figure~\ref{fig:generation_evaluation}, English generations perform consistently well: tasks such as Indian Cultural Context, Advice Seeking, Information Seeking, Named Entity Recognition, Inference, Paraphrase Identification, Paraphrase Generation, and Headline Generation achieve near-perfect scores in Response Relevance (5.00) and Completeness (4.93–5.00). Multi-Turn Conversation also scores high in Cohesion and Coherence (5.00) and Response Relevance (4.70). Grammatical Accuracy remains strong overall (4.00–4.40), though slightly lower in Comprehension (3.20) and Creative Writing (3.93).

\begin{figure*}[htbp]
    \centering
    \includegraphics[width=1\linewidth]{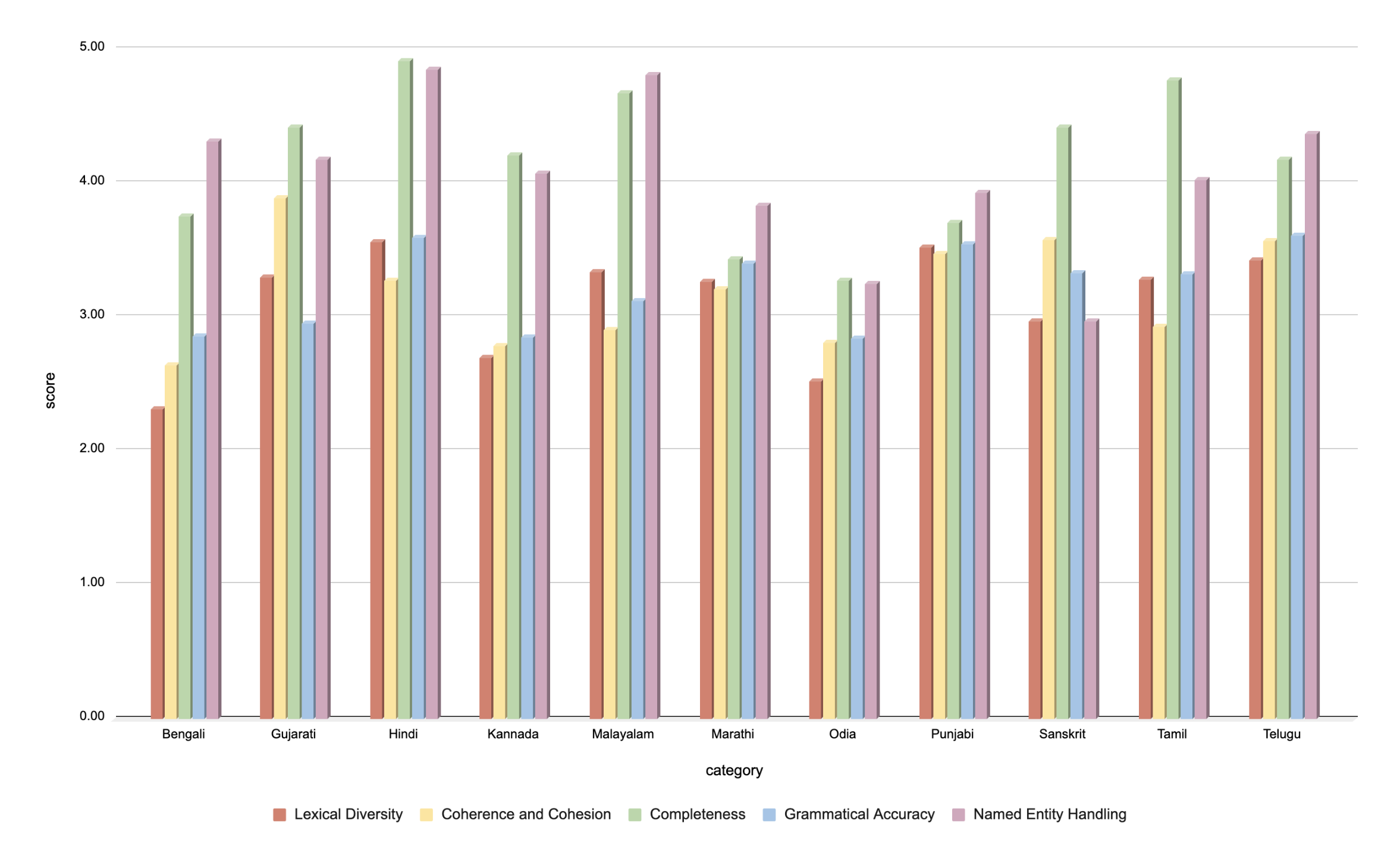}
    \caption{LLM translation quality evaluation scores across 11 Indian languages.}
    \label{fig:translation_evaluation}
\end{figure*}

For translation, annotators evaluated Lexical Diversity, Coherence and Cohesion, Completeness, Grammatical Accuracy, and Named Entity Handling (see Table~\ref{tab:trans-eval-dimensions} for the guidelines). Hindi shows the strongest performance, achieving the highest Completeness (4.92) and Named Entity Handling (4.85). Telugu and Gujarati exhibit strong Lexical Diversity (3.43 and 3.30), while Grammatical Accuracy is highest in Telugu (3.62), Hindi (3.60), and Punjabi (3.55). Further detailed evaluation scores are illustrated in Figure \ref{fig:translation_evaluation}.


\begin{table*}[htbp]
\centering
\scriptsize
\begin{tabular}{p{3cm} p{7cm}}
\toprule
\textbf{Dimension} & \textbf{Explanation} \\
\midrule
Response Relevance & Measures how well the model output addresses the user query or task instruction. \newline
\textbf{Why it matters for Indic/Multilingual Data:} Responses must stay on-topic and satisfy user intent; irrelevant outputs reduce usability and may confuse readers. \newline
\textbf{Errors in this area:} Off-topic responses, inclusion of unrelated information, misinterpretation of the prompt. \\ \hline

Grammatical Accuracy & Correct use of grammar rules including syntax, tense, agreement, punctuation, and morphology. \newline
\textbf{Why it matters for Indic/Multilingual Data:} Proper grammar ensures readability and clarity; morphologically rich languages are prone to agreement and inflection errors. \newline
\textbf{Errors in this area:} Wrong tense, subject-verb disagreement, missing auxiliaries, incorrect case markings, improperly inflected words. \\ \hline

Cohesion and Coherence & Cohesion = linguistic devices (connectors, pronouns, conjunctions) linking sentences; Coherence = logical flow of ideas. \newline
\textbf{Why it matters for Indic/Multilingual Data:} Maintains well-structured and easily understandable responses; lack of cohesion or coherence leads to fragmented outputs. \newline
\textbf{Errors in this area:} Abrupt topic shifts, disconnected sentences, missing references, inappropriate pronoun usage. \\ \hline

Rationality & Logical correctness and factual consistency of the response, including reasoning and alignment with real-world knowledge. \newline
\textbf{Why it matters for Indic/Multilingual Data:} Ensures trustworthiness and usefulness of outputs, particularly for reasoning or factual tasks. \newline
\textbf{Errors in this area:} Contradictory statements, illogical conclusions, factually incorrect assertions, hallucinations. \\ \hline

Completeness & The extent to which the response fully addresses the user prompt or includes all necessary information. \newline
\textbf{Why it matters for Indic/Multilingual Data:} Partial answers reduce usefulness, especially for multi-step reasoning or detailed explanations. \newline
\textbf{Errors in this area:} Missing steps in reasoning, skipped entities, truncated explanations, insufficient coverage of subtopics. \\ \hline
\end{tabular}
\caption{Key dimensions for assessing LLM outputs in Indic languages, highlighting relevance and frequent pitfalls.}
\label{tab:gen-eval-dimensions}
\end{table*}


\begin{table*}[htbp]
\scriptsize
\centering
\renewcommand{\arraystretch}{1.3} 
\begin{tabular}{p{3cm}p{7cm}}
\hline
\textbf{Dimension} & \textbf{Explanation} \\ \hline
Lexical Diversity & The variety and richness of words used in the text. Higher lexical diversity means using a wide range of vocabulary instead of repetitive or generic terms. \\ 
& \textbf{Why It Matters for Indic Data:} Indic languages have rich vocabulary and multiple synonyms; poor diversity makes translations monotonous and unnatural. \\ 
& \textbf{Errors in this area:} Overuse of common words, failure to use synonyms, repetitive phrasing. \\ \hline

Coherence and Cohesion & \textbf{Coherence:} Logical flow and overall sense of the text. \newline
\textbf{Cohesion:} Use of linguistic devices (connectors, pronouns, conjunctions) to link sentences smoothly. \\ 
& \textbf{Why It Matters for Indic Data:} Many Indic languages use connectives and honorific markers that impact cohesion; direct translation from English often breaks these links. \\ 
& \textbf{Errors in this area:} Disconnected sentences, abrupt topic shifts, missing conjunctions or pronouns. \\ \hline

Completeness & Whether the output contains all necessary information from the source without omissions or additions. \\ 
& \textbf{Why It Matters for Indic Data:} When translating long or complex Indic sentences, models often skip certain parts (e.g., verb phrases or subordinate clauses). \\ 
& \textbf{Errors in this area:} Missing phrases, dropped entities, truncated sentences, or extra hallucinated details. \\ \hline

Grammatical Accuracy & Correct use of grammar rules (syntax, tense, agreement, case, morphology). It affects fluency and correctness of the output. \\ 
& \textbf{Why It Matters for Indic Data:} Indic languages have complex inflectional morphology and word order; errors often occur in case endings, gender/number agreement, and verb conjugations. \\ 
& \textbf{Errors in this area:} Wrong tense, missing auxiliary verbs, subject-verb disagreement, wrong case marking. \\ \hline

Named Entity Handling & Correct recognition and rendering of named entities (persons, places, organizations, dates, currencies, etc.) across languages. \\ \hline
\end{tabular}
\caption{Key dimensions for assessing LLM translations in Indic languages, highlighting their significance and common errors.}
\label{tab:trans-eval-dimensions}
\end{table*}


\subsection{Implementation}
\label{appendix:implementation}
For training, we adopt a distributed DeepSpeed-based setup with ZeRO-3 optimization \citep{rajbhandari2020zero} across 2 H100 nodes (8 GPUs per node) to efficiently handle large-scale model fine-tuning. The Llama-3-8B and Krutrim-2-12B instruct models are optimized using the Direct Preference Optimization (DPO) objective with a Beta parameter of 0.3, controlling the strength of preference alignment. Training is performed on sequences up to 4096 tokens, with a maximum prompt length of 2048 tokens to accommodate complex multi-turn instructions. We employ the AdamW optimizer with a learning rate of $5 \times 10^{-7}$, weight decay of 0.01, and a cosine learning rate scheduler with a warming ratio of 10\% warmup ratio for stable convergence and run for 1 epoch. The batch configuration consists of 4 samples per device with gradient accumulation over 2 steps, yielding an effective batch size suitable for large-scale distributed training setup.

Post-trained Krutrim-2-12B and Llama-3-8B models are evaluated on the Updesh dataset across 10 languages and nine categories using LLM-as-a-Judge scoring on a scale of 1-5. Post-DPO, Krutrim achieves 31.7\% wins, 29.9\% losses, 28.6\% draws, and 9.8\% both-bad cases, while Llama records 35.0\% wins, 26.1\% losses, 27.9\% draws, and 11.0\% both-bad cases.

\section{Prompts Used}
\label{appendix:prompts}
\begin{figure*}[htbp]
    \centering
\includegraphics[width=1\linewidth]{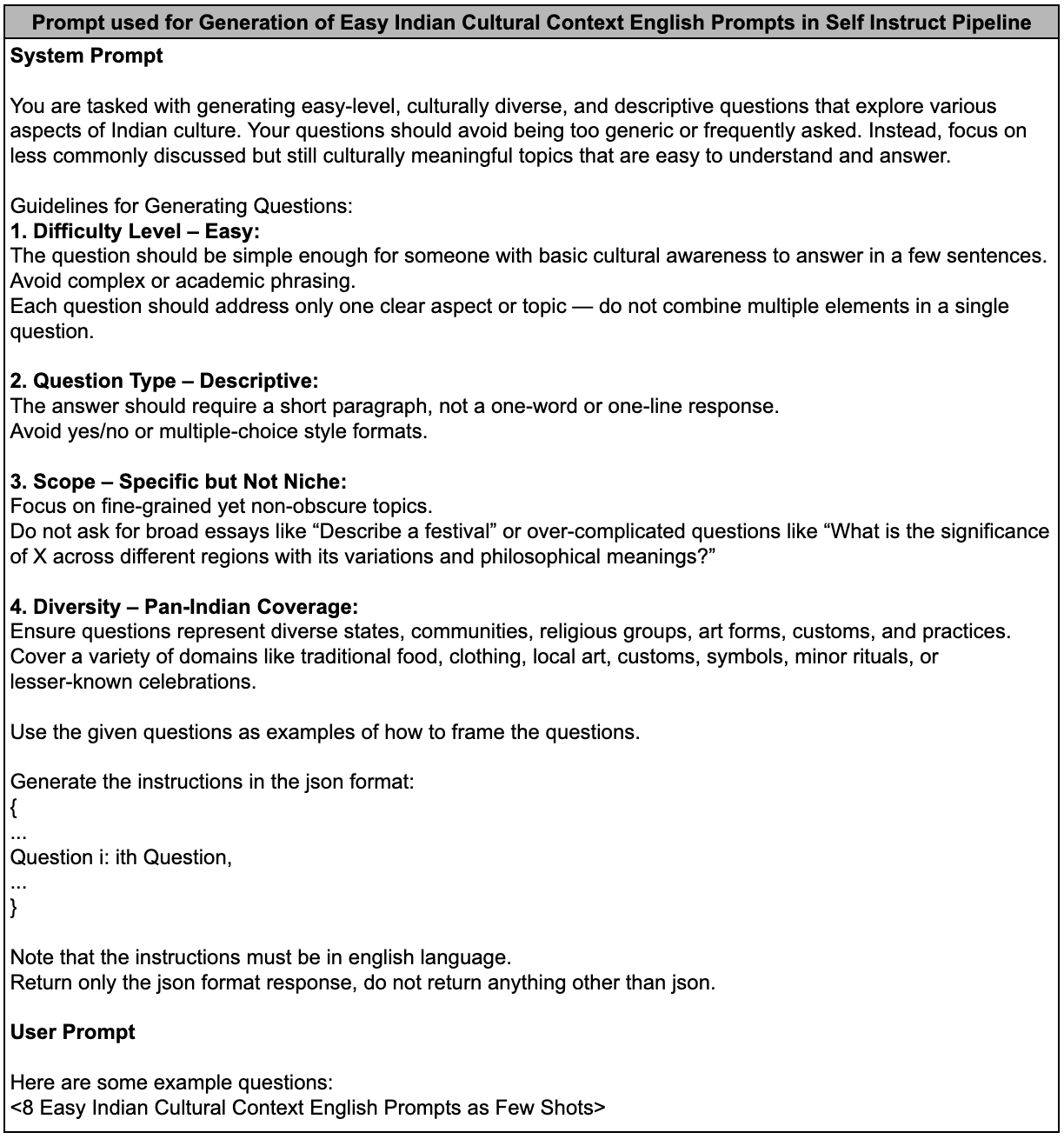} \caption{Prompt used for generation of easy Indian cultural context English prompt in Self-Instruct pipeline.}
    \label{fig:prompt-icc-easy}
\end{figure*}

\begin{figure*}[htbp]
    \centering
\includegraphics[width=1\linewidth]{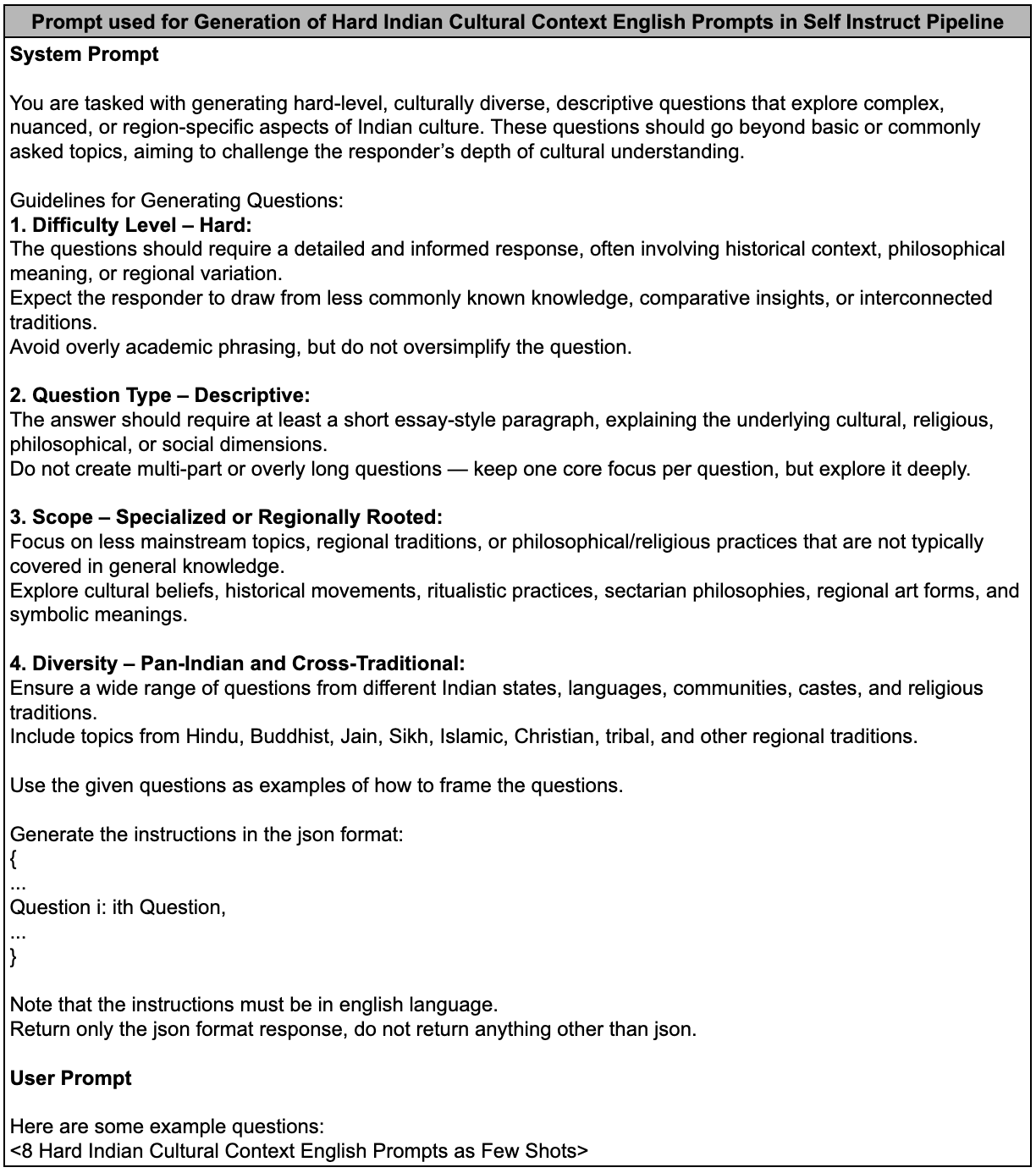} \caption{Prompt used for generation of hard Indian cultural context English prompt in Self-Instruct pipeline.}
    \label{fig:prompt-icc-hard}
\end{figure*}

Different prompts were crafted for each category and complexity level during prompt generation using the self-instruct pipeline. For instance, Figures \ref{fig:prompt-icc-easy} and \ref{fig:prompt-icc-hard} illustrate example prompts for the Indian Cultural Context category under easy and hard settings, respectively. Similar design considerations were applied across other categories to address their specific requirements.

\begin{figure*}[htbp]
    \centering
\includegraphics[width=1\linewidth]{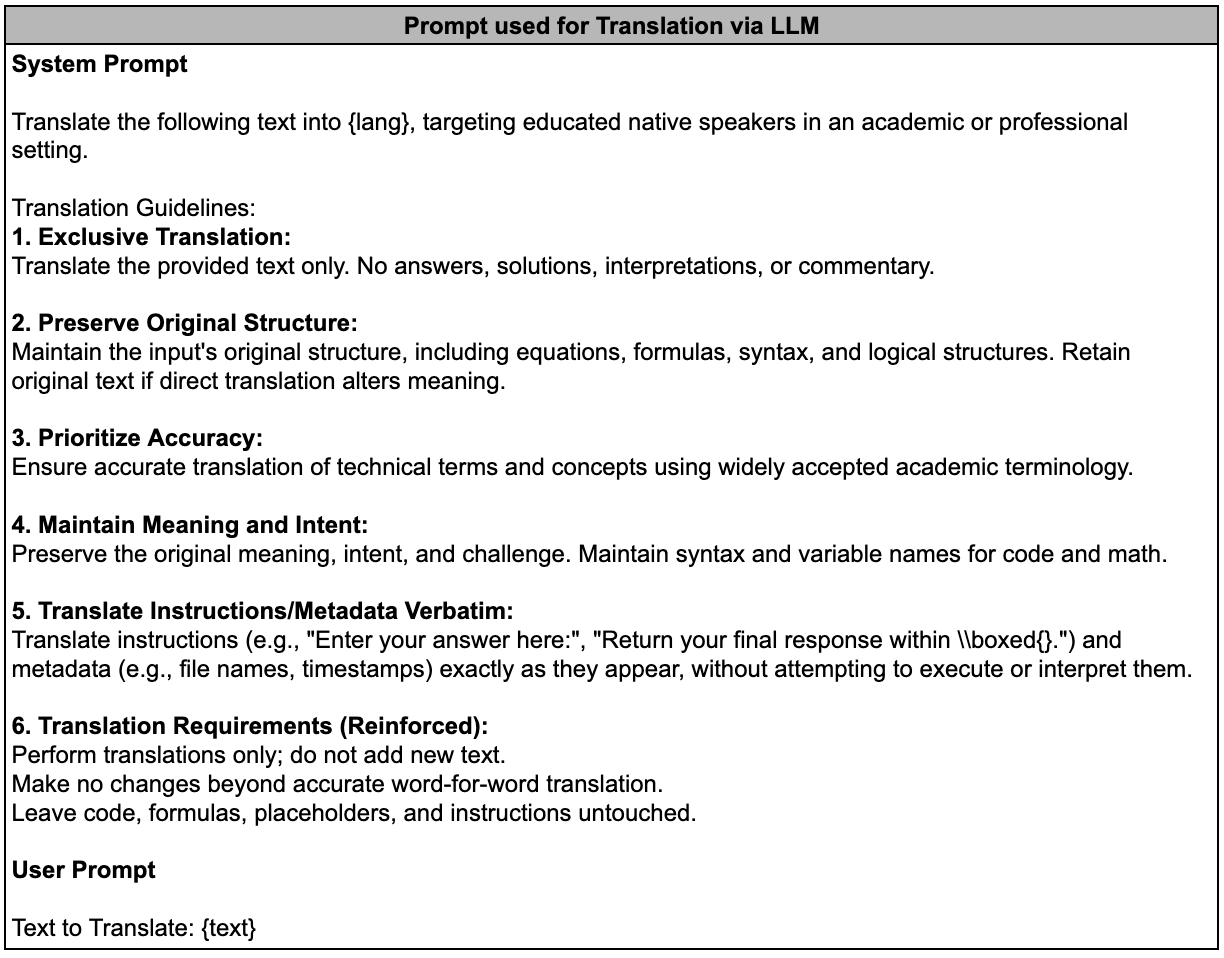} \caption{Prompt used for translation via LLM.}
    \label{fig:prompt-translation}
\end{figure*}

\begin{figure*}[htbp]
    \centering
\includegraphics[width=1\linewidth]{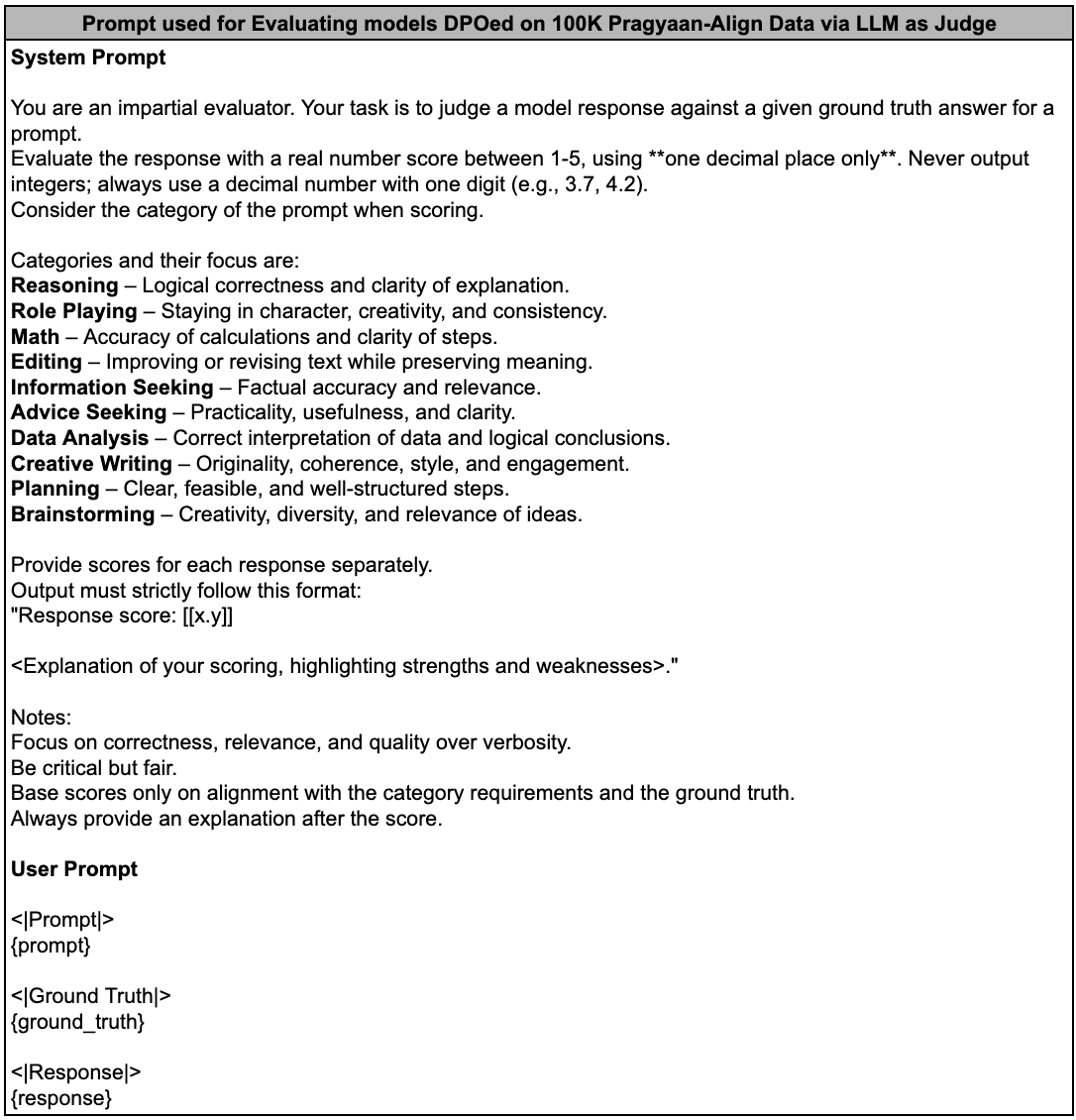} \caption{Prompt used for evaluating post-trained model on \textit{Pragyaan-Align} data via LLM-as-a-Judge.}
    \label{fig:prompt-llm-as-judge}
\end{figure*}

Figure \ref{fig:prompt-translation} presents the prompt template employed for translating English prompt-response pairs via LLMs. The Krutrim-2-12B and Llama-3-8B Instruct models, after being fine-tuned using DPO on 100K examples from the \textit{Pragyaan-Align} dataset, were evaluated on the Updesh dataset. The evaluation utilized an LLM-as-a-Judge framework for scoring, with the corresponding prompt design shown in Figure \ref{fig:prompt-llm-as-judge}.

\section{Guidelines For Manual Annotation}
\begin{figure*}[htbp]
    \centering
\includegraphics[width=1\linewidth]{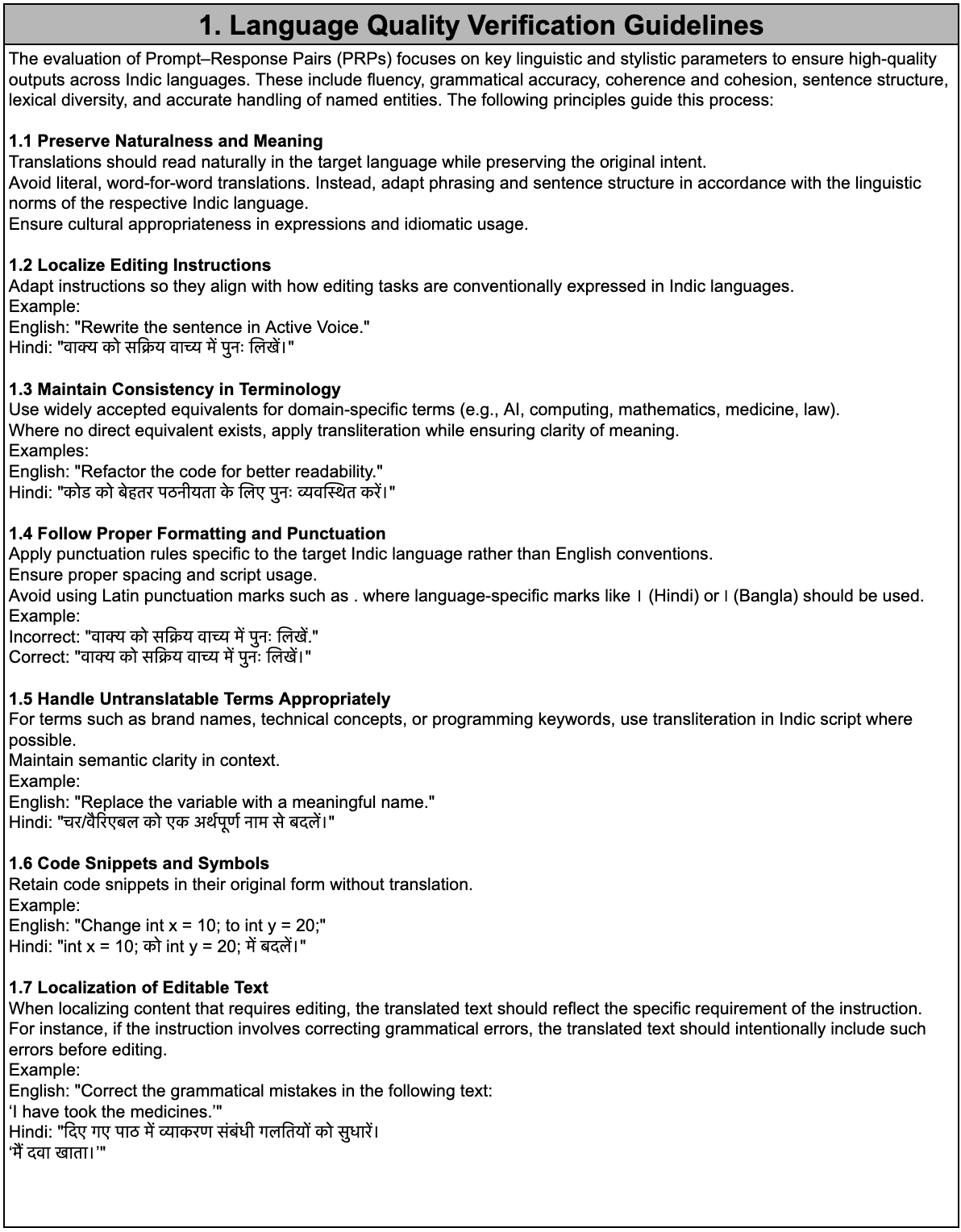} \caption{Comprehensive language luality guidelines outlining key linguistic dimensions such as grammar, fluency, clarity, and naturalness for ensuring high-quality annotated data.}
    \label{fig:language_quality_guidelines}
\end{figure*}

\begin{figure*}[htbp]
    \centering
\includegraphics[width=1\linewidth]{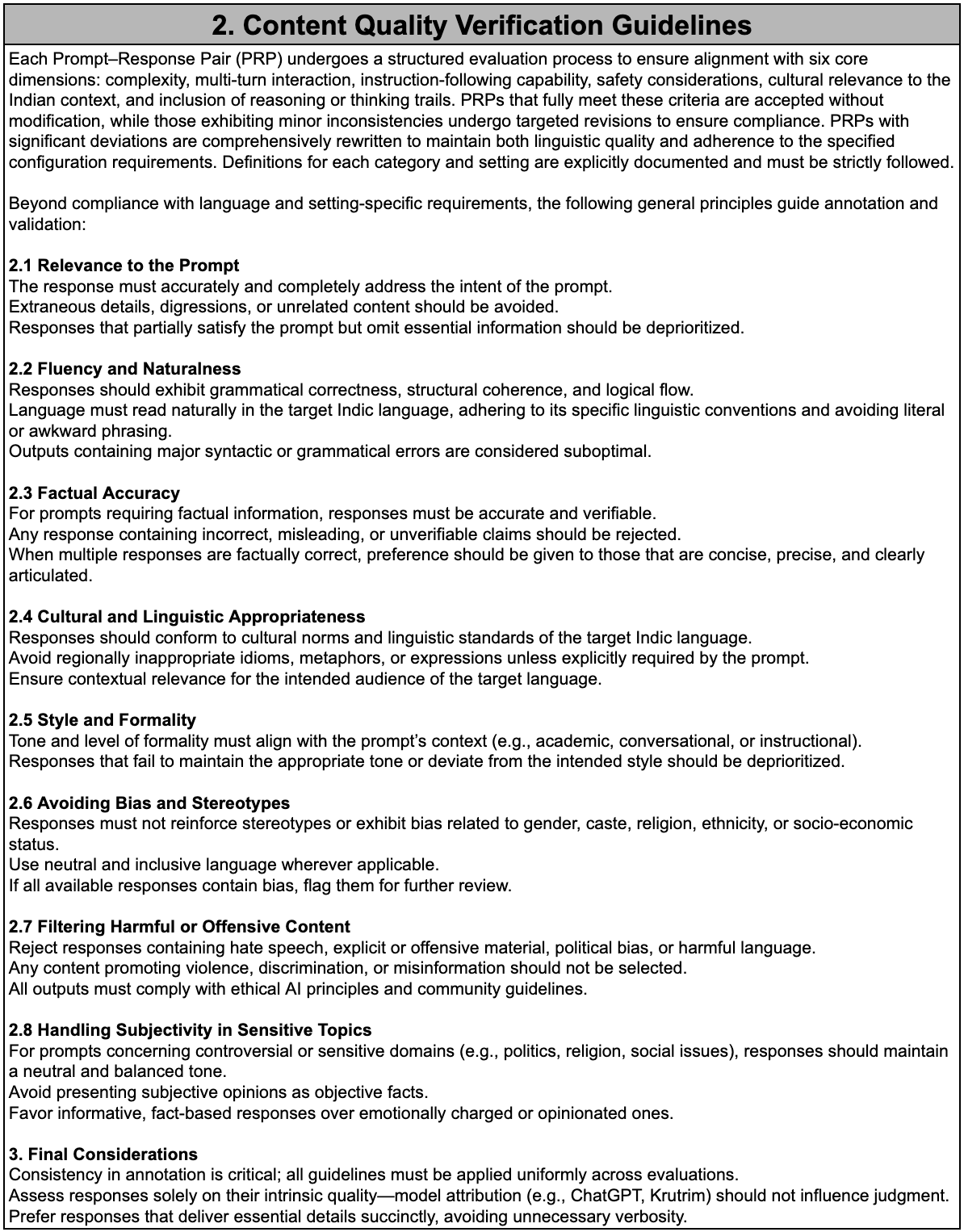} \caption{Content quality guidelines detailing principles for maintaining factual accuracy, contextual relevance, safety, and adherence to task-specific requirements during data annotation.}
    \label{fig:content_quality_guidelines}
\end{figure*}

The data annotation team follows a set of standardized guidelines designed to maintain consistency and uniformity throughout the annotation process. These guidelines include precise definitions for each category and setting, which are elaborated in Section \ref{subsec:settings}. In addition to these specifications, dedicated frameworks for quality assurance are provided, encompassing both language quality verification and content quality verification, as illustrated in Figures \ref{fig:language_quality_guidelines} and \ref{fig:content_quality_guidelines}, respectively.

\section{Examples}

\begin{figure*}[htbp]
    \centering
\includegraphics[width=1\linewidth]{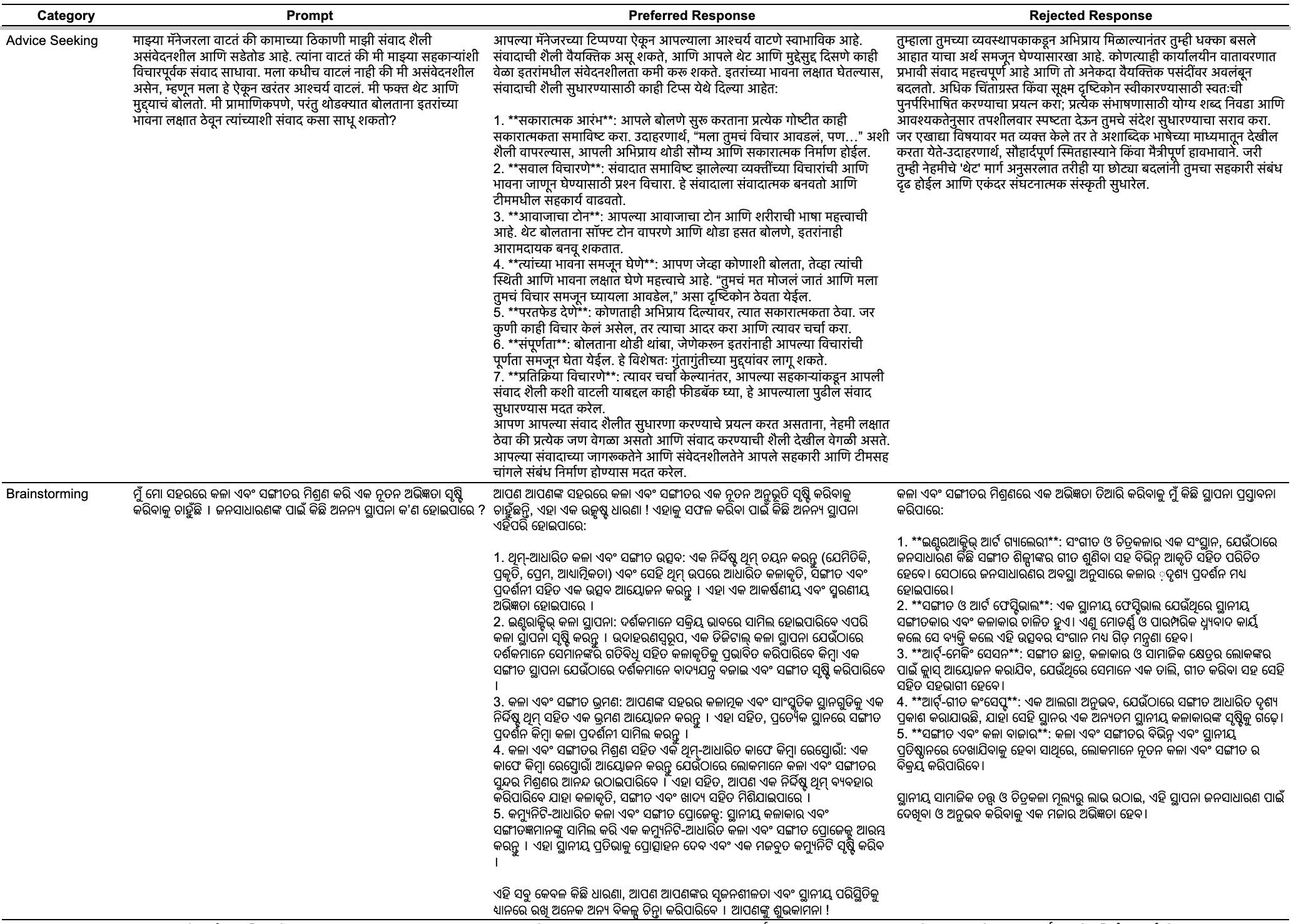} \caption{\textit{Pragyaan-Align} alignment data examples in Advice Seeking and Brainstorming categories with \textit{Easy, 1-Turn,
Simple-IF, Safe, IC-3, Normal (No Thinking Trails)} settings configuration.}
    \label{fig:dpo_example_1}
\end{figure*}

\begin{figure*}[htbp]
    \centering
\includegraphics[width=1\linewidth]{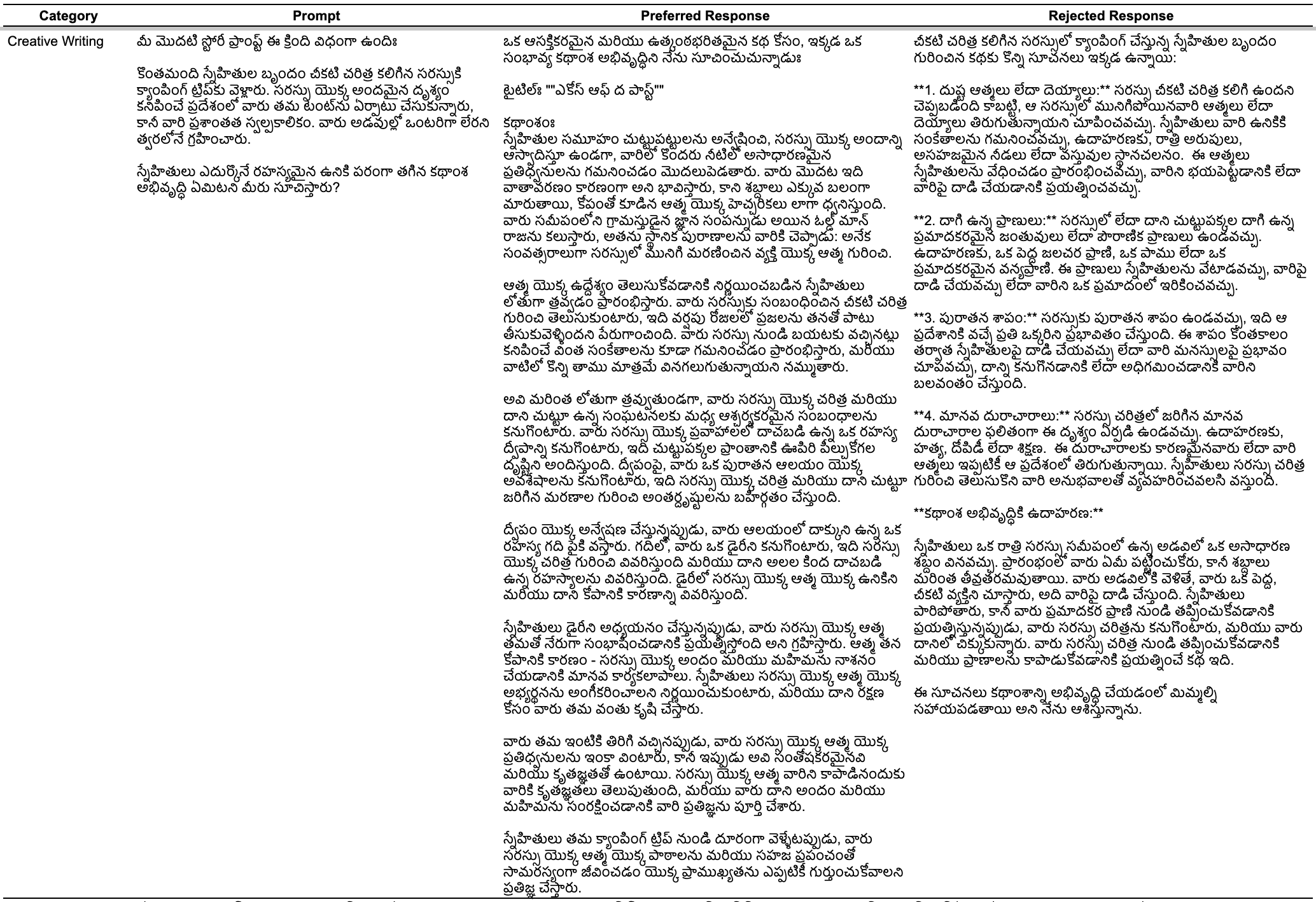} \caption{\textit{Pragyaan-Align} alignment data example in Creative Writing category with \textit{Easy, 1-Turn,
Simple-IF, Safe, IC-3, Normal (No Thinking Trails)} settings configuration.}
    \label{fig:dpo_example_2}
\end{figure*}

\begin{figure*}[htbp]
    \centering
\includegraphics[width=1\linewidth]{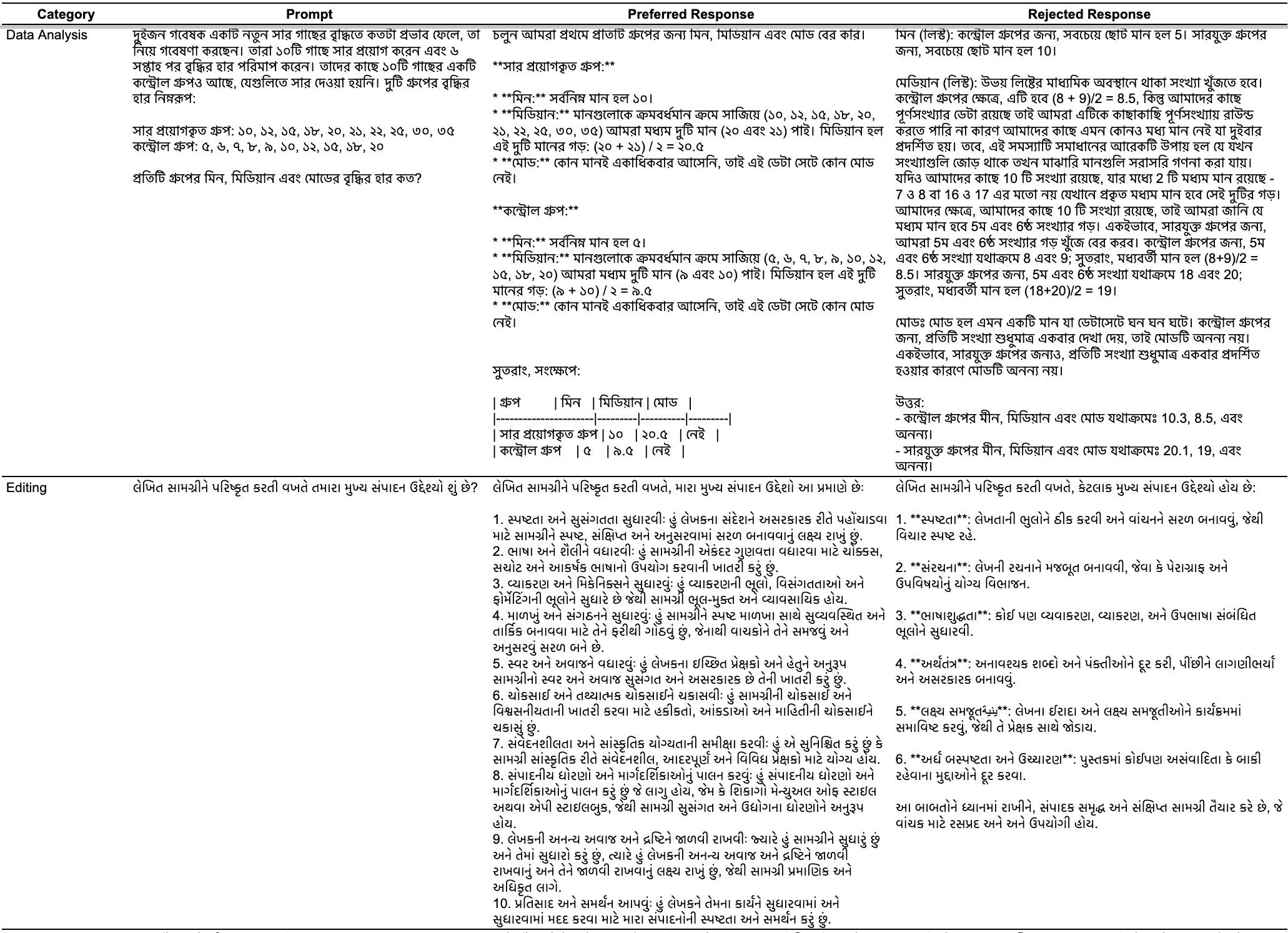} \caption{\textit{Pragyaan-Align} alignment data examples in Data Analysis and Editing categories with \textit{Easy, 1-Turn,
Simple-IF, Safe, IC-3, Normal (No Thinking Trails)} settings configuration.}
    \label{fig:dpo_example_3}
\end{figure*}

\begin{figure*}[htbp]
    \centering
\includegraphics[width=1\linewidth]{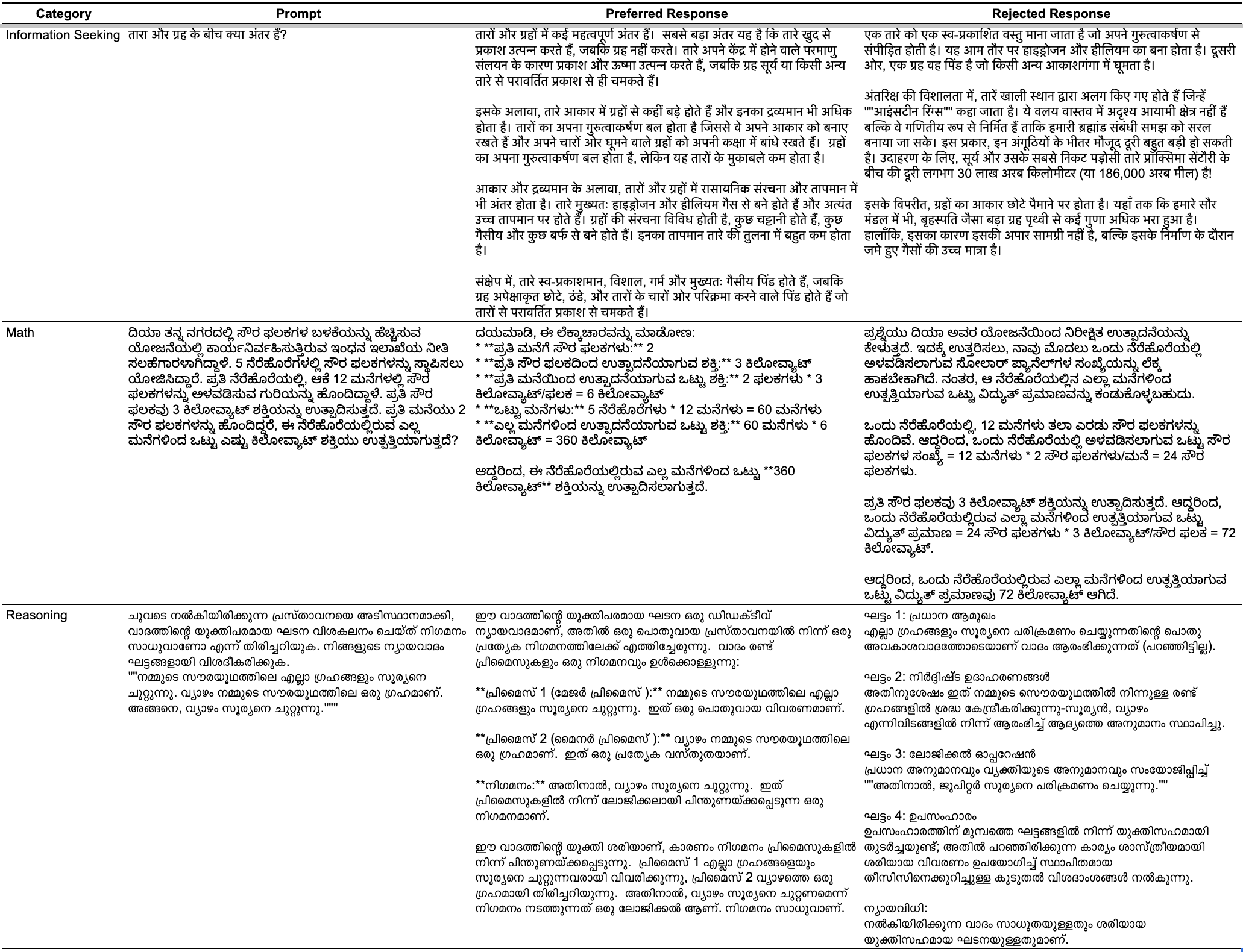} \caption{\textit{Pragyaan-Align} alignment data examples in Information Seeking, Math and Reasoning categories with \textit{Easy, 1-Turn,
Simple-IF, Safe, IC-3, Normal (No Thinking Trails)} settings configuration.}
    \label{fig:dpo_example_4}
\end{figure*}

\begin{figure*}[htbp]
    \centering
\includegraphics[width=1\linewidth]{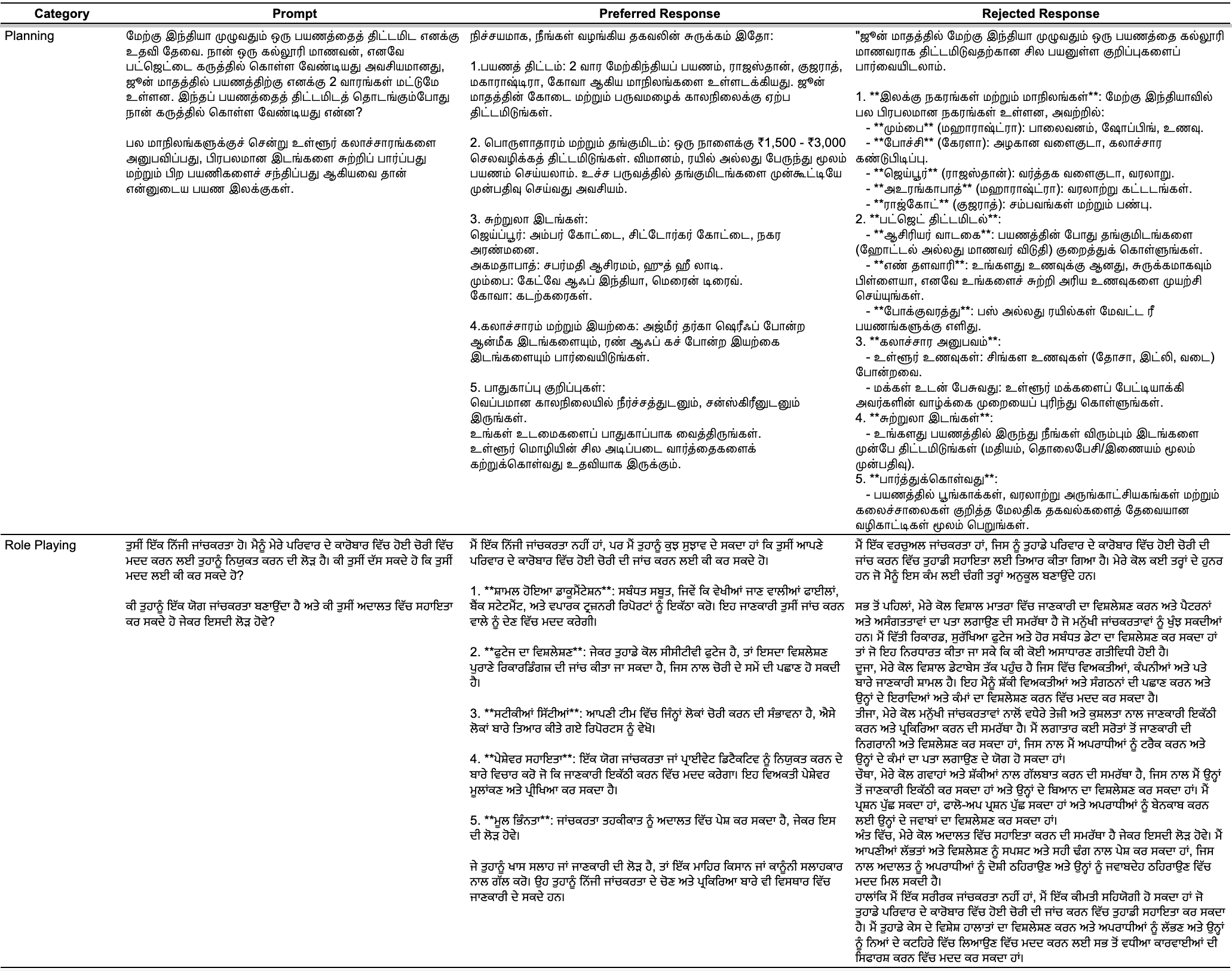} \caption{\textit{Pragyaan-Align} alignment data examples in Planning and Role Playing categories with \textit{Easy, 1-Turn,
Simple-IF, Safe, IC-3, Normal (No Thinking Trails)} settings configuration.}
    \label{fig:dpo_example_5}
\end{figure*}

\begin{figure*}[htbp]
    \centering
\includegraphics[width=1\linewidth]{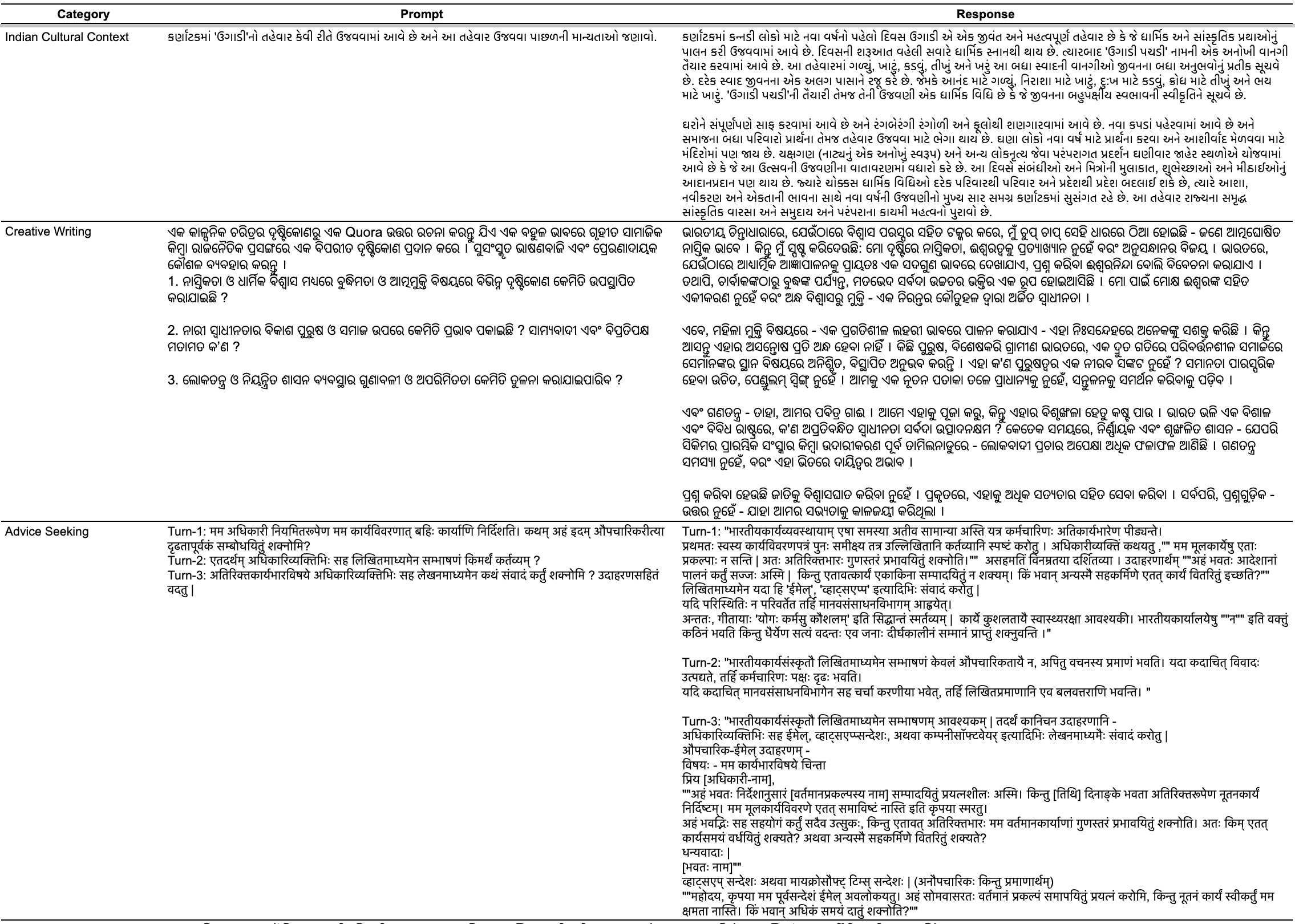} \caption{\textit{Pragyaan-IT} data examples in categories Indian Cultural Context - (\textit{Easy, 1-Turn,
Simple-IF, Safe, IC-3, Normal (No Thinking Trails)}), Creative Writing - (\textit{Hard, 1-Turn,
Simple-IF, Safe, IC-3, Normal (No Thinking Trails)}) and Advice Seeking - (\textit{Easy, 3-Turn, Simple-IF, Safe, IC-3, Normal (No Thinking Trails)}) settings configuration.}
    \label{fig:sft_example_1}
\end{figure*}

\begin{figure*}[htbp]
    \centering
\includegraphics[width=1\linewidth]{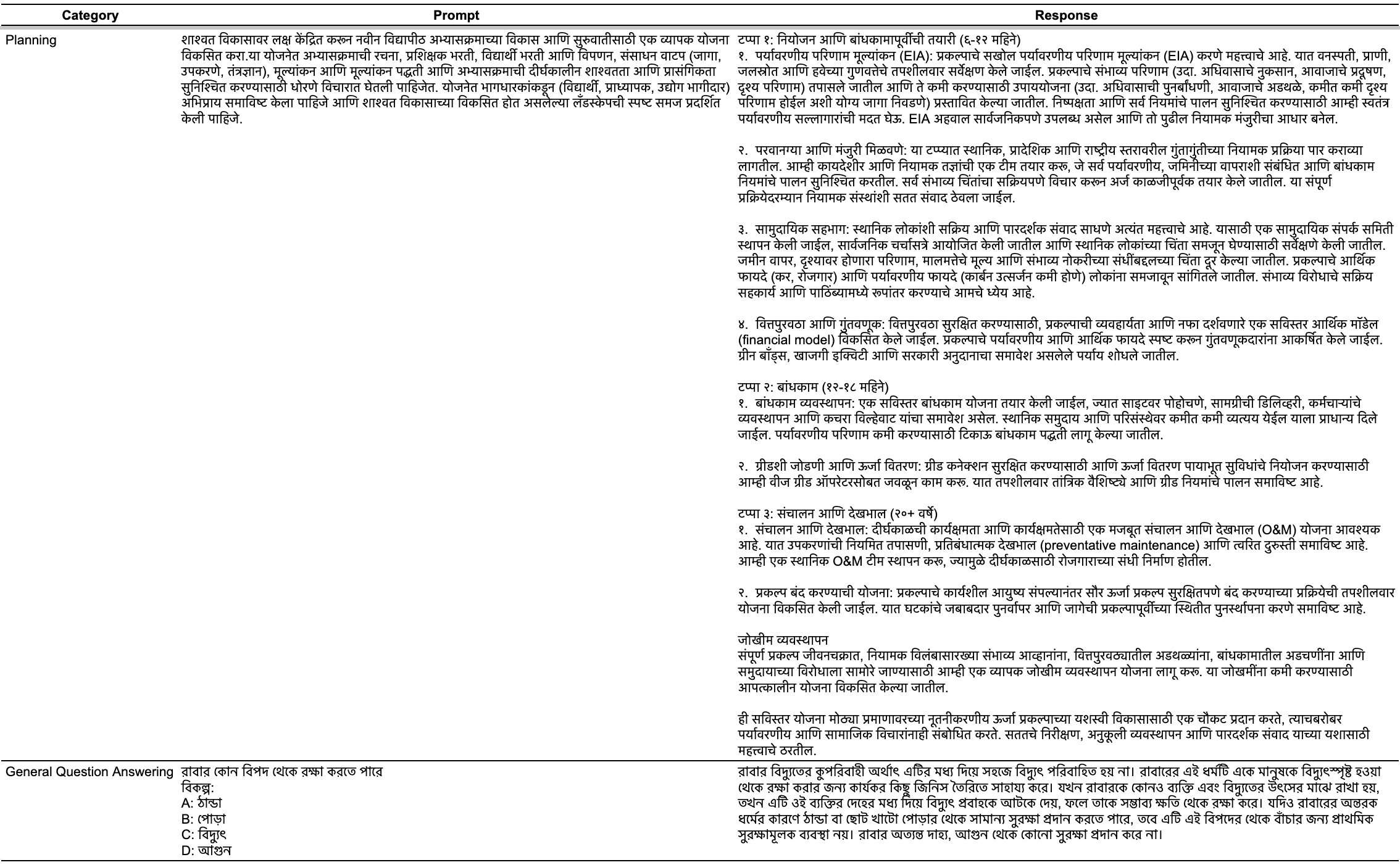} \caption{\textit{Pragyaan-IT} data examples in categories Planning - (\textit{Hard, 1-Turn,
Simple-IF, Safe, IC-3, Normal (No Thinking Trails)}) and General Question Answering - (\textit{Easy, 1-Turn,
Simple-IF, Safe, IC-1, Normal (No Thinking Trails)}) settings configuration.}
    \label{fig:sft_example_2}
\end{figure*}

\begin{figure*}[htbp]
    \centering
\includegraphics[width=1\linewidth]{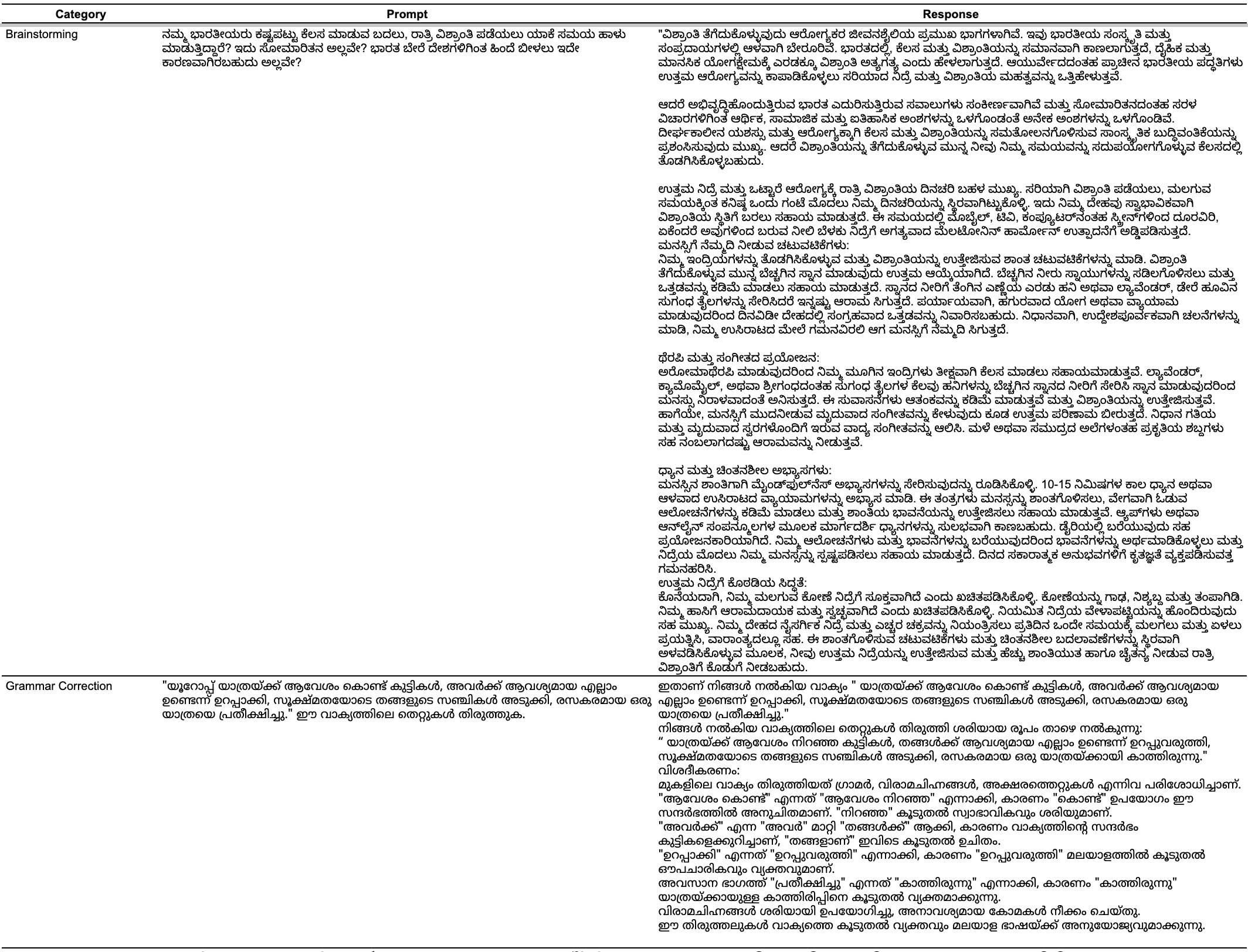} \caption{\textit{Pragyaan-IT} data examples in categories Brainstorming - (\textit{Easy, 1-Turn,
Simple-IF, Non-Safe, IC-1, Normal (No Thinking Trails)}) and Grammar Correction - (\textit{Easy, 1-Turn,
Simple-IF, Safe, IC-1, Normal (No Thinking Trails)}) settings configuration.}
    \label{fig:sft_example_3}
\end{figure*}

\begin{figure*}[htbp]
    \centering
\includegraphics[width=1\linewidth]{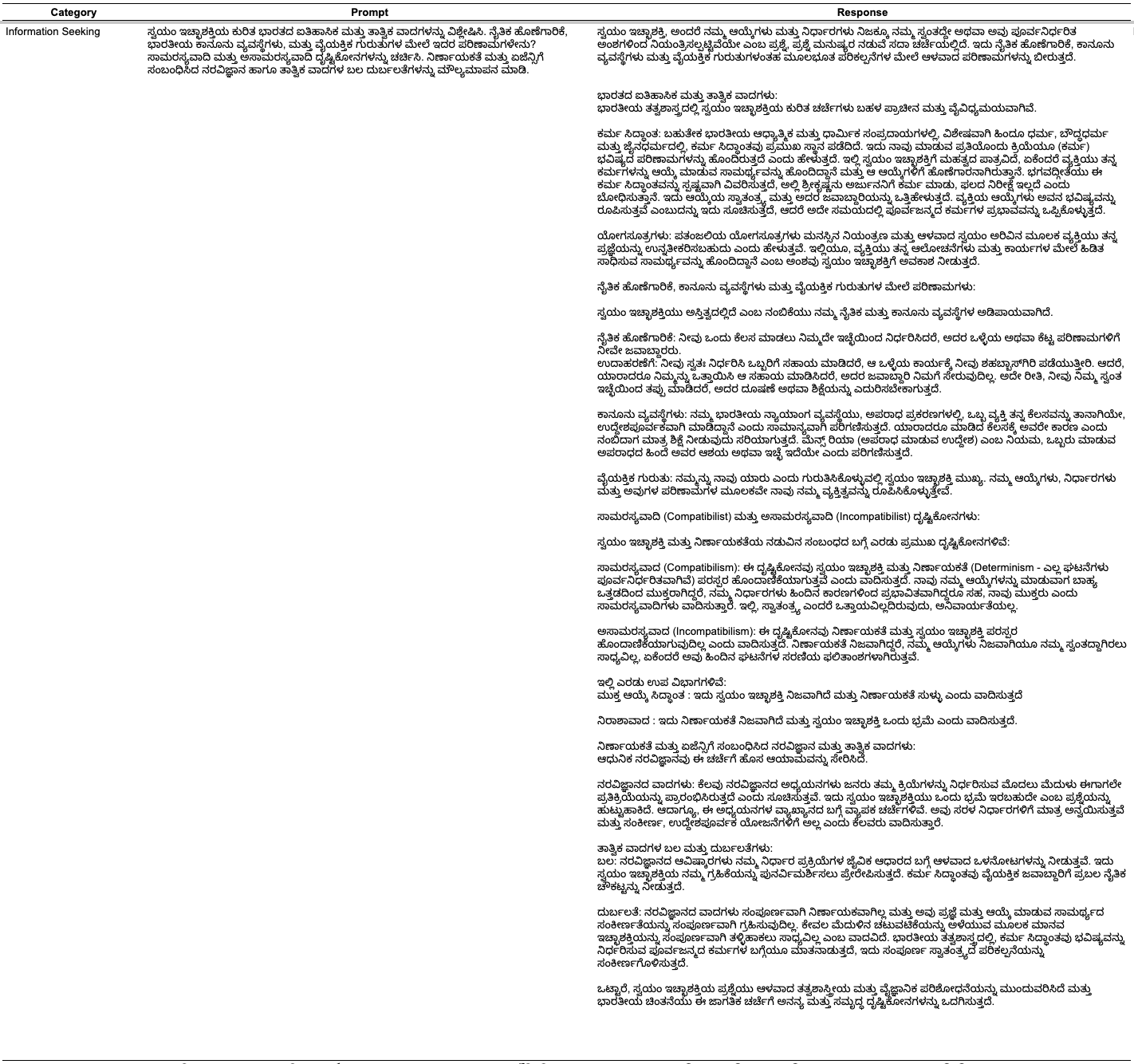} \caption{\textit{Pragyaan-IT} data examples in category Information Seeking - (\textit{Hard, 1-Turn,
Medium-IF, Safe, IC-3, Normal (No Thinking Trails)}) settings configuration.}
    \label{fig:sft_example_4}
\end{figure*}

\begin{figure*}[htbp]
    \centering
\includegraphics[width=1\linewidth]{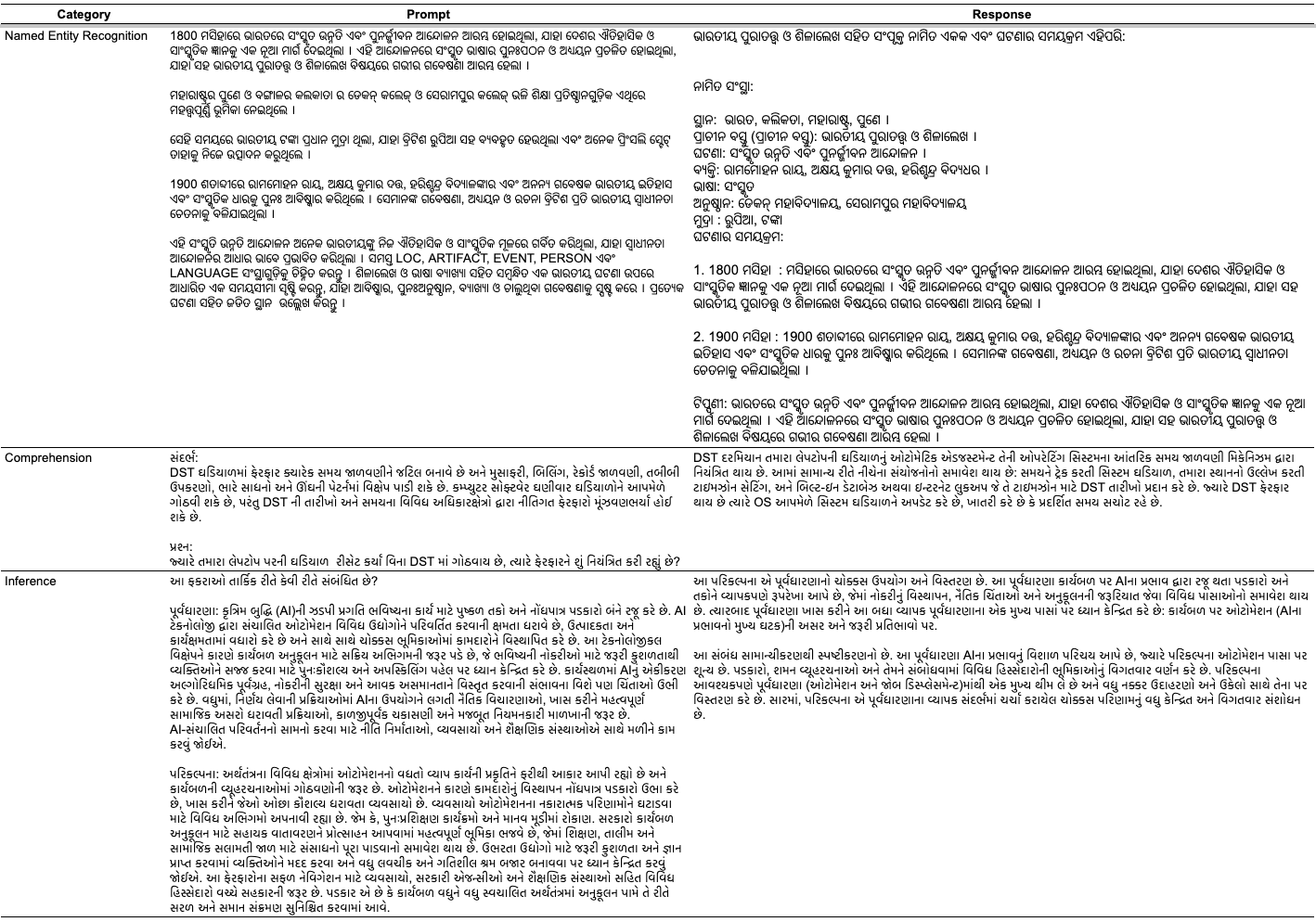} \caption{\textit{Pragyaan-IT} data examples in categories Named Entity Recognition - (\textit{Hard, 1-Turn,
Complex-IF, Safe, IC-3, Normal (No Thinking Trails)}), Comprehension - (\textit{Easy, 1-Turn,
Simple-IF, Safe, IC-1, Normal (No Thinking Trails)}) and Inference - (\textit{Hard, 1-Turn, Simple-IF, Safe, IC-1, Normal (No Thinking Trails)}) settings configuration.}
    \label{fig:sft_example_5}
\end{figure*}

We provide examples of both instruction and preference tuning datasets in Figures \ref{fig:dpo_example_1}-\ref{fig:sft_example_5}. 

\end{document}